\documentclass[10pt,journal]{IEEEtran}

\ifCLASSOPTIONcompsoc

  \usepackage[nocompress]{cite}
\else
  \usepackage{cite}
\fi

\usepackage{multirow}
\usepackage{makecell}
\usepackage{graphicx}
\usepackage{times}
\usepackage{epsfig}
\usepackage{amsmath}
\usepackage{amssymb}
\usepackage{multicol}
\usepackage{balance}
\usepackage{multirow}
\usepackage{float}
\usepackage{caption}
\usepackage{ctable}
\usepackage{array}
\usepackage{makecell}
\usepackage{subfig}
\usepackage{tabularx}
\usepackage{bbding}
\usepackage{pifont}
\usepackage{url}
\usepackage{times}
\usepackage{epsfig}
\usepackage{amsmath}
\usepackage{amssymb}
\usepackage{colortbl}
\usepackage{color}
\usepackage{threeparttable}
\usepackage{booktabs}
\usepackage{adjustbox}
\usepackage{subfig}
\usepackage{bm}
\usepackage{xfrac}
\usepackage{tabto,enumitem}
\usepackage[linesnumbered,lined,boxed,commentsnumbered,ruled]{algorithm2e}
\usepackage[breaklinks=true,colorlinks,citecolor=blue,urlcolor=blue,linkcolor=blue,bookmarks=false]{hyperref} % pagebackref

\usepackage{xspace}
\usepackage{threeparttable}

\usepackage{xspace}
\makeatletter
\DeclareRobustCommand\onedot{\futurelet\@let@token\@onedot}
\def\@onedot{\ifx\@let@token.\else.\null\fi\xspace}

\begin{document}

\title{Deep Learning Based Domain Adaptation Methods in Remote Sensing: A Comprehensive Survey}
\newcommand{\gl}[1]{{\color{blue}(guangliang: {#1})}} % guangliang's comments
\author{Shuchang Lyu,~\IEEEmembership{Member,~IEEE}, Qi Zhao,~\IEEEmembership{Member,~IEEE}, Zheng Zhou, Meng Li, You Zhou, Dingding Yao, Guangliang Cheng, Huiyu Zhou, Zhenwei Shi,~\IEEEmembership{Senior Member,~IEEE}
% <-this % stops a space
\thanks{This work was supported by ``the Fundamental Research Funds for the Central Universities'' (grant number 501QYJC2025102002).}
\thanks{Shuchang Lyu, Qi Zhao, Zheng Zhou, Meng Li and You Zhou are with the Department of Electronic and Information Engineering, Beihang University. Zhenwei Shi are with the Image Processing Center, School of Astronautics, Beihang University. E-mail: \{lyushuchang, zhaoqi, zhengzhou, limenglm, sy2402322, shizhenwei\}@buaa.edu.cn}
\thanks{Dingding Yao is with Institute of Acoustics, Chinese Academy of Sciences. \{yaodingding\}@hccl.ioa.ac.cn}
\thanks {Guangliang Cheng is with the Department of Computer Science, at the University of Liverpool. \{Guangliang.Cheng\}@liverpool.ac.uk.}
\thanks {Huiyu Zhou is with the School of Computing and Mathematical Sciences, at the University of Leicester \{hz143\}@leicester.ac.uk.}
\thanks{Corresponding author: Qi Zhao} 
}

% The paper headers
\markboth{IEEE Geoscience and Remote Sensing Magazine}%
{Lyu \MakeLowercase{\textit{et al.}}: Bare Demo of IEEEtran.cls for IEEE Journals}

\maketitle

\IEEEpeerreviewmaketitle

%\IEEEtitleabstractindextext{
\begin{abstract} % 1/4 page
\par Domain adaptation is a crucial and increasingly important task in remote sensing, aiming to transfer knowledge from a source domain a differently distributed target domain. It has broad applications across various real-world applications, including remote sensing element interpretation, ecological environment monitoring, and urban/rural planning. However, domain adaptation in remote sensing poses significant challenges due to differences in data, such as variations in ground sampling distance, imaging modes from various sensors, geographical landscapes, and environmental conditions. In recent years, deep learning has emerged as a powerful tool for feature representation and cross-domain knowledge transfer, leading to widespread adoption in remote sensing tasks. In this paper, we present a comprehensive survey of significant advancements in deep learning based domain adaptation for remote sensing. We first introduce the preliminary knowledge to clarify key concepts, mathematical notations, and the taxonomy of methodologies. We then organize existing algorithms from multiple perspectives, including task categorization, input mode, supervision paradigm, and algorithmic granularity, providing readers with a structured understanding of the field. Next, we review widely used datasets and summarize the performance of state-of-the-art methods to provide an overview of current progress. We also identify open challenges and potential directions to guide future research in domain adaptation for remote sensing. Compared to previous surveys, this work addresses a broader range of domain adaptation tasks in remote sensing, rather than concentrating on a few subfields. It also presents a systematic taxonomy, providing a more comprehensive and organized understanding of the field. As a whole, this survey can inspire the research community, foster understanding, and guide future work in the field.
\end{abstract}
%}
\section{Introduction} % 1.5 pages 
  \label{intro}
  \IEEEPARstart{R}{emote} sensing (RS) technology has been extensively utilized across various real-world applications, including remote sensing element interpretation~\cite{AID, NWPU, MGML}, ecological environment monitoring~\cite{CNRIEEEMC, RSMonitoring}, urban/rural planning~\cite{LoveDA, CITY-OSM, ISPRS}, etc. Over the past decade, advancements in deep learning, such as deep convolutional neural networks~\cite{VGG, ResNet, DenseNet}, Transformers~\cite{ViT, Swin} and Mamba~\cite{Mamba, Mamba2, Vim}, have significantly accelerated the development of remote sensing applications. However, their effectiveness often relies on costly and labor-intensive training samples with annotations that adequately cover complex scenes. In practical scenarios, the scarcity of such samples frequently fails to adequately address the discrepancies between the training (source) and testing (target) images, leading to a notable deterioration in performance. This domain shift phenomenon and the subsequent gap between source and target domains pose significant challenges. To mitigate this issue and bridge the domain gap effectively, domain adaptation methods for remote sensing have emerged as a pivotal research topic. These methods strive to enhance the generalization capabilities of RS models, enabling them to perform robustly across diverse and unseen domains, thus overcoming the limitations imposed by insufficient and biased training data.
  \begin{figure}[!t]
  \centering
\includegraphics[width=0.48\textwidth]{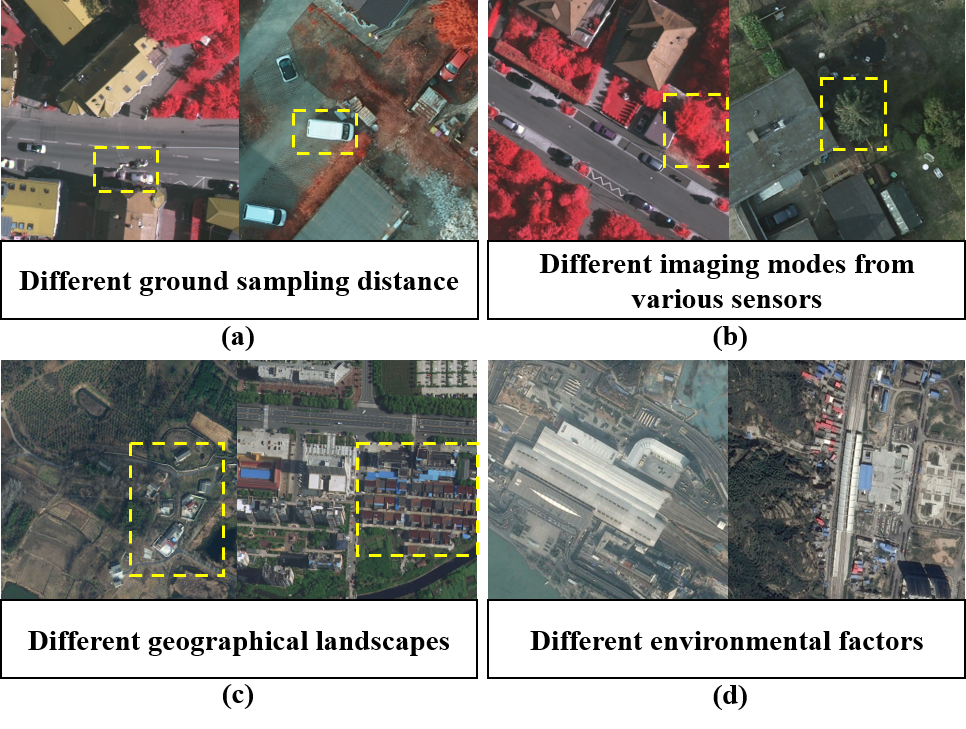}
  \caption{Illustration of the cases to explain the main challenges of domain adaptation methods in remote sensing. Different colors represent specific sections. Best viewed in color.}
  \label{Fig1}
\end{figure}
\begin{figure*}
  \centering
\includegraphics[width=1.0\textwidth]{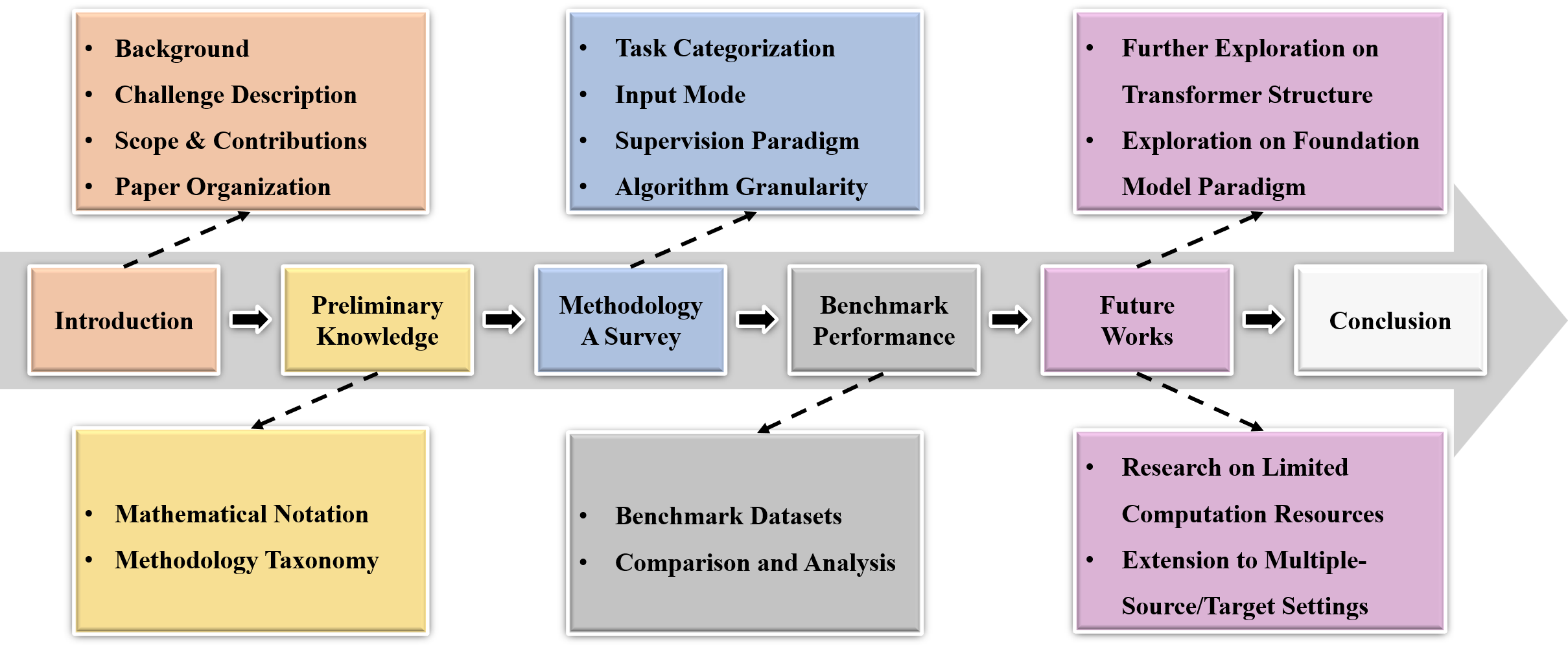}
  \caption{Illustration of the pipeline of this survey. Different colors represent specific sections. Best viewed in color.}
  \label{Fig2}
\end{figure*}
\par Deep learning based domain adaptation methods are designed to train a model using source dataset and then transfer its capabilities to generate accurate predictions on target dataset. In the realm of remote sensing scenes, several challenges persist, encompassing the following aspects. (1) Severe variation caused from different ground sampling distances. As shown in Fig.~\ref{Fig1} (a), objects such as ``cars'' in images captured with varying ground sampling distances exhibit obvious differences in characteristics. (2) Discrepancy of certain categories between source and target images due to the utilization of different imaging modes. As illustrated in Fig.~\ref{Fig1} (b), the color of ``trees'' varies between R-G-B and IR-R-G images. (3) Discrepancy of same categories under different geographical landscapes. As depicted in Fig.~\ref{Fig1}(c), ``buildings`` in rural and urban exhibit distinct architectural patterns. (4) Large domain shifts occur due to different environmental factors (weather conditions, illumination, shadows, etc.). Fig.~\ref{Fig1} (d) shows the contrast between a ``railway station'' scene with and without mist. As a whole, these challenges result in varying degrees of data distribution discrepancies, leading to a domain shift problem between source and target domain images.
\par To alleviate the domain shift and bridge the domain gap between the source and target images in remote sensing scenes, numerous domain adaptation methods have been proposed. Before deep learning era, domain adaptation methods mainly focus on traditional methods~\cite{zhang2016DAsurvey} such as invariant feature selection~\cite{ham2005TGRS, bruzzone2009TGRS, Tuia2014TGRS, persello2015TGRS}, data distribution adaptation~\cite{nielsen2007TIP, nielsen2009ISPRSXV, matasci2015TGRS, volpi2015ISPRS}, etc., with the aim of adjusting the distribution discrepancy between the source and target domains. In the past decade, the rapid development of deep learning~\cite{zhu2017survey, zhang2016survey}, especially deep neural networks, has promoted the development of domain adaptation methods in remote sensing field. Initially, adversarial learning emerges as the predominant technology~\cite{Wang2018IGARSS, Wang2019GRSL, Wang2021JSTARs, Li2021ISPRS, Bai2022TGRS, Zhao2023RS, Liu2022PR, Huang2024GRSL}, which is primarily utilized for feature-level or image-level alignment. As another remarkable technique, self-training has garnered considerable attention in the realm of remote sensing domain adaptation tasks~\cite{wang2024Arxiv, Jin2024TGRS, Yang2024RS, Wang2024JSTARs, Du2024TGRS, Luo2024TGRS}. This non-adversarial paradigm enhances adaptation capabilities by generating reliable, consistent, and class-balanced pseudo labels for target domain images. As adversarial learning and self-training enhance the efficiency of domain adaptation methods from distinct perspectives, the integration of these two techniques has become as a novel research focus~\cite{Huang2022IGARSS, Kwak2022RS, Zhang2022TGRS, Liang2023GRSL, Zhao2024JAG}. The introduction of the large vision models (LVMs), such as Segment Anything Model (SAM)~\cite{SAM, Ravi2024Arxiv} has marked a revolutionized era in the realm of computer vision. Leveraging their exceptional generalization capabilities across diverse scenarios, the field of remote sensing stands to gain significantly from domain adaptation methods~\cite{WangetalRS, Chen2024TGRS, Lyu2025TGRS} informed by these advancements. Despite the fact that LVM-based domain adaptation methods in remote sensing remain relatively underexplored, it still reveals a significant and promising future research trend.
\par With the plethora of recent domain adaptation methods in remote sensing, some survey papers have been proposed. Notably,~\cite{Tuia2016GRSM, Peng2022JSATRs} delve into the domain adaptation methods specifically tailored for remote sensing classification tasks.~\cite{Xu2022RS} clarifies and reviews the idea of unsupervised domain adaptation in remote sensing area. Since the most recent pertinent survey only encompasses methods up to 2022, there arises a necessity to provide a more updated survey in this rapid-developed field. Compared to previous survey papers, we provide a updated overview of the most recent methods. For various specific remote sensing tasks, we review and present a detailed and comprehensive experimental comparison across diverse benchmark datasets. Considering that large vision models (LVMs) are anticipated to become a focal research topic in the forthcoming years, we will specifically explore LVM-based domain adaptation methods and outline the emerging trends in remote sensing area.
\par In summary, the main contributions of this survey can be listed as follows.
\begin{itemize}
\item We provide systematical overview of the latest research on domain adaptation methods in remote sensing. Compared to prior survey works, our study provides a broader scope and more updated content.
\item We present several systematic taxonomies of current domain adaptation methods in remote sensing, organized from four perspectives: task categorization, input model, supervision paradigms and algorithm granularity.
\item We highlight influential works with state-of-the-art performance on several key benchmarks, providing insightful guidance for ongoing and future research efforts.
\item We summarize the existing methods in this field and present our perspective on future trends and topics that worth further exploration.
\end{itemize}
\par As shown in Fig.\ref{Fig2}, this paper is organized as follows. In Sec.~\ref{Not}, we review the preliminary knowledge of domain adaptation methods in remote sensing including ``Preliminary Knowledge'' and ``Methodology Taxonomy \& Mathematical Notation''. In Sec.~\ref{method}, we conduct a thorough review of existing literature in a paper-to-paper manner and make categorization into four perspectives. In Sec.~\ref{performance}, we first introduce the benchmark datasets. Then we present a detailed performance comparison across diverse benchmark datasets on various remote sensing tasks. In Sec.~\ref{future}, we discuss the future research trends and topics as well as our viewpoints. Finally, we conclude this survey in Sec.~\ref{conclusion}.
%%%%%
\begin{figure*}
  \centering
\includegraphics[width=1.0\textwidth]{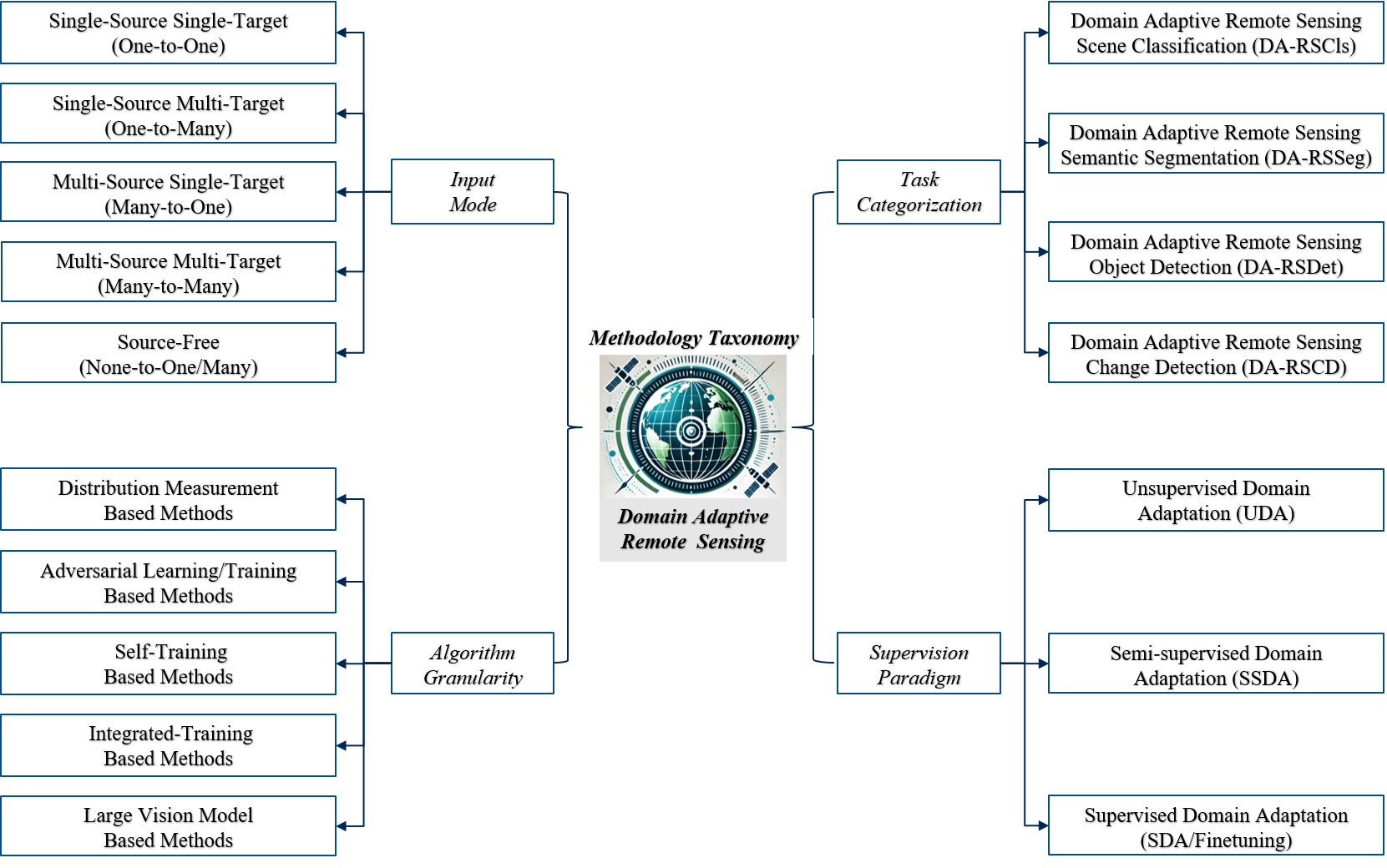}
  \caption{Overview on the method taxonomy on domain adaptation remote sensing tasks of this survey.}
  \label{Fig3}
\end{figure*}
%%%%%
\section{Preliminary Knowledge} % 2-3 pages
\label{Not}
\par As shown in Fig.~\ref{Fig2}, we will first introduce the preliminary knowledge including some basic notations. Subsequently, we organize and present methodology taxonomy as well as mathematical definitions across different methods in this field.
\subsection{Mathematical Notation}
\par As shown in Fig.~\ref{Fig2}, we provide a detailed taxonomies of existing methods, categorizing them based on four perspectives: task categorization, input mode, supervision paradigm and algorithm granularity. The specific methodology taxonomy is shown in Fig.~\ref{Fig3}.
\par Tab.~\ref{Tab1} shows the preliminary mathematical notations. For the four tasks focused in this survey (Fig.~\ref{Fig3}), existing methods always construct a model, optimize it using the training set ($\mathcal{D}_{train}$) and evaluate its performance on the testing set ($\mathcal{D}_{test}$). These training and testing datasets are partitioned from a comprehensive dataset ($\mathcal{D} = \mathcal{D}_{train} \cup \mathcal{D}_{test}$). Therefore, despite potential disparities in data distribution existing between the training and testing datasets, we still treat them as samples from the same domain. When domain adaptation is integrated into the previously mentioned four tasks, the problem definition will undergo a significant change.
\par For domain adaptation remote sensing (DA-RS) tasks, It always encounters source dataset ($\mathcal{D}_{S}$) and target dataset ($\mathcal{D}_{T}$). If the category sets of source and target dataset are respectively denoted as $\mathcal{C}_{S}, \mathcal{C}_{T}$, two sets have same number of categories ($N_{\mathcal{C}_{S}} = N_{\mathcal{C}_{T}}$) in general terms.
\begin{equation}
    \mathcal{D}_{S} = \{\mathcal{D}_{S}^{i}\}_{i=1}^{N_{S}}, \quad \mathcal{D}_{T} = \{\mathcal{D}_{T}^{i}\}_{i=1}^{N_{T}}
\label{Eq1}    
\end{equation}
%%%%%
\par The $i^{th}$ sub-set of source and target dataset respectively contain $K_{S}^{i}$ and $K_{T}^{i}$ samples, where $K_{S}^{i} \geq 0$ and $K_{T}^{i} \geq 0$. This process can be formulated in Eq.~\ref{Eq2}.
%%%%%
\begin{equation}
    \mathcal{D}_{S}^{i} = \{(\bm{x}_{S}^{i, j}, y_{S}^{i, j})\}_{j=1}^{K_{S}^{i}}, \quad \mathcal{D}_{T}^{i} = \{(\bm{x}_{T}^{i, j}, y_{T}^{i, j})\}_{j=1}^{K_{T}^{i}}
\label{Eq2}    
\end{equation}
%%%%%
\par To optimize a model for DA-RS tasks, $\mathcal{D}_{T}$ is divided into training and testing sub-sets ($\mathcal{D}_{T-train}, \mathcal{D}_{T-test}$), where $\mathcal{D}_{T} = \mathcal{D}_{T-train} \cup \mathcal{D}_{T-test}$. Consequently, for a given DA-RS task, the training dataset encompasses both the source dataset $\mathcal{D}_{S}$ and the training subset of the target dataset ($\mathcal{D}_{T-train}$), while the testing dataset $\mathcal{D}_{test}$ consists solely of $\mathcal{D}_{T-test}$. It means that both $\mathcal{D}_{S}$ and a portion of $\mathcal{D}_{T}$ will be are leveraged during the training phase. It is worthy of noting that the proportion of $\mathcal{D}_{T}$ in the training phase may be 0, implying that the target dataset will not be used for training purposes in certain cases.
%%%%%
\begin{table}
  \caption{Preliminary mathematical notations utilized for task definitions in this paper.}
  \centering
  \scalebox{0.83}{
  \begin{tabular}{c c | c c}
  \cmidrule(r){1-4}
  \multicolumn{2}{c}{Data-level Notations} & \multicolumn{2}{c}{Method-level Notations} \\
  \cmidrule(r){1-4}
  Notations & Descriptions & Notations &  Descriptions \\
  \cmidrule(r){1-4}
  $\mathcal{D}$ & Dataset & $\mathcal{L}(\cdot)$ & Loss \\ 
  \cmidrule(r){1-4}
  $\mathcal{D}_{train}, \mathcal{D}_{test}$ & Training, Testing dataset & $\theta$ & Model parameters \\
  \cmidrule(r){1-4}
  $\mathcal{D}_{S}, \mathcal{D}_{T}$ & Source, Target dataset & $f(\cdot), h(\cdot), g(\cdot)$ & Module functions \\
  \cmidrule(r){1-4}
  $\bm{x}, y$ & Input images, labels & $\alpha, \beta, \gamma, \cdots$ & Hyper-parameters \\ 
  \cmidrule(r){1-4}
  $\bm{X}, \bm{Y}$ & Input, Label set & $minmax$ & ``min-max'' criterion \\
  \cmidrule(r){1-4} 
  $\mathcal{C}$ & Category set & $argmax$ & ``argmax'' operation \\
  \cmidrule(r){1-4}
  $\mathcal{C}_{train}, \mathcal{C}_{test}$ & Training, Testing category set & $\mathbb{E}$ & Expectation \\
  \cmidrule(r){1-4}
  $\mathcal{C}_{S}, \mathcal{C}_{T}$ & Source/Target category set & $\mathbb{R}$ & Real numbers' set \\
  \cmidrule(r){1-4}
  $N_{S}, N_{T}$ & Source, Target sub-set number & $\bm{F}$ & Features \\
  \cmidrule(r){1-4}
  $K_{S}, K_{T}$ & Source, Target sample number & $\bm{P}$ & Predictions \\
  \cmidrule(r){1-4}
  $C, H, W$ & Image channel, height, width & B & Batch size \\
  \cmidrule(r){1-4}
  \end{tabular}}
 \label{Tab1}
\end{table}
\par As shown in Eq.~\ref{Eq1} and Eq.~\ref{Eq2}, the source dataset comprises $N_{S}$ sub-sets, whereas the target dataset includes $N_{T}$ sub-sets, with the constraints that $N_{S} \geq 1$ and $N_{T} \geq 1$. This indicates that both datasets contain at least one sub-set each.
%%%%%
\subsection{Methodology Taxonomy}
\subsubsection{Problem Definitions for DA-RS Tasks}
\par
\begin{figure*}
  \centering
  \includegraphics[width=1.0\textwidth]{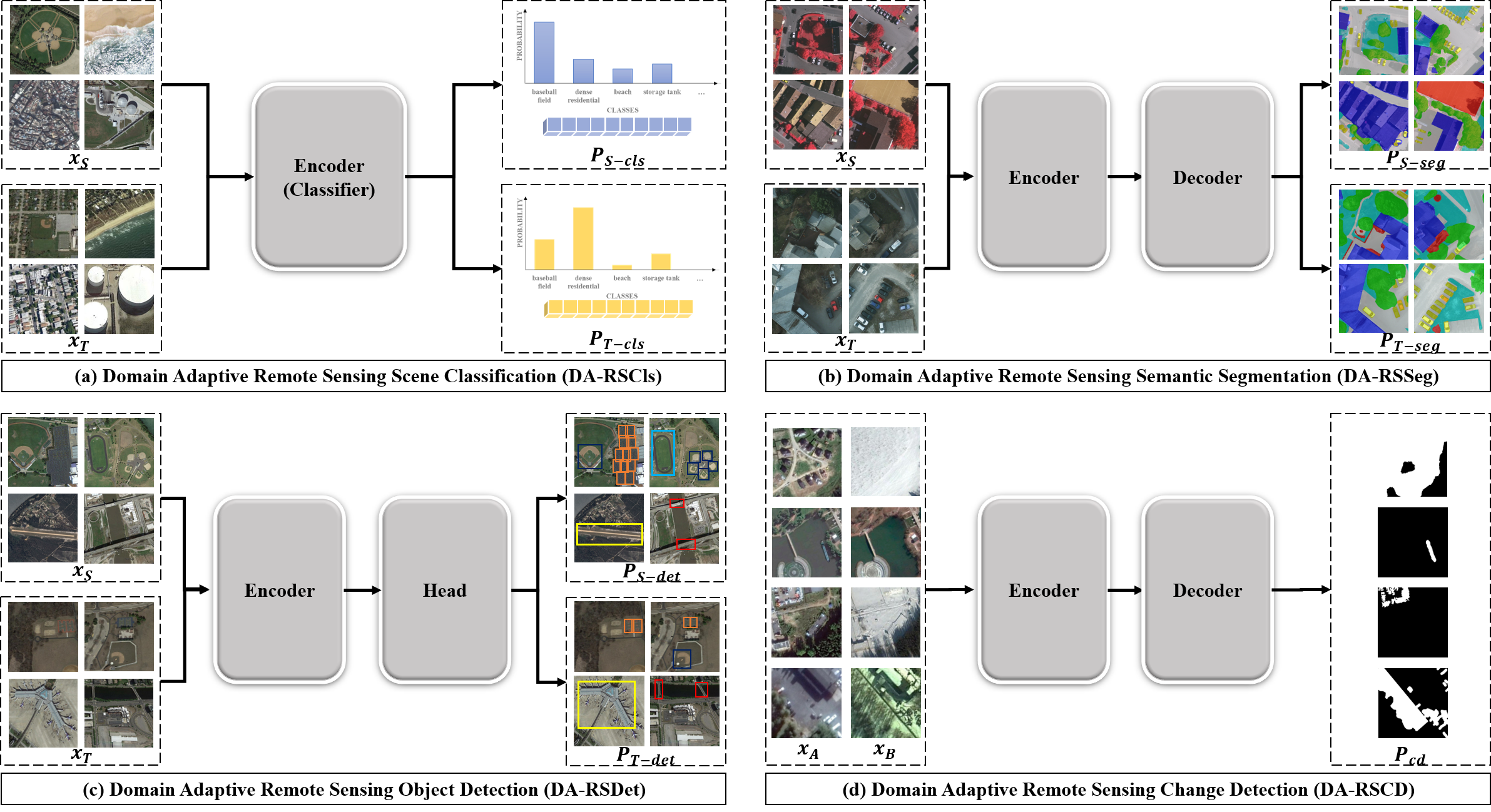}
  \caption{Overview on the four domain adaptation remote sensing tasks of this survey.}
  \label{Fig4}
\end{figure*}

 As shown in Fig.~\ref{Fig3}, this survey mainly review four DA-RS Tasks, which are DA-RSCls, DA-RSSeg, DA-RSDet, DA-RSCD. The specific illustration are shown in Fig.~\ref{Fig4}.
\par \textbf{DA-RSCls.} As shown in Fig.~\ref{Fig4} (a), the classifier comprises a solitary encoder. During forward pass, source and target images ($\bm{x_{S}}, \bm{x_{T}} \in \mathbb{R}^{B \times C \times H \times W}$) are fed into the encoder ($f_{cls}(\cdot)$) to generate predictions, formulated in Eq.~\ref{Eq3}. In classification task, the predictions ($\bm{P_{S-cls}}, \bm{P_{T-cls}} \in \mathbb{R}^{B \times N_{C}}$) are denoted as posterior probabilities across various categories. Here, $N_{C}$ indicates the number of categories.
%%%%%
\begin{equation}
    \bm{P_{S-cls}} = f_{cls}(\bm{x_{S}}), \quad \bm{P_{T-cls}} = f_{cls}(\bm{x_{T}})
\label{Eq3}    
\end{equation}
%%%%%
\par To optimize the encoder, the vanilla loss ($\mathcal{L}_{DA-cls}$) is shown in Eq.~\ref{Eq4}, where $\mathcal{L}_{ce}$ denotes cross-entropy loss. Based on this optimization paradigm, existing methods integrate various loss components to construct a combined loss including adversarial loss, self-training loss, etc. 
%%%%%
\begin{equation}
    \mathcal{L}_{DA-cls} = -\frac{1}{B} \sum_{i=1}^{B}\mathcal{L}_{ce}(\bm{P_{S-cls}^{i}}, \bm{P_{T-cls}^{i}}, y_{S}^{i}, y_{T}^{i})
\label{Eq4}    
\end{equation}
%%%%%
%%%%%
%%%%%
\par \textbf{DA-RSSeg.} As shown in Fig.~\ref{Fig4} (b), the segmentor employs an ``Encoder-Decoder'' architecture ($f_{seg}(\cdot)$). $\bm{x_{S}}, \bm{x_{T}}$ respectively pass through the segmentor to generate pixel-level predictions ($\bm{P_{S-seg}}, \bm{P_{T-seg}} \in \mathbb{R}^{B \times N_{C} \times H \times W}$), as formulated in Eq.~\ref{Eq5}. Each pixel in these predictions represents a posterior probability, thereby allowing semantic segmentation to be perceived as a pixel-wise classification task.
%%%%%
\begin{equation}
    \bm{P_{S-seg}} = f_{seg}(\bm{x_{S}}), \quad \bm{P_{T-seg}} = f_{seg}(\bm{x_{T}})
\label{Eq5}    
\end{equation}
%%%%%
\par Similar to the vanilla loss function adopted in DA-RSCls task (Eq.~\ref{Eq5}), cross-entropy loss is also utilized as a baseline loss ($\mathcal{L}_{DA-seg}$) to optimize the segmentor. This formulation is represented in Eq.~\ref{Eq6} and Eq.~\ref{Eq7}.

%%%%%
\begin{small}
    \begin{equation}
        \mathcal{L}_{DA-seg}^{i} = \frac{1}{HW}\sum_{h, w=1}^{H, W}\mathcal{L}_{ce}(\bm{P_{S-cls}^{i, h, w}}, \bm{P_{T-cls}^{i, h, w}}, y_{S}^{i, h, w}, y_{T}^{i, h, w})
\label{Eq6}    
    \end{equation}
\end{small}
%%%%%
%%%%%
\begin{equation}
    \mathcal{L}_{DA-seg} = -\frac{1}{B}\sum_{i=1}^{B}\mathcal{L}_{DA-seg}^{i}
    \label{Eq7}    
\end{equation}
%%%%%
\par \textbf{DA-RSDet.} Object detection poses a significant challenge, as it strives to simultaneously locate and classify objects within an image. By leveraging domain adaptation, the detector can achieve accurate detections even when faced with samples from different data distributions. As shown in Fig.~\ref{Fig4} (c), the basic detectors is combined with encoder and heads, which provides a general overview of both the ``one-stage detector'' and ``two-stage detector'' architectures.
\par Similar to DA-Cls and DA-Seg tasks, both source and target images will be delivered to the detector ($f_{det}(\cdot)$) and generate final predictions ($\bm{P_{S-det}}, \bm{P_{T-det}}$). Specifically in ``one-stage detector'', these predictions are structured as a single tensor that integrates both localization and classification outcomes. Conversely, in a ``two-stage detector'', the predictions are generated from distinct branches dedicated to localization and classification. Regardless of the variations among different types of detectors, both regression loss for bounding boxes ($\mathcal{L}_{reg}$) and classification loss ($\mathcal{L}_{cls}$) for posterior probabilities of instances are required. As a whole, the vanilla loss function ($\mathcal{L}_{DA-det}$) for DA-Det optimization can be formulated from Eq.~\ref{Eq8} to Eq.~\ref{Eq10}.
%%%%%
\begin{equation}
    \bm{P_{S-det}} = f_{det}(\bm{x_{S}}), \quad \bm{P_{T-det}} = f_{det}(\bm{x_{T}})
\label{Eq8}    
\end{equation}
%%%%%
%%%%%
\begin{equation}
\begin{split}
    \mathcal{L}_{DA-det}^{i} &= \mathcal{L}_{reg}(\bm{P_{S-det}^{i}}, \bm{P_{T-det}^{i}}, y_{S-box}^{i}, y_{T-box}^{i}) \\ &+ \mathcal{L}_{cls}(\bm{P_{S-det}^{i}}, \bm{P_{T-det}^{i}}, y_{S-cls}^{i}, y_{T-cls}^{i})
    \label{Eq9} 
\end{split}
\end{equation}
%%%%%
%%%%%
\begin{equation}
    \mathcal{L}_{DA-det} = -\frac{1}{B}\sum_{i=1}^{B}\mathcal{L}_{DA-det}^{i}
    \label{Eq10} 
\end{equation}
%%%%%
\par \textbf{DA-RSCD.} Change detection is an essential task in remote sensing that aims to detect and analyze changes occurring in the same geographical area over time. As shown in Fig.~\ref{Fig4} (d), the change detection model operates by accepting image pairs as input. $\bm{x_{A}}$ and $\bm{x_{B}}$ respectively denote the pre-change (pre-event) and post-change (post-event) images. In RSCD task, $\bm{x_{A}}$ and $\bm{x_{B}}$ show minimal discrepancy, indicating less degree of domain-shift. In DA-RSCD task, $\bm{x_{A}}$ and $\bm{x_{B}}$ exhibit a significant domain-shift problem, which is often caused by the use of different sensors or varying imaging conditions. 
\par The architecture of change detector ($f_{cd}(\cdot)$) is similar to segmentor. $\bm{x_{A}}$ and $\bm{x_{B}}$ are fed into an ``Encoder-Decoder'' structure to generate a binary prediction ($\bm{P_{cd}} \in \mathbb{R}^{B \times N_{C} \times H \times W}$), where $N_{C} = 2$. This process is expressed in Eq.~\ref{Eq11}.
%%%%%
\begin{equation}
    \bm{P_{cd}} = f_{cd}(\bm{x_{A}}, \bm{x_{B}})
\label{Eq11}    
\end{equation}
%%%%%
\par The optimization of DA-RSCD can be considered as a binary segmentation optimization problem. The specific formulation of the vanilla loss function ($\mathcal{L}_{DA-cd}$) is detailed in Eq.~\ref{Eq12}, where $\mathcal{L}_{bce}$ denotes the binary cross-entropy loss.
%%%%%
\begin{equation}
    \mathcal{L}_{DA-cd} = -\frac{1}{BHW}\sum_{i=1}^{B}\sum_{h, w=1}^{H, W}\mathcal{L}_{bce}(\bm{P_{cd}^{i, h, w}}, y^{i, h, w})
    \label{Eq12}    
\end{equation}
%%%%%
\par The DA-RSCD task distinguishes itself from the aforementioned three DA-RS tasks in that the domain-shift issue arises within image pairs. Conversely, the domain-shift challenges faced by DA-RSCls, DS-RSSeg, and DA-RSDet manifest between the source and target datasets.
%%%%%
\begin{table}
  \caption{Mathematical definitions for different input modes. The formulations in this table can be referred to Eq.~\ref{Eq1}.}
  \centering
  \scalebox{0.84}{
  \begin{tabular}{c c c}
  \cmidrule(r){1-3}
  Input modes & Source Input: $\mathcal{D}_{S} = \{\mathcal{D}_{S}^{i}\}_{i=1}^{N_{S}}$ & Target Input: $\mathcal{D}_{T} = \{\mathcal{D}_{T}^{i}\}_{i=1}^{N_{T}}$ \\ \cmidrule(r){1-3}
  One-to-one & $N_{S} = 1$ & $N_{T} = 1$ \\ \cmidrule(r){1-3}
  One-to-Many & $N_{S} = 1$ & $N_{T} \textgreater 1$ 
  \\ \cmidrule(r){1-3}
  One-to-Many & $N_{S} \textgreater 1$ & $N_{T} = 1$ 
  \\ \cmidrule(r){1-3}
  Many-to-Many & $N_{S} \textgreater 1$ & $N_{T} \textgreater 1$ 
  \\ \cmidrule(r){1-3}
  None-to-One/Many & Not Available & $N_{T} \geq 1$ 
  \\ \cmidrule(r){1-3}
  \end{tabular}}
 \label{Tab2}
\end{table}
%%%%%
\subsubsection{Mathematical Definitions for Different Input Modes}
\par In the taxonomy of input modes, there are primarily five distinct types of DA-RS methods, categorized with respective of the number of source and target sub-sets (Fig.~\ref{Fig3}). The specific definitions are shown in Tab.~\ref{Tab2}.
\par \textbf{Single-Source Single-Target (``One-to-One'').} When $N_{S}$ and $N_{T}$ are set as $1$, this corresponds to a standard domain adaptation configuration on distinct tasks. In this mode, model are trained to adapt from the source-domain to target-domain.
\par \textbf{Single-Source Multi-Target (``One-to-Many'').} When $N_{S}$ is set as $1$ while $N_{T}$ is set to a value greater than $1$, the primary objective is to develop models that can perform well across different target domains. One typical domain adaptation task in this mode is known as domain generalization (DG). 
\par \textbf{Multi-Source Single-Target (``Many-to-One'').} When $N_{S}$ is set to a value greater than $1$ while $N_{T}$ is set as $1$, the goal is to develop a model that can leverage the collective knowledge from multiple source domains to enhance the model's generalization capabilities on an unseen target domain. Another crucial challenge in this scenario is to mitigate the adverse effects of domain-shift among various source sub-sets.
\par \textbf{Multi-Source Multi-Target (``Many-to-Many'').} When $N_{S}$ and $N_{T}$ are both set to a value greater than $1$, the task will become further complex. Models are required to effectively optimized for generalizing from multiple source domains to multiple target domains.
\par \textbf{Source-Free Single-Target / Multi-Target (``None-to-One/Many'').} In this mode, $N_{T}$ is set to a value greater than $1$ while $N_{S}$ is not available in adaptation phase due to data privacy or transmission difficulty. In the aforementioned four modes, the annotated source data is all accessible. In source-free domain adaptation, only the pretrained model on the source data is available. 
\subsubsection{Mathematical Definitions for Different Supervision Paradigms}
\par As shown in Fig.~\ref{Fig3}, there are mainly three supervision paradigms in the DA-RS methods. 
\par \textbf{Unsupervised domain adaptation in remote sensing (``UDA-RS'').} UDA is a notable and widely-applied paradigm in realm of remote sensing. In general terms, UDA indicates that the labels of target images are unavailable, yet these target images can still be incorporated into the training phase. Mathematically, in Eq.~\ref{Eq2}, each target sample ($y_{T}^{i, j}$) is denoted as ``\textit{None}'', formulated in Eq.~\ref{Eq13}. Under this setting, $\bm{y_{T}}$ is denoted as an empty set in loss functions for each task shown in Eq.~\ref{Eq4}, Eq.~\ref{Eq6} and Eq.~\ref{Eq9}.
%%%%%
\begin{equation}
    \bm{y_{T}} = \emptyset, \quad \forall i \in [1, N_{T}], j \in [1, K_{T}^{i}], y_{T}^{i, j} \rightarrow \textit{None}
\label{Eq13}    
\end{equation}
%%%%%
\par As narrowly defined, UDA-RS indicates a typical single-source single-target RS task, where $N_{S} = 1, N_{T} = 1$. Obviously, the methods utilized within this paradigm are intended to address the limitation posed by the insufficient availability of labeled target domain samples, as well as to address the issue of high annotation costs.
\par \textbf{Semi-supervised domain adaptation in remote sensing (``SSDA-RS'').} SSDA is another hot topic in remote sensing field, which aims to improve the generalization performance by labeling a few target domain samples. As shown in Eq.~\ref{Eq2}, each sub-set of target dataset ($D_{T}^{i}$) contains labeled and unlabeled samples, where the number of unlabeled samples ($K_{T-l}^{i}$) is much larger the the number of labeled samples ($K_{T-u}^{i}$). This paradigm can be formulated in Eq.~\ref{Eq14}.
%%%%%
\begin{equation}
    \mathcal{D}_{T}^{i} = \{(\bm{x}_{T}^{i, l}, y_{T}^{i, l})\}_{l=1}^{K_{T-l}^{i}} \cup \{(\bm{x}_{T}^{i, u})\}_{u=1}^{K_{T-u}^{i}}
\label{Eq14}    
\end{equation}
%%%%%
where $K_{T-l}^{i} + K_{T-u}^{i} = K_{T}^{i}$ and $K_{T-l}^{i} \ll K_{T-u}^{i}$. Similar to UDA-RS, existing methods concerning SSDA-RS primarily utilize the single-source single-target input mode as well. Compared to UDA-RS, SSDA-RS emphasizes exploring information from limited labeled samples while maximizing the utilization of unlabeled samples.
\par \textbf{Supervised domain adaptation in remote sensing (``SDA/Finetuning-RS'').} SDA assumes that labeled samples are available for both domains. SDA methods focus on challenging situations where labeled target-domain samples are less numerous than those available in the source domain~\cite{Tuia2016GRSM}. In such conditions, the proper use of source domain information and the finetuning technique on target domain are very important in solving the target problem. Mathematically, this paradigm can be represented as $\sum_{i=1}^{N_{T}}K_{T}^{i} \ll \sum_{i=1}^{N_{S}}K_{S}^{i}$. 
\par In remote sensing scene, acquiring data from various sensors such as satellites and drones is relatively straightforward, resulting in an abundant supply of data from diverse domains. However, due to the vast geographical land cover, the challenge of high annotation costs becomes even more pronounced. Consequently, the paradigms of UDA-RS and SSDA-RS are more aligned with practical requirements.
\subsubsection{Mathematical Definitions for Typical Algorithms} 
\par As shown in Fig.~\ref{Fig3}, DA-RS methods primarily consist of five typical types of algorithms.
\begin{figure}[!t]
  \centering
  \includegraphics[width=0.45\textwidth]{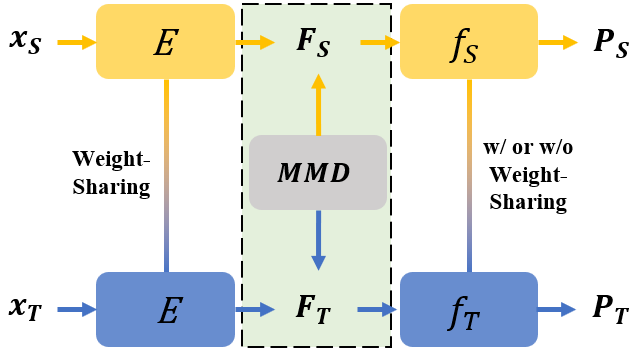}
  \caption{Illustration of the common-utilized distribution measurement paradigm. $E$ and $\{f_{S}, f_{T}\}$ respectively denote encoders and feature mapping modules.}
  \label{Fig5}
\end{figure}
\par \textbf{Distribution Measurement Based Methods.} Distribution measurement based methods mainly aim to match distributions between hidden features from different domains. As shown in Fig.~\ref{Fig5} and Eq.~\ref{Eq15}, the common operation is to embed adaptation metric (e.g., Maximum Mean Discrepancy, MMD) into the neural networks.

%%%%%
 \begin{small}
\begin{equation}
 \mathcal{MMD}(\mathcal{D}_{S}, \mathcal{D}_{T}) = ||\frac{1}{N_{S}}\sum_{i=1}^{N_{S}}E(\bm{x_{S}^{i}}) - \frac{1}{N_{T}}\sum_{j=1}^{N_{T}}E(\bm{x_{T}^{j}})||_{2}^{2}
\label{Eq15}
\end{equation}
\end{small}
\par DAN~\cite{Long2015ICML} is  the pioneer in leveraging the deep convolutional neural networks to learn transferable hidden features across domains in UDA task. It embeds MMD metric for measurement. Subsequent to DAN, numerous DA-RS tasks adopt MMD for distribution measurement across intermediate feature maps of images belonging to different domains.  
\par
\begin{figure*}
  \centering
  \includegraphics[width=0.98\textwidth]{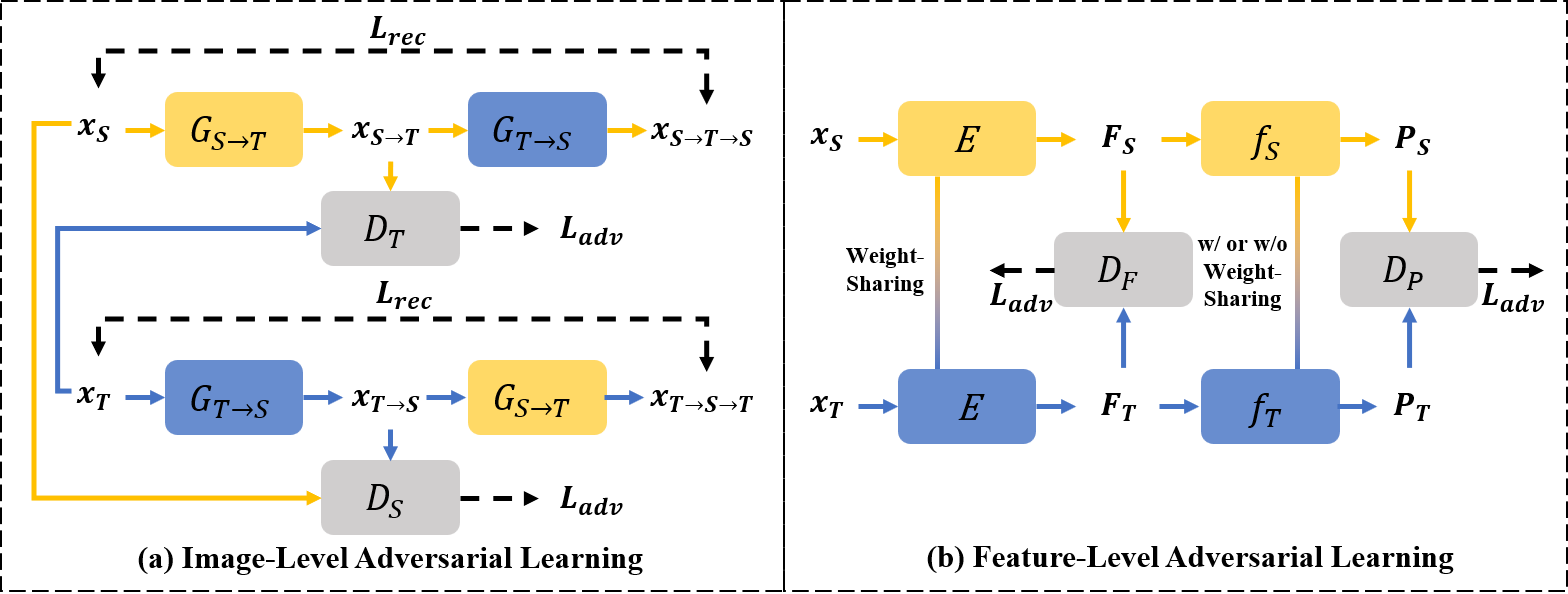}
  \caption{Illustration of the common-utilized adversarial learning paradigms. $\{G_{S \rightarrow T}, G_{T \rightarrow S}\}$ and $\{D_{S}, D_{T}, D_{F}, D_{P}\}$ respectively denote generators and discriminators. $E$ and $\{f_{S}, f_{T}\}$ respectively denote encoders and feature mapping modules.}
  \label{Fig6}
\end{figure*}

 \textbf{Adversarial Learning Based Methods.} Adversarial learning is frequently utilized in DA-RS methods, primarily encompassing two types: image-level and nd feature-level adversarial learning-based methods. The illustration of these two typical adversarial learning paradigms is shown in Fig.~\ref{Fig6}.
%%%%%
\par Image-level adversarial learning-based methods utilize image generalization techniques, such as image translation~\cite{cyclegan, DA-pix2pix}, to synchronize the data distribution between source and target images. These methods initially apply image-level adaptation and subsequently train models using cross-domain synthetic data. As depicted in Fig.~\ref{Fig1} (a), $G_{S \rightarrow T}$ and $G_{T \rightarrow S}$ are responsible for adapting images from the source domain to the target domain, and vice versa. Meanwhile, the role of $D_{S}$ and $D_{T}$ is to distinguish between authentic source data and the fake data produced by the generators. For optimization, adversarial loss is employed, which is formulated in Eq.~\ref{Eq16}.

%%%%%
\begin{small}
    \begin{equation}
\begin{cases}{}
\begin{split}
\mathcal{L}_{adv}(G_{S \rightarrow T}, D_{T}) &= 
    \mathbb{E}_{x_{T} \sim X_{T}}[log(D_{T}(\bm{x_{T}}))]  \\
    &+ \mathbb{E}_{x_{S} \sim X_{S}}[log(1 - D_{T}(G_{S \rightarrow T}(\bm{x_{S}})))]
\end{split}
\\ \\ 
\begin{split}
\mathcal{L}_{adv}(G_{T \rightarrow S}, D_{S}) &= 
    \mathbb{E}_{x_{S} \sim X_{S}}[log(D_{S}(\bm{x_{S}}))]  \\
    &+ \mathbb{E}_{x_{T} \sim X_{T}}[log(1 - D_{S}(G_{T \rightarrow S}(\bm{x_{T}})))]
\end{split}
\end{cases}
\label{Eq16}
\end{equation}
\end{small}
%%%%%
\par To optimize $G_{S \rightarrow T}$ and $G_{T \rightarrow S}$. ``Min-max'' criterion is adopted, which can be expressed in Eq.~\ref{Eq17}.
%%%%%
\begin{equation}
\begin{cases}{}
    min_{G_{S \rightarrow T}}max_{D_{T}}\mathcal{L}_{adv}(G_{S \rightarrow T}, D_{T}) \\ \\ min_{G_{T \rightarrow S}}max_{D_{S}}\mathcal{L}_{adv}(G_{T \rightarrow S}, D_{S})
\end{cases}
\label{Eq17}
\end{equation}
%%%%%
\par In this paradigm, reconstruction loss ($\mathcal{L}_{rec}$) or cycle consistency loss is consistently utilized alongside the adversarial loss. The formulation for this is presented in Eq.~\ref{Eq18}.

%%%%%
\begin{small}
    \begin{equation}
\begin{split}
\mathcal{L}_{rec}(G_{S \rightarrow T}, G_{T \rightarrow S}) &= 
    \mathbb{E}_{x_{T} \sim X_{T}}[||G_{S \rightarrow T}G_{T \rightarrow S}(\bm{x_{T}}) - \bm{x_{T}}||_{a}]  \\
    &+ \mathbb{E}_{x_{S} \sim X_{S}}[||G_{T \rightarrow S}G_{S \rightarrow T}(\bm{x_{S}}) - \bm{x_{S}}||_{a}]
\end{split}
\label{Eq18}
\end{equation}
\end{small}
%%%%%
\par \noindent where $||\cdot||_{a}$ indicates $L_{a}-norm$, which is served as loss function. Among the commonly applied norms, The $L_{1}-norm$ and $L_{2}-norm$ are particularly prevalent. This loss function aims to encourage the preservation of structural properties during the style transfer process.
\par Feature-level adversarial learning-based methods delve into the domain-invariant features shared between source-style and target-style features. These methods leverage feature-level alignment strategies to address the domain shift problem. Typically, discriminators are integrated into the networks to ensure consistency alignment on intermediate feature maps or output posterior probabilities. 
\par As shown in Fig.~\ref{Fig6} (b), $E$ are used for feature extraction, which share their weights in most scenarios. $f_{S}$ and $f_{T}$ are utilized to transform the extracted features into predictions. The design of these modules, as well as whether they share weights or not, may vary depending on the specific DA-RS tasks. To optimize the framework, feature-level adversarial loss can be depicted in Eq.~\ref{Eq19} and Eq.~\ref{Eq20}.

%%%%%
\begin{small}
    \begin{equation}
\begin{split}
\mathcal{L}_{adv}(E, D_{F}) &= 
    \mathbb{E}_{x_{S} \sim X_{S}}[log(D_{F}(E(\bm{x_{S}})))]  \\
    &+ \mathbb{E}_{x_{T} \sim X_{T}}[log(1 - D_{F}(E(\bm{x_{T}})))]
\end{split}
\label{Eq19}
\end{equation}
\end{small}
%%%%%
%%%%%
\begin{small}
    \begin{equation}
\begin{split}
\mathcal{L}_{adv}(E, f_{S}, f_{T}, D_{P}) &= \mathbb{E}_{x_{S} \sim X_{S}}[log(D_{P}(f_{S}(E(\bm{x_{S}}))))]  \\
    &+ \mathbb{E}_{x_{T} \sim X_{T}}[log(1 - D_{P}(f_{T}(E(\bm{x_{T}}))))]
\end{split}
\label{Eq20}
\end{equation}
\end{small}
%%%%%
\par Similar to the optimization process in image-level adversarial learning (Eq.~\ref{Eq17}), the ``min-max'' criterion is also utilized for optimizing the loss function, shown in Eq.~\ref{Eq21}.
%%%%%
\begin{equation}
\begin{cases}{}
    min_{E}max_{D_{F}}\mathcal{L}_{adv}(E, D_{F}) \\ \\ min_{\{E, f_{S}, f_{T}\}}max_{D_{P}}\mathcal{L}_{adv}(E, f_{S}, f_{T}, D_{P})
\end{cases}
\label{Eq21}
\end{equation}
\par \textbf{Self-Training 
Based Methods.} As a non-adversarial UDA paradigm, self-training has attracted much attention in DA-RS tasks. This algorithm promotes the adaption ability by generating reliable, consistent, and class-balanced pseudo labels.  By supervising the model with high-quality pseudo ground-truth derived from the target domain, models can quickly adapt to target images. 
\par Self-training mechanism mainly contains two processes. The first process involves the updating stage, where the exponential moving average (EMA) technique is commonly applied to the original network, referred to as the student network. Once the EMA-updating process is completed, the resulting network functions as the EMA-teacher. Specifically, during each training step $t$, the EMA-teacher's weights are updated by student's weights by EMA operation, which can be formulated in Eq.~\ref{Eq22}.
%%%%%
\begin{equation}
    \phi_{t} = \alpha\phi_{t-1} + (1-\alpha)\theta_{t}
\label{Eq22}
\end{equation}
%%%%%
where $\alpha$ represents the EMA decay factor, which controls the rate of updating. $\phi_{t}$ denotes the weights of the EMA-teacher at the $t^{th}$ step. $\theta_{t}$ signifies the weights of the student network at the same $t^{th}$ step.
\par After updating the weights of the teacher networks, the second process involves generating pseudo-labels for target images and utilizing these pseudo-labels for supervision. For specific DA-RS tasks, input images are fed into the teacher networks for pseudo-labels generation process, which can be referenced in Eq.~\ref{Eq3},~\ref{Eq5},~\ref{Eq8},~\ref{Eq11}. Subsequently, these pseudo-labels are treated as ground-truths of target samples to optimize and enhance the performance of the student network.
\par \textbf{Integrated-Training 
Based Methods.} To achieve improved model performance in addressing the domain shift problem, numerous domain adaptation methods employ an integrated-training strategy, which is also referred to a hybrid-training strategy. Alongside basic task-specific loss functions for optimization (referenced in Eq.~\ref{Eq4},~\ref{Eq7},~\ref{Eq10},~\ref{Eq12}), adversarial learning plays a crucial role in achieving consistency alignment, while the self-training mechanism primarily aims to mitigate the representation bias of source-trained networks. Therefore, these two training strategies are compatible, and integrating them with a combined loss function can enhance the model from different perspectives.
\par \textbf{Large Vision Model 
Based Methods.} In the field of DA-RS, the domain shift issue is particularly severe due to various factors (Fig.~\ref{Fig1}). Essentially, a single model often fails to perform well across different domains due to its limited generalization ability. Large vision models~\cite{Lu2025GRSM}, can effectively mitigate this issue. Their generalized representation capabilities arise from extensive pre-training on large amount of samples, covering a wide range of domains. 
\par Recently, LVM-based methods on DA-RS tasks remain relatively underexplored. The challenge lies in how to effectively transfer the generalized capabilities of LVMs to a specific downstream DA-RS task, as well as how to design an optimization paradigm tailored for LVM-based approaches.
\begin{table*}[!htbp]
  \scriptsize
  \centering
  \caption{Representative methods with different supervision paradigms in different DA-RS tasks.}
  \scalebox{1.0}{
  %\begin{tabular}{c|cccc}
  \begin{tabular}{p{0.06\textwidth}|p{0.055\textwidth}|p{0.26\textwidth} |p{0.27\textwidth}|p{0.25\textwidth}}
  \hline \hline
  {} & \makecell[c]{\multirow{2}{*}{\textbf{Methods}}} & \multicolumn{3}{c}{\textbf{Taxonomy Based on Task Categorization}}
  \\ \cline{3-5}
  {} & {} & \makecell[c]{DA-RSCls} & \makecell[c]{DA-RSSeg} & \makecell[c]{DA-RSDet}
  \\ \cline{1-5}
  \makecell[c]{\multirow{3}{*}{\shortstack{\textbf{Taxonomy} \\ 
  \textbf{Based on} \\
  \textbf{Supervision} \\ \textbf{Paradigm}}}} & \makecell[c]{UDA-RS} & \makecell[c]{
  \\ DAN~\cite{Othman2017TGRS}, DDME~\cite{Wang2019GRSL}, 
  DACNN~\cite{Song2019GRSL}, \\ TCANet~\cite{Garea2019RS},  DDA-Net~\cite{Ma2019TGRS}, SSMT-RS~\cite{Zheng2020IGARSS}, \\ ADA-Net~\cite{Ma2020TGRS}, MSCN~\cite{Lu2020TGRS}, ECB-FAM~\cite{Ma2021RS}, \\ CDA~\cite{Liu2021TGRS},  JCGNN~\cite{Wang2021JSTARs},   JDA~\cite{Miao2021JSTARs}, \\
  CMC~\cite{Wei2021JSTARs},   AST~\cite{Huang2022IGARSS}, 
  ADA-DDA~\cite{Yang2022TGRS}, \\ DFENet~\cite{ZhangXufei2022TGRS}, UDACA~\cite{Yu2022GRSL}, GNN-MTDA~\cite{Saha2022GRSL},  \\ TAADA~\cite{Huang2022TGRS}, 
  PFDA~\cite{Chen2022GRSL},  EHACA~\cite{Ngo2023TGRS}, \\ SDG-MA~\cite{Xu2023TGRS},   PDA~\cite{Zheng2023TGRS},   SSWADA~\cite{Huang2023ISPRS}, \\ 
  VSFA~\cite{Zhang2023TGRS},   TSTnet~\cite{Zhang2023TNNLS}, UDA-SAR~\cite{Shi2024TGRS}, \\ FDDAN~\cite{Xin2024JAG}, PPLM-Net~\cite{Leng2024JSTARs},   AdaIN~\cite{Zhang2024GRSL}, \\ DST~\cite{Zhao2024GRSL},   HFPAN~\cite{Miao2024TGRS},  MLUDA~\cite{Cai2024TGRS}, \\  C$^3$DA~\cite{Guo2024GRSL}, S$^4$DL~\cite{Feng2025TNNLS}, S$^{2}$AMSnet~\cite{ChenXi2024TGRS}, \\ SSM~\cite{Liu2021JSTARs}, SDEnet~\cite{Zhang2023TIP}, DAN\_MFAC~\cite{Wu2025KBS}, \\ DATSNET~\cite{ZhengZhendong2021TGRS}, MRDAN~\cite{Niu2022TGRS}, \\ AMRAN~\cite{Zhu2022TGRS}, SRKT~\cite{Zhao2023TGRS},  DDCI~\cite{Zhao2025TGRS}  \\ \\} & \makecell[c]{
  \\ UDA-GAN~\cite{Benjdira2019RS}, Tri-ADA~\cite{Yan2021TGRS}, DNT~\cite{Chen2022JAG}, \\ DCA~\cite{Wu2022TGRS}, ResiDualGAN~\cite{Zhao2023RS}, BiFDANet~\cite{Cai2022RS}, \\ MBATA-GAN~\cite{Ma2023TGRS}, FGUDA~\cite{Wang2023JSTARs}, JDAF~\cite{Huang2024TGRS}, \\ CSLG~\cite{ZhangBo2022TGRS}, MemoryAdaptNet~\cite{Zhu2023TGRS}, \\ De-GLGAN~\cite{Ma2024TGRS}, RoadDA~\cite{Zhang2022TGRS}, RCA-DD~\cite{Chen2022TGRS}, \\ PFM-JONet~\cite{Lyu2025TGRS}, STADA~\cite{Liang2023GRSL}, MIDANet~\cite{ChenHongyu2022TGRS}, \\ ST-DASegNet~\cite{Zhao2024JAG}, CPCA~\cite{Zhu2024TGRS}, MEBS~\cite{Li2025TGRS}, \\
  MS-CADA~\cite{Gao2024TGRS}, CDANet~\cite{Yang2024NN}   MHDA~\cite{Liang2023TGRS}, \\ IterDANet~\cite{Cai2022TGRS}, MMDANet~\cite{Zhou2023TGRS},  \\ DDF~\cite{Ran2024TGRS}, TDAIF~\cite{GaoXianjun2024TGRS},  HighDAN~\cite{Hong2023RSE}, \\ GeoMultiTaskNet~\cite{Marsocci2023CVPRW}, EUDA-PLR~\cite{Cui2025JAG} \\ \\} & \makecell[c]{FRCNN-SAR~\cite{Shi2021JSTARs}, HSANet~\cite{ZhangJun2022TGRS}, \\ PT-SAR~\cite{Shi2022TGRS},  IDA~\cite{Pan2023TGRS}, DFD-CAC~\cite{Zhang2025TGRS}, \\ FACL~\cite{Zhao2022TGRS},   PDCSR~\cite{Luo2023GRSL}, RIRA~\cite{Chen2021RS}, \\ FADA~\cite{Xu2022TGRS}, RFA-Net~\cite{Zhu2022JSTARs}, APA~\cite{Koga2021IGARSS}, \\ DualDA-Net~\cite{ZhuYangguang2023TGRS}, CORAL-ADA~\cite{Koga2020RS}, \\   DCLDA~\cite{Biswas2024JSTARs}, ML-UDA~\cite{Luo2024JSTARs},  CDST~\cite{Luo2024TGRS}, \\ MGDAT~\cite{Fang2025RS}, RST~\cite{Han2024TGRS}, SFOD~\cite{Liu2024Arxiv}, \\ FIE-Net~\cite{ZhangJun2025TGRS}, HDADE~\cite{ZhangGuangbin2025TGRS}, RSL-DA~\cite{Jiao2025JSTARs} }
  \\ \cline{2-5}
  {} & \makecell[c]{SSDA-RS} & \makecell[c]{\\ SS-MA~\cite{Tuia2014TGRS}, ADDA~\cite{Wang2018IGARSS}, TDDA~\cite{Li2020RS}, \\ CDADA~\cite{Teng2020GRSL}, DACNN-MME~\cite{Lasloum2021NISS}, \\\ SSDAN~\cite{Lasloum2021RS}, DJ-CORAL~\cite{Yang2022GRSL}, \\ AL-LCC~\cite{Kalita2022GRSL}, BSCA~\cite{HuangWei2023ISPRS}, AG-GTL~\cite{Yaghmour2024JSTARs}, \\ CFAN~\cite{Li2024TGRS}, SSCA~\cite{Mo2024RS}, SASS~\cite{Qu2024TGRS} \\ \\ }& \makecell[c]{Fusion-DA~\cite{Hafner2022RSE}, CDMPC~\cite{Gao2023RS}, \\ TempCNN~\cite{Lucas2023ML}, EasySeg~\cite{Yang2024TGRS} } & \makecell[c]{DA-FRCNN~\cite{Guo2021RS}, FDDA~\cite{Liao2022RS}, \\ SAR-CDSS~\cite{Luo2024RS}, SRA-YOLO~\cite{Huang2024ICANN}, DT~\cite{ZhengXiangtao2023TGRS}, \\ WeedTeacher~\cite{Deng2025Arxiv}, CDTL-YOLOV5~\cite{Liao2024IGARSS} \\}
  \\ \cline{1-5} \hline \hline
   \end{tabular}
  }
  \label{Tab3}
\end{table*}
  \section{Methodology: A Survey} % 3 pages
  \label{method}
   In this section, we will conduct a comprehensive review of methodologies from the taxonomy based on supervision paradigm and the taxonomy based on algorithm granularity. First, we will include different methods that are pertinent to different task categories according to the taxonomy based on supervision paradigm. Following this, we will delve into a detailed introduction of these methods, structured in accordance with the taxonomy based on algorithm granularity. Given that the majority of existing DA-RS methods use ``Single-Source Single Target'' as input mode, we will also delve into methods that employ other unique input modes. It is important to acknowledge that there may be inevitable overlaps among the methods within these three taxonomies. Specifically, some methods may fall into the realm of more than one taxonomy.
  \subsection{Taxonomy Based on Supervision Paradigm}
  \subsubsection{UDA-RS Methods} In remote sensing field, the high cost of annotation raises a significant challenge for remote sensing (RS) tasks. Consequently, researching methods to reduce the reliance on a vast number of annotated samples has emerged as a popular topic. As previously mentioned, UDA-RS strives to train a model using source images while having access to target images but not their corresponding annotations. The representative UDA-RS methods are included in Tab.~\ref{Tab3}.
  \par Before the deep learning era, pioneering traditional methods define the UDA-RS task and introduce a variety of machine learning models~\cite{Sun2013GRSL, Tuia2013TGRS, Banerjee2015TGRS, Othman2016GRSL, Yang2016JSTARs, Peng2019GRSL, MaLi2019TGRS}, which significantly advances the development within this field. Upon entering the deep learning era, deep learning techniques have shown superior performance over traditional methods.
  \par \textbf{UDA-RSCls.} Unsupervised domain adaptation Hyperspectral image (HSI) classification stands as a pivotal topic within the UDA-RSCls task. As shown in Tab.~\ref{Tab3}, many methods~\cite{Garea2019RS, Ma2019TGRS, Ma2020TGRS, Liu2021TGRS, Wang2021JSTARs, Miao2021JSTARs, Wei2021JSTARs, Liu2021JSTARs, Yu2022GRSL, Huang2022TGRS, Huang2023ISPRS, Zhang2023TNNLS, Xin2024JAG, Cai2024TGRS, Wu2025KBS, Feng2025TNNLS} with variety of algorithms have been proposed to address the challenge of UDA HSI classification. Unsupervised domain adaptation for aerial image classification represents another fascinating research topic within the UDA-RSCls task. Many notable methods~\cite{Othman2017TGRS, Wang2019GRSL, Song2019GRSL, Zheng2020IGARSS, Lu2020TGRS, Ma2021RS, Huang2022IGARSS, Yang2022TGRS, ZhangXufei2022TGRS, Ngo2023TGRS, Xu2023TGRS, Zheng2023TGRS, Leng2024JSTARs, Miao2024TGRS, Guo2024GRSL, ZhengZhendong2021TGRS, Zhu2022TGRS, Zhao2023TGRS, Niu2022TGRS, Zhao2025TGRS} are proposed to exploit the abundant information contained in high-resolution aerial and satellite images. Given the unique imaging characteristics and inherent domain shifts of SAR data, unsupervised domain adaptation for SAR image classification emerges as another compelling direction within the broader scope of UDA-RSCls. Several recent studies~\cite{Chen2022GRSL, Zhang2023TGRS, Shi2024TGRS, Zhang2024GRSL, Zhao2024GRSL} have explored strategies to mitigate the significant domain gap in different SAR data, aiming to enhance cross-domain generalization.
  \par \textbf{UDA-RSSeg.} Unsupervised domain adaptation for remote sensing semantic segmentation focuses primarily on adapting models across high-resolution (HR) aerial and satellite optical images~\cite{Benjdira2019RS, Yan2021TGRS, Chen2022JAG, Wu2022TGRS, Cai2022RS, Zhao2023RS, Liang2023GRSL, Ma2023TGRS, Wang2023JSTARs, Huang2024TGRS, Zhu2023TGRS, Ma2024TGRS, Zhao2024JAG, Zhang2022TGRS, ZhangBo2022TGRS, Zhu2024TGRS, Yang2024NN, Liang2023TGRS, Chen2022TGRS, Cai2022TGRS, Gao2024TGRS, ChenHongyu2022TGRS, Zhou2023TGRS, Ran2024TGRS, Li2025TGRS}, addressing domain shifts caused by geographic or acquisition differences. Some methods~\cite{Hong2023RSE, Marsocci2023CVPRW, Cui2025JAG, GaoXianjun2024TGRS} also aim to bridge modality gaps for cross-domain segmentation on multispectral or SAR imagery, addressing challenges such as variations in texture, resolution, and spectral characteristics.
  \par \textbf{UDA-RSDet.} Unsupervised domain adaptation for remote sensing object detection primarily focuses on transferring knowledge from labeled optical aerial and satellite images to unlabeled imagery from different domains~\cite{Koga2020RS, Chen2021RS, Koga2021IGARSS, Xu2022TGRS, Zhu2022JSTARs, ZhuYangguang2023TGRS, Biswas2024JSTARs, Luo2024JSTARs, Luo2024TGRS, Han2024TGRS, Liu2024Arxiv, Fang2025RS, ZhangJun2025TGRS}, where significant domain gaps arise from fundamentally different imaging mechanisms, object textures, and background clutter. Unlike optical-to-optical adaptation, SAR data introduce significant challenges in feature representation and alignment, prompting the development of specialized techniques such as domain-invariant feature learning, cross-modality alignment, and pseudo-label refinement~\cite{Shi2021JSTARs, ZhangJun2022TGRS, Shi2022TGRS, Pan2023TGRS, Zhang2025TGRS}.
  \subsubsection{SSDA-RS Methods} Semi-supervised domain adaptation in remote sensing aims to transfer knowledge from a labeled source domain to a partially labeled target domain, reducing annotation costs while adapting to new geographic regions or sensor types. Some methods focus on classification tasks~\cite{Tuia2014TGRS, Wang2018IGARSS, Li2020RS, Teng2020GRSL, Lasloum2021NISS, Lasloum2021RS, Yang2022GRSL, Kalita2022GRSL, HuangWei2023ISPRS, Li2024TGRS, Mo2024RS, Qu2024TGRS, Yaghmour2024JSTARs}, while others target semantic segmentation~\cite{Hafner2022RSE, Gao2023RS, Lucas2023ML, Yang2024TGRS} or object detection~\cite{Guo2021RS, Liao2022RS, Huang2024ICANN, Luo2024RS, ZhengXiangtao2023TGRS, Liao2024IGARSS, Deng2025Arxiv}, addressing different remote sensing tasks under the SSDA paradigm.
  \subsubsection{SDA/Finetuning-RS Methods} SDA assumes access to labeled data in both domains. As such, SDA is often treated more as a practical engineering solution rather than a fundamental research challenge. In many real-world applications, once a moderate amount of labeled target data is available, engineers typically opt to directly fine-tune or retrain a specific model on the target domain, which can yield strong performance without requiring complex domain adaptation techniques. Nevertheless, some methods~\cite{Tuia2011RSE, Bahirat2012TGRS, Aryal2023RS, Scheibenreif2024CVPR, Lyu2025TGRS} are still worth noting, as they provide valuable insights into leveraging target-domain labels for adaptation and generalization.
  \subsection{Taxonomy Based on Algorithm Granularity}
  \subsubsection{Distribution Measurement based Methods} As shown in Tab.~\ref{Tab3}, many notable methods utilize distribution measurement technologies for aligning distributions across diverse domains. DAN~\cite{Othman2017TGRS} introduces an auxiliary network, comprising multiple hidden layers, stacked on top of a pre-trained convolutional neural network (CNN). This design incorporates graph Laplacian and Maximum Mean Discrepancy (MMD) regularization terms besides the standard cross-entropy error. With adaptation on hidden layers, DAN is able to combat the domain shift problem. DDME~\cite{Wang2019GRSL} learns a manifold embedding and aligns the discriminative distribution by training a classifier for the source and target images. DACNN~\cite{Song2019GRSL} framework that integrates subspace alignment (SA) with CNNs to tackle the domain adaptation challenge in RS scene image classification. This method enables the CNN model to adapt seamlessly to the aligned feature subspace, thereby effectively mitigating discrepancies in domain distributions. TCANet~\cite{Garea2019RS} consists of several stages built based on convolutional filters that operate on patches of the hyperspectral image. Leveraging Transfer Component Analysis (TCA), the transformation matrix across different domains is learned by minimizing the metric of distribution distance. DDA-Net~\cite{Ma2019TGRS} first designs a domain alignment module to minimize the domain discrepancy guided by an irrelevant task by similarity measurement. Additionally, it employs a task allocation module to classify within the source domain with the regulation of alignment. Furthermore, it integrates a domain adaptation module to transfer the both the alignment capability and classification ability to the target domain. JCGNN~\cite{Wang2021JSTARs} enhances the measurement and alignment of joint distributions by embedding both domain-wise and class-wise CORAL into the GNN framework. With this design, it enables more precise feature matching and improved discriminability across domains. ADA-DDA~\cite{Yang2022TGRS} enhances domain adaptation by guiding marginal distribution alignment through attention mechanisms and dynamically balancing the importance of marginal and conditional distributions. TSTnet~\cite{Zhang2023TNNLS} introduces an integrated CNN-based semantic features with GCN-based topological structure modeling. By designing graph optimal transmission and a consistency constraint between CNN and GCN outputs, the method effectively aligns both distribution and topological relationships between domains. SSM~\cite{Liu2021JSTARs} addresses spectral shift in cross-scene HSI classification by proposing a spectral shift mitigation strategy, which combines amplitude normalization and adjacency effect correction. DAN\_MFAC~\cite{Wu2025KBS} involves multi-level feature alignment that explicitly constrains both global and category-wise distributions, thereby enhancing domain-invariant feature learning and improving generalization across scenes. DATSNET~\cite{ZhengZhendong2021TGRS} introduces task-specific classifiers and employs adversarial minimaxing of classifier discrepancy to achieve better alignment of source and target feature distributions while refining task-specific decision boundaries. MRDAN~\cite{Niu2022TGRS} employs a feature-fusion adaptation module to construct a broader domain-invariant space and a dynamic alignment mechanism to automatically balance local and global adaptation losses, thereby enhancing cross-domain performance. AMRAN~\cite{Zhu2022TGRS} jointly aligns marginal and conditional distributions while leveraging attention and multiscale strategies to enhance feature robustness and information completeness. SRKT~\cite{Zhao2023TGRS} learns sensor-invariant representations via adversarial and contrastive alignment. It explicitly models and aligns class-wise relationship distributions. DDCI~\cite{Zhao2025TGRS} introduces integrates an adaptation diffusion distillation module and a consistent causal intervention module to enable effective cross-domain knowledge transfer and eliminate spurious correlations, thereby improving model generalization and robustness. DCA~\cite{Wu2022TGRS} explicitly aligns category features to learn domain-invariant representations, which enhances intra-class compactness and inter-class separability under imbalanced and inconsistent distributions. To address the lack of labeled SAR data for ship detection, HSANet~\cite{ZhangJun2022TGRS} introduces a hierarchical domain adaptation framework that leverages labeled optical images. It aligns SAR and optical domains at both global (structure-level via Fourier-based alignment) and local (instance-level via prototype alignment) scales to enhance cross-modal detection performance.  CORAL-ADA~\cite{Koga2020RS} integrates  correlation alignment and adversarial domain adaptation into a region-based detector, further enhanced by reconstruction loss, achieving obvious performance gain in the target domain.
  \par Some semi-supervised domain adaptation methods are centered around distribution alignment or distance measurement, as the availability of a small amount of labeled target data allows for more reliable estimation of domain discrepancies and better guidance in aligning feature spaces between source and target domains. TDDA~\cite{Li2020RS}  proposes a two-stage deep domain adaptation method for hyperspectral image classification, which combines MMD-based domain alignment and a Spatial–Spectral Siamese Network with pairwise loss to learn a discriminative embedding space using few labeled target samples. DJ-CORAL~\cite{Yang2022GRSL} projects heterogeneous features into a shared subspace and jointly aligns marginal and conditional distributions. This method effectively leverages unlabeled target data to reduce domain shifts and improve classification performance. BSCA~\cite{HuangWei2023ISPRS} effectively reduces domain shift and leverages both labeled and unlabeled data across source and target domains by combining unsupervised maximum mean discrepancy alignment and supervised class-aware alignment. SSCA~\cite{Mo2024RS} integrates a rotation-robust convolutional feature extractor (RCFE) to handle rotation variations and a neighbor-based subcategory centroid alignment (NSCA) module to mitigate intra-class discrepancies across domains. CDMPC~\cite{Gao2023RS} introduces a contradictory structure learning mechanism and self-supervised learning strategy to enhance domain alignment and improve the use of limited labeled target samples. SAR-CDSS~\cite{Luo2024RS} proposes a semi-supervised cross-domain object detection framework for SAR imagery by leveraging optical data and a few labeled SAR samples. It reduces domain shift progressively at the image, instance, and feature levels through image mixing and instance swapping for data augmentation and an adaptive optimization strategy for selective feature alignment. CDTL-YOLOV5~\cite{Liao2024IGARSS} raises a semi-supervised SAR target detection method based on YOLOv5. By introducing feature domain adaptation constraints, the model transfers rich knowledge from labeled optical data to improve SAR feature learning under limited annotation.
  \subsubsection{Adversarial Learning Based Methods} As previously mentioned, adversarial learning techniques aims to derive information at either the feature-level, pixel-level, or both, to reduce the discrepancy that exist between various domains. Among the methods shown in Tab.~\ref{Tab3}, adversarial learning plays an important role on different tasks with different supervision paradigms. 
  \par \textbf{Feature-level Adversarial Learning Based Methods.} Feature-level adversarial learning aims to align the feature maps from both the source and target domains into a unified latent space, thereby mitigating biases during the process of mapping target features to predictions.
  \par At the global-level, feature adversarial alignment encourages the model to learn domain-invariant category-level semantics, thereby improving the consistency of predictions across domains. SSMT-RS~\cite{Zheng2020IGARSS} leverages meta-learning to distinguish multiple target domains and utilizes adversarial learning to confuse the distinction between the source domain and a mixed multi-target domain. These two processes, meta-learning and adversarial learning, operate iteratively and dynamically. ADA-Net~\cite{Ma2020TGRS} develops generator based on variational autoencoders, which is designed to acquire spectral and spatial characteristics from both domains. By incorporating two objective functions for adversarial learning, the global and local alignment will be taken into account. To address the domain-shift issue between multiple source domain datasets and single target domain dataset, MSCN~\cite{Lu2020TGRS} initially designs a pre-trained CNN to extract image features across diverse domains. Subsequently, the features from each source domain and the target domain are aligned by a cross-domain alignment module, which is composed of different source-specific discriminators. ECB-FAM~\cite{Ma2021RS} integrates an error-correcting boundaries mechanism into a feature-level adversarial learning framework. This integration enables the simultaneous construction of both domain-invariant features and semantically meaningful features. CDA~\cite{Liu2021TGRS} integrates class-wise adversarial alignment and PMMD (Probability MMD) strategies to achieve unsupervised domain adaptation by aligning domain-invariant features on a per-class basis. To address cross-domain feature distribution discrepancies in HSI classification, JDA~\cite{Miao2021JSTARs} first maps data into a shared latent space via a coupled-VAE (variational autoencoders) module, then refines class-wise alignment through a fine-grained  joint distribution alignment module. DFENet~\cite{ZhangXufei2022TGRS} combines a context-aware feature refinement (CAFR) module for adaptive extraction of global and local discriminative features with a multilevel adversarial dropout (MAD) module. This design enhances feature generalization by selectively suppressing low-quality information at both feature and decision levels. UDACA~\cite{Yu2022GRSL} proposes a novel UDA-HSI method, which performs content-wise feature alignment by jointly reducing class-level and style-perceive-level discrepancies through an adversarial framework. TAADA~\cite{Huang2022TGRS} extracts spectral-spatial joint information through attention-based dual-branch feature extraction and adversarial learning, enabling more effective cross-domain classification performance. SSWADA~\cite{Huang2023ISPRS} addresses the challenges of cross-scene wetland mapping with hyperspectral images. It introduces a spatial–spectral weighted adversarial domain adaptation framework that leverages joint 2D–3D convolutions, instance-weighted discrimination, and multi-classifier learning to enhance domain alignment and target classification. FDDAN~\cite{Xin2024JAG} designs a feature disentanglement-based domain adaptation network for cross-scene coastal wetland classification, aiming to extract class-specific domain-invariant features by explicitly separating out domain-specific and class-invariant components. Through a transformer-convolution fusion encoder and adversarial disentanglement strategy, the method enhances both transferability and class discriminability. AdaIN~\cite{Zhang2024GRSL} combines dual-level feature adversarial learning and adaptive instance normalization based simulation of cross-domain characteristics, the method achieves physically plausible adaptation between SAR and optical domains. ADDA~\cite{Wang2018IGARSS} extends the adversarial discriminative domain adaptation method to a semi-supervised setting for remote sensing object recognition, enabling effective adaptation to new geographic regions using both labeled and unlabeled data. CDADA~\cite{Teng2020GRSL} encourages the generator to produce transferable features that align across domains while remaining distant from class boundaries through incorporating dual land-cover classifiers as discriminators. DACNN-MME~\cite{Lasloum2021NISS} fuses cross-entropy loss on labeled data with an adversarial min-max entropy loss on unlabeled target data to jointly promote domain-invariant and class-discriminative feature learning. SSDAN~\cite{Lasloum2021RS} jointly optimizes standard cross-entropy loss and an adversarial min-max entropy loss, the method effectively learns domain-invariant yet discriminative features using both labeled and unlabeled target data. SASS~\cite{Qu2024TGRS} proposes a semi-supervised learning method for hyperspectral and LiDAR data classification by introducing a shared-private feature decoupling mechanism and multi-level feature alignment. This method effectively leverages both labeled and unlabeled data to enhance classification performance with limited annotations.
  \par At the pixel-level, feature-level adversarial learning mainly aim to enhance dense prediction consistency. MBATA-GAN~\cite{Ma2023TGRS} proposes a transformer-based framework, leveraging a mutually boosted attention module to effectively capture cross-domain semantic dependencies. By combining global attention with enhanced feature-level adversarial learning, it significantly improves feature transferability. JDAF~\cite{Huang2024TGRS} jointly aligns marginal and conditional distributions and incorporates an uncertainty-adaptive learning strategy to improve pseudo-label reliability and reduce domain gaps. CSLG~\cite{ZhangBo2022TGRS} progressively aligns features from local semantic to global structural levels based on patch-level adaptation difficulty. With this design, it effectively addresses both local distribution shifts and global domain gaps caused by diverse land covers and sensor variations. MemoryAdaptNet~\cite{Zhu2023TGRS} combines output space adversarial learning with a category-aware invariant feature memory module to bridge domain gaps and enhance feature consistency, achieving superior performance across multiple cross-domain tasks. De-GLGAN~\cite{Ma2024TGRS} enhances semantic segmentation in remote sensing by leveraging multiscale high/low-frequency decomposition and global–local generative adversarial learning to improve domain-invariant representation learning and cross-domain generalization. RCA-DD~\cite{Chen2022TGRS} proposes a region and category adaptive domain discriminator, which incorporates an entropy-based regional attention module and a class-clear module. This method focuses on hard-to-align regions and selectively updating category distributions during training. MIDANet~\cite{ChenHongyu2022TGRS} employs multitask learning and entropy-based adversarial training, leveraging semantic segmentation and elevation estimation to extract domain-invariant features and enhance generalization without requiring target-domain labels. MMDANet~\cite{Zhou2023TGRS} improves UDA-based semantic segmentation of high-resolution RSI by integrating DSM data through a multipath encoder and multitask decoder, and further aligns source and target domains using feature-level adversarial learning, enabling richer representations and improved segmentation via height-aware fusion refinement. HighDAN~\cite{Hong2023RSE} enhances the transferability by combining multi-resolution fusion and adversarial learning on multimodal remote sensing data (hyperspectral, multispectral, SAR) for cross-city segmentation.
  \par At the instance-level, feature-level adversarial learning mainly promotes feature alignment for more accurate object localization and recognition across domains. IDA~\cite{Pan2023TGRS} addresses domain shift by enforcing prediction consistency through a newly designed loss function. By aligning semantic information at both the image and instance levels, it enhances feature discrimination and reduces the risk of negative transfer, with theoretical analysis supporting its convergence properties. DFD-CAC~\cite{Zhang2025TGRS} utilizes dynamic feature discrimination and center-aware calibration to mitigate bounding box regression errors and feature misalignment caused by scattering differences across SAR sensors. By leveraging bidirectional feature aggregation and centerness-guided alignment, the approach enhances detection robustness under cross-domain conditions. FACL~\cite{Zhao2022TGRS} incorporates adversarial learning with specialized attention and compensation modules. The Adversarial Learning Attention (ALA) uses entropy-based weighting for precise instance- and pixel-level alignment, while the Compensation Loss Module (CLM) enforces prototype feature consistency across domains to enhance detection accuracy. To tackle feature incompatibility across domains, PDSCR~\cite{Luo2023GRSL} captures domain-specific features via patch-wise channel recalibration and leverages dynamic weighted prototype alignment (DWPA) to alleviate the impact of noisy pseudo-labels during early training. RIRA~\cite{Chen2021RS} framework introduces image-level and prototype-level feature alignment via a relation-aware graph, along with a rotation-invariant regularizer, effectively improving cross-domain detection robustness. FADA~\cite{Xu2022TGRS} integrates adversarial training (AT) via adversarial-based foreground alignment and further refines domain generalization through prototype-based confusing feature alignment. RFA-Net~\cite{Zhu2022JSTARs} introduces a unified framework combining data-level augmentation, sparse feature reconstruction, and pseudo-label generation. This design enhances instance-level alignment while reducing the influence of background noise in unsupervised domain adaptation scenarios. APA~\cite{Koga2021IGARSS} proposes an adversarial prediction-space alignment method that jointly adapts location and class confidence outputs for vehicle detection in satellite images, yielding improvement and demonstrating enhanced cross-domain detection performance. FIE-Net~\cite{ZhangJun2025TGRS} addresses the foreground–background imbalance in a foreground-enhanced alignment framework. It emphasizes salient instance features by integrating a foreground-focused multigranularity alignment module and a progressive label filtering strategy. DA-FRCNN~\cite{Guo2021RS} proposes a domain adaptive Faster R-CNN~\cite{FasterRCNN} framework for SAR target detection with limited labeled data by transferring knowledge from labeled optical images. It uses a GAN-based constraint after proposal generation, achieving superior detection performance on small SAR datasets.
 \par \textbf{Image-level Adversarial Learning Based Methods.} Image-level adversarial learning seeks to reduce domain discrepancies by translating source images into the target domain style (or vice versa) while preserving semantic content. By aligning visual appearances at the pixel level, the model can perceive source and target images more consistently, thus facilitating downstream tasks under domain shift conditions.
 \par VSFA~\cite{Zhang2023TGRS} integrates visual features and scattering topological features to improve target recognition between synthetic and measured SAR images. By leveraging graph neural networks and a class-wise alignment strategy, the method achieves superior performance. UDA-GAN~\cite{Benjdira2019RS} is the first method to apply a GAN-based segmentation network for addressing the UDA-RSSeg task. It introduces an image-level adversarial learning scheme, where a generator translates source images into the target domain style, and a discriminator encourages the translated images to be visually indistinguishable from real target images, thereby reducing the domain gap at the pixel level. ResiDualGAN~\cite{Zhao2023RS} introduces an in-network resizer to handle scale discrepancies and residual connections to stabilize real-to-real translation, achieving notable improvements in semantic segmentation accuracy under unsupervised domain adaptation settings. BiFDANet~\cite{Cai2022RS} simultaneously optimizes source-to-target and target-to-source translations to leverage complementary domain information. By integrating a bidirectional semantic consistency loss and combining dual classifiers at inference, it outperforms conventional unidirectional approaches across multiple datasets. MHDA~\cite{Liang2023TGRS} introduces a unified multilevel unsupervised domain adaptation framework for remote sensing segmentation, with a particular emphasis on image-level adversarial learning via cycle consistency. By combining instance-level alignment, feature-level contrastive learning, and decision-level task consistency, it effectively addresses complex domain shifts without relying on target-domain supervision. FDDA~\cite{Liao2022RS} integrates a reconstruction-based decoding module and a domain-adaptation module to leverage both unlabeled SAR data and labeled optical remote sensing (ORS) images. By jointly optimizing detection, reconstruction, and domain alignment losses, the model effectively transfers knowledge from ORS to SAR and achieves improved detection performance with limited labeled SAR data.
 \par \textbf{Feature-level and Image-level Integrated Adversarial Learning Based Methods.} Integrated adversarial learning methods combine both image-level and feature-level alignment to comprehensively reduce domain discrepancies. Image-level adaptation aligns the visual appearance of source and target images to mitigate low-level distribution gaps, while feature-level adaptation ensures that high-level semantic representations are mapped into a shared latent space. This joint optimization enables more effective representation.
 \par PFDA~\cite{Chen2022GRSL} proposes a pixel-level and feature-level domain adaptation method that combines image translation and feature alignment to address the challenges of distribution discrepancies in heterogeneous SAR target recognition. SDEnet~\cite{Zhang2023TIP} employs adversarial domain expansion with spatial-spectral randomization and supervised contrastive learning to enhance domain-invariant representation DNT~\cite{Chen2022JAG} proposes a dual-space alignment framework that reduces input-level discrepancies using a Digital Number Transformer and enhances feature-level alignment through multi-scale feature aggregation and fine-grained discrimination. IterDANet~\cite{Cai2022TGRS} incorporates a progressive framework that first enhances inter-domain alignment and then iteratively refines intra-domain consistency through subdomain clustering and pseudo-label optimization. DCLDA~\cite{Biswas2024JSTARs} leverages curated support sets for multi-level domain alignment, and a novel contrastive loss to mitigate false negative bias and enhance cross-domain robustness for UDA-RSDet task. ML-UDA~\cite{Luo2024JSTARs} proposes a multilevel unsupervised domain adaptation framework for single-stage object detection in remote sensing images, integrating pixel-level and feature-level adaptations in a progressive manner. To address local deformation and scale variation, this method incorporates semantic region-aware translation and attention-guided multiscale feature alignment.
 \subsubsection{Self-Training Based Methods} Self-training based domain adaptation methods improve target domain performance by iteratively generating pseudo-labels for unlabeled data and retraining the model. These methods exploit the model's own predictions to gradually refine decision boundaries, enabling better class discrimination under domain shift. 
 \par UDA-SAR~\cite{Shi2024TGRS} utilizes pseudo-label-based contrastive learning to enhance class-wise alignment between the optical source and SAR target domains. By integrating consistency constraints and cross-domain pseudo-label supervision, the method effectively mitigates source bias and progressively refines target domain representations for accurate SAR ship classification. DST~\cite{Zhao2024GRSL} presents a novel unsupervised domain adaptation method for heterogeneous SAR image classification. It introduces a dynamic self-training framework embedded with a domain-specific weak alignment module that adaptively distinguishes known and unknown classes. MLUDA~\cite{Cai2024TGRS} is developed to leverage self-training with pseudo-labels and supervised contrastive learning to enhance inter-class separability. CPCA~\cite{Zhu2024TGRS} disentangles causal and bias features and leveraging contrastive learning with causal prototypes and counterfactual interventions. In addition to this design, it involves a self-training strategy that leverages reliable pseudo-labels generated from causal features to guide target domain learning. MEBS~\cite{Li2025TGRS} proposes a self-supervised teacher–student UDA framework for remote sensing image segmentation, addressing class confusion and imbalance through two key techniques: mask-enhanced class mix (MECM) for improved contextual learning and scale-based rare class sampling (SRCS) to better represent small-scale categories. MS-CADA~\cite{Gao2024TGRS} constructs expert-specific branches, employs cross-domain mixing, adaptive pseudo-labeling, and multi-view knowledge integration to effectively transfer knowledge across mismatched class spaces in a class-asymmetric multi-source setting. EUDA-PLR~\cite{Cui2025JAG} designs a multi-factor guided pseudo-label refinement strategy to effectively transfer knowledge from historical SAR datasets to new oil spill events with limited annotations. This method can improve segmentation accuracy across various sensors and marine conditions. DualDA-Net~\cite{ZhuYangguang2023TGRS} designs a dual-head rectification framework combining multilevel feature alignment with coarse-to-fine consistency and collaborative pseudo-label refinement. It progressively enhances target supervision and mitigates label bias via a teacher–student co-training scheme. RST~\cite{Han2024TGRS} introduces the first DETR-based framework, which enhances cross-domain alignment via a self-adaptive pseudo-label assigner, effectively improving detection in sparse-object and noisy-background scenarios. SFOD~\cite{Liu2024Arxiv} performs unsupervised domain adaptation using only a pre-trained source model without access to source domain data. It embeds multi-level perturbations and a teacher–student consistency strategy to align the target and perturbed domains. EasySeg~\cite{Yang2024TGRS} integrates interactive and active learning. By introducing a point-level labeling strategy (SFAL) and an interactive segmentation network (ISS-Net), the method efficiently annotates informative pixels and refines pseudo-labels to improve domain adaptation performance. SRA-YOLO~\cite{Huang2024ICANN} uses a teacher-student strategy with knowledge distillation and adaptive zoom-in/out techniques to mitigate spatial resolution differences. This method effectively utilize both labeled and unlabeled data to address domain discrepancies by aligning Ground Sample Distance (GSD) across varied aerial imagery. DT~\cite{ZhengXiangtao2023TGRS} divides the learning process into cross-domain and semi-supervised subtasks, each guided by a separate teacher–student model. These models collaboratively refine the detector by generating and utilizing pseudo-labels from unlabeled SAR data. WeedTeacher~\cite{Deng2025Arxiv} designs a YOLOv8-based semi-supervised object detection framework for weed detection that integrates EMA-based self-training to effectively leverage unlabeled data. 
 \subsubsection{Integrated-Training Based Methods} Integrated-training based methods unify adversarial learning and self-training to jointly promote domain alignment and class-level discrimination. Adversarial learning reduces domain gaps, while self-training progressively refines target predictions, leading to improved adaptation performance.
 \par AST~\cite{Huang2022IGARSS} proposes an adversarial self-training framework that integrates self-training and adversarial learning to simultaneously refine decision boundaries and enhance cross-domain feature alignment for high-resolutional aerial scene classification. PPLM-Net~\cite{Leng2024JSTARs} solves domain shift and complex backgrounds through three key components: domain adversarial training for extracting domain-invariant features, partial patch local masking for enhancing global contextual learning, and a teacher–student network self-training strategy for generating pseudo-labels to boost target domain performance. Tri-ADA~\cite{Yan2021TGRS} introduces a triplet-based adversarial framework that jointly leverages source and target domain information via a domain similarity discriminator, alongside a class-aware self-training strategy for reliable pseudo-labeling. FGUDA~\cite{Wang2023JSTARs} proposes a fine-grained adaptation framework that integrates global-local alignment with category-level self-training mechanism. By focusing on hard-to-align regions and compacting dispersed category features, it effectively mitigates negative transfer and enhances semantic consistency in the target domain. RoadDA~\cite{Zhang2022TGRS} introduces a stagewise domain adaptation framework for remote sensing road segmentation, combining image-level adversarial alignment and adversarial self-training. By progressively adapting both interdomain and intradomain features, it effectively mitigates domain shift and enhances segmentation performance in the target domain. STADA~\cite{Liang2023GRSL} integrates adversarial feature alignment with self-training on denoised pseudo-labels. It improves model adaptability to out-of-distribution target domains. ST-DASegNet~\cite{Zhao2024JAG} addresses domain shifts caused by sensor, resolution, and geographic differences. It combines feature-level adversarial learning with an EMA-based self-training strategy to enhance cross-domain generalization. CDANet~\cite{Yang2024NN} combines adversarial learning, pixel-wise contrastive loss, and a self-training strategy, which can effectively align cross-domain features and enhances the extraction of small, densely distributed buildings. DDF~\cite{Ran2024TGRS} introduces a hybrid training strategy, which leverages both image-level style-transferring and pseudo-labels refinement based on spatial context. TDAIF~\cite{GaoXianjun2024TGRS} tackles cross-domain cloud detection by coupling image-level pseudo-target generation with a feature-level domain discriminator and self-ensembling strategy, effectively mitigating distribution gaps caused by radiometric and scale differences. FRCNN-SAR~\cite{Shi2021JSTARs} is the first unsupervised domain adaptation framework for SAR target detection, improving performance in unlabeled target domains without costly annotations. It combines pixel-level adaptation, multilevel feature alignment, and iterative pseudo-labeling to address domain shifts from varying radar parameters and scenes. PT-SAR~\cite{Shi2022TGRS} transfers knowledge from optical images to SAR data across three stages: pixel-level appearance adaptation using GAN-based translation, feature-level domain alignment via adversarial learning, and prediction-level enhancement through robust self-training to mitigate noisy pseudo-labels. CDST~\cite{Luo2024TGRS} employs GAN-based domain transfer to reduce domain shift, followed by a hard example selection strategy that improves pseudo-label quality through confidence and relational scoring mechanisms. This framework enhances the robustness and accuracy of object detection in cross-domain remote sensing scenarios. MGDAT~\cite{Fang2025RS} enhances remote sensing object detection under domain shift by incorporating pixel-level, image-level, and instance-level feature alignment into a teacher–student architecture, enabling more robust and fine-grained cross-domain representation learning. CFAN~\cite{Li2024TGRS} aligns source and target domain features at the category level while enhancing pseudo-label quality to mitigate noise by combining a dual-directional prototype alignment module with adversarial training.
 \subsubsection{LVM Based Methods} The large vision model (vision foundation model) has revolutionized AI and deep learning, empowering remote sensing with stronger generalization across diverse scenes. Lu~\textit{et al.}~\cite{Lu2025GRSM} highlights emerging trends and key advancements driven by LVMs in remote sensing field. Although the use of large vision models (LVMs) for DA-RS tasks remains under-explored, several notable methods~\cite{Lyu2025TGRS, Scheibenreif2024CVPR, Gong2024Arxiv} have shown promising potential and merit further attention. PFM-JONet~\cite{Lyu2025TGRS} tackles UDA for semantic segmentation in very-high-resolution (VHR) remote sensing imagery by integrating the Segment Anything Model (SAM)~\cite{SAM} as a foundation model to mitigate feature inconsistency and domain gap issues. Through a hybrid training strategy combining adversarial learning and self-training at both feature and prediction levels, the framework achieves robust adaptation across diverse urban scenes. SLR~\cite{Scheibenreif2024CVPR} explores efficient adaptation strategies for large pre-trained foundation transformer models in remote sensing. It aims to overcome the distribution gap between pre-training and downstream tasks without the high cost of full training. CrossEarth~\cite{Gong2024Arxiv} is the first to propose foundation model tailored for remote sensing domain generalization in semantic segmentation.
 \par As a whole, domain adaptation in remote sensing has evolved from traditional shallow methods to deep learning based frameworks, and now toward foundation model driven directions. Along this trajectory, several representative paradigms have emerged. Distribution measurement based methods reduce domain gaps by aligning statistical distributions, laying the foundation for early DA research. Adversarial learning based methods exploit domain discriminators for global alignment, particularly effective in classification tasks with coarse domain shifts. Self-training based methods leverage pseudo-labels for progressive adaptation, proving especially powerful in dense prediction tasks such as segmentation and detection. Integrated-training methods combine alignment, self-training, and auxiliary techniques like contrastive learning or style transfer to achieve more robust semantic consistency. Most recently, large vision model based methods are gaining momentum, adapting pretrained foundation models to remote sensing tasks and opening a new paradigm with transferable and open-vocabulary capabilities.
 \begin{table}[!t]
  \scriptsize
  \centering
  \caption{Representative DA-RS methods with different input modes.}
  \scalebox{1.0}{
  %\begin{tabular}{c|cccc}
  \begin{tabular}{p{0.08\textwidth}|p{0.08\textwidth}|p{0.25\textwidth}}
  \hline \hline
  {} & \makecell[c]{\textbf{Methods}} &  \makecell[c]{\textbf{DA-RS methods}}
  \\ \cline{1-3}
  \makecell[c]{\multirow{5}{*}{\shortstack{\\ \\ \\ \\ \\ \\ \\ \\ \\ \\ \\ \\ \\ \\ \\ \\ \\ \\ \\ \\ \\
  \textbf{Taxonomy} \\ 
  \textbf{Based on} \\
  \textbf{Input Mode} }}} & \makecell[c]{One-to-Many} & \makecell[c]{
  \\ SSMT-RS~\cite{Zheng2020IGARSS}, GNN-MTDA~\cite{Saha2022GRSL},  \\ EHACA~\cite{Ngo2023TGRS},  
  HFPAN~\cite{Miao2024TGRS}, \\ TSAN~\cite{Zheng2022TGRS}, DAL~\cite{Zhang2024TGRS}, CCDR~\cite{Liang2024TGRS} \\ \\} 
  \\ \cline{2-3}
  {} & \makecell[c]{Many-to-Many} & \makecell[c]{\\ SSCA~\cite{Mo2024RS}, DAugNet~\cite{Tasar2021TGRS}, MultiDAN~\cite{Cai2024MM}, \\ M$^{3}$SPADA~\cite{Capliez2023TGRS}, MDGTnet~\cite{Qi2024GRSL}, \\ FDINet~\cite{Gao2025TNNLS}, GeIraA-Net~\cite{Yang2024TIP} \\ \\}
  \\ \cline{2-3}
  {} & \makecell[c]{Many-to-One} & \makecell[c]{\\ MSCN~\cite{Lu2020TGRS}, SSDAN~\cite{Lasloum2021RS}, MS-CADA~\cite{Gao2024TGRS}, \\ AMDA~\cite{Elshamli2020TGRS}, IMIS~\cite{Gong2021TGRS}, Swin-DA~\cite{Li2024JSTARs}, \\ ALKA~\cite{Rahhal2021GRSL}, 
  MECKA~\cite{NgoBaHung2023TGRS},  MSSDANet~\cite{Wang2025RS}\\ \\}
  \\ \cline{2-3}
  {} & \makecell[c]{Source-free} & \makecell[c]{\\ HSI-SFDA~\cite{Xu2022IGARSS}, MSF-UDA~\cite{Zhang2023ICGRSM}, \\ SFOD~\cite{Liu2024Arxiv}, MLDP-SFOD~\cite{Hong2024GIS},  SFUDA~\cite{Mohammadi2024RSE}, \\ APD-SFDA~\cite{Gao2024GRSL}, LPLDA~\cite{Kim2025GRSL}, \\ SD-SFDA~\cite{Zhang2025GRSL}, S$^{3}$AHI~\cite{Ding2024IJCNN}, SMNA~\cite{Zhu2025TGRS} \\ \\ }
  \\ \cline{1-3} \hline \hline
   \end{tabular}
  }
  \label{Tab4}
\end{table}
 \subsection{Taxonomy Based on Input Mode}
 As shown in Fig.~\ref{Fig3}, existing DA-RS methods can also be categorized based on different input modes. The aforementioned methods primarily focus on the widely studied ``One-to-One'' configuration. In the following sections, we will review methods with other input modes listed in Tab.~\ref{Tab4}.
 \subsubsection{One-to-Many}
 To address the limitations of single-target domain adaptation in remote sensing, SSMT-RS~\cite{Zheng2020IGARSS} pioneers the research of single-source-multiple-target domain adaptation in remote sensing by constructing a challenging mixed multi-target dataset and proposing a meta-adversarial learning framework. GNN-MTDA~\cite{Saha2022GRSL} proposes a multi-targets adaptation framework that learns a unified classifier across multiple unlabeled target domains by leveraging graph-based co-teaching and a sequential adaptation strategy, effectively aggregating features from both source and diverse target domains. Building upon the objective of improving MTDA, EHACA~\cite{Ngo2023TGRS} proposes an easy-to-hard adaptation framework that preserves semantic integrity within each target domain and effectively models both source–target and inter-target relationships through hierarchical intra-target and collaborative inter-target feature alignment. HFPAN~\cite{Miao2024TGRS} integrates fine-grained global-local feature extraction with hierarchical feature embedding and progressive alignment. With this design, this method effectively addresses domain shifts across multiple target domains. S$^{2}$AMSnet~\cite{ChenXi2024TGRS} expands the source-domain distribution via spectral–spatial generative networks and enforces semantic consistency through adversarial learning, enabling generalization to unseen single or multiple domains. TSAN~\cite{Zheng2022TGRS} integrates adversarial learning for global domain alignment and self-supervised learning for pseudo-domain label generation, enabling iterative refinement of multiple target domain representations without manual annotation. DAL~\cite{Zhang2024TGRS} comprises three key modules, which are multi-domain style transfer (MST), multi-domain feature approximation (MFA), and multi-domain cascaded instance extraction (MCIE). Three modules collaboratively bridge domain discrepancies and enhance extraction accuracy across multiple target domains. CCDR~\cite{Liang2024TGRS} bridges domain shifts and achieves superior performance without complex pipelines or external data. It effectively enhances generalization from a single source to multiple unseen target domains.
 \subsubsection{Many-to-Many} SSCA~\cite{Mo2024RS} is a typical method addressing ``many-to-many'' semi-supervised domain adaptation task. It is particularly effective in remote sensing scenarios involving heterogeneous multiple data sources, enhancing the model’s robustness and generalization across varied target domains. DAugNet~\cite{Tasar2021TGRS} enhances generalization across evolving and heterogeneous domains, outperforming existing methods in adapting to new geographic locations. MultiDAN~\cite{Cai2024MM} proposes a novel unsupervised multistage, multisource, and multitarget domain adaptation framework for remote sensing image segmentation. By integrating multi-adversarial learning, entropy-based subdomain clustering, and a dynamic pseudo-labeling strategy, it effectively addresses both inter-domain and intra-domain shifts across multiple target domains. M$^{3}$SPADA~\cite{Capliez2023TGRS} is proposed for cross-temporal land cover classification using multisensor, multitemporal, and multiscale remote sensing data. It enables effective model transfer across different time periods within the same geographic area, addressing distribution shifts caused by environmental and sensor variations. MDGTnet~\cite{Qi2024GRSL} captures both domain-specific and shared features, enabling robust classification on unseen target domains. FDINet~\cite{Gao2025TNNLS} combines a weight-sharing baseline for common feature extraction, modality-based similarity estimation, and a sharpness-aware feature discriminating (SAFD) strategy. With this design, FDINet enhances generalization without sacrificing feature discrimination. GeIraA-Net~\cite{Yang2024TIP} enhances cross-scene classification of multisource remote sensing data by jointly modeling global-local features and inter-class relations. A tailored pseudo-label-guided alignment strategy further refines domain calibration.
 \subsubsection{Many-to-One} MSCN~\cite{Lu2020TGRS} is the first article to address UDA task on multiple source domain datasets with unshared categories in remote sensing. Unlike most UDA methods assuming class symmetry, MS-CADA~\cite{Gao2024TGRS} addresses the more realistic yet underexplored problem of class-asymmetric adaptation for remote sensing images using multiple sources. SSDAN~\cite{Lasloum2021RS} proposes a multi-source semi-supervised domain adaptation method for remote sensing scene classification that leverages a pre-trained EfficientNet-B3~\cite{EfficientNet} to extract discriminative features and employs a prototype-based classification module with cosine distance. AMDA~\cite{Elshamli2020TGRS} efficiently adapts training data from multiple domains and achieves significant performance gains for local climate zone (LCZ) classification task. IMIS~\cite{Gong2021TGRS} addresses cross-domain scene classification with incomplete multiple source domains, where target categories may be partially unknown to each source. To handle this, it proposes a separation mechanism that distinguishes known and unknown categories in the target domain before performing alignment. Swin-DA~\cite{Li2024JSTARs} aligns source and target domains without repeatedly accessing source data, achieving superior performance and stability across multiple high-resolution remote sensing datasets. ALKA~\cite{Rahhal2021GRSL} proposes a novel multi-source unsupervised domain adaptation architecture. To reduce domain discrepancy, a Minmax entropy optimization strategy is employed, adversarially aligning target features with source classifiers for improved generalization. MECKA~\cite{NgoBaHung2023TGRS} designs a unified framework for multisource-single-target domain adaptation in remote sensing. It smoothly transfers information to target domains with incomplete class overlap, enhancing classification robustness. MSSDANet~\cite{Wang2025RS} is a novel unsupervised multi-source domain adaptation framework. It combines a two-stage feature extraction strategy with newly designed discriminant semantic transfer (DST) and class correlation (CC) losses, which enhances semantic alignment across domains without requiring target domain labels.
 \subsubsection{Source-Free} Source-free domain adaptation refers to adapting a pretrained source model to an unlabeled target domain without access to source data, relying solely on the target data and the transferred model parameters. HSI-SFDA~\cite{Xu2022IGARSS} generates source spectral features and aligning them with target features via contrastive learning, combined with a logits-weighted prototype classifier for iterative pseudo-labeling, the method enables effective target domain adaptation without access to labeled source data. MSF-UDA~\cite{Zhang2023ICGRSM} combines weighted information maximization and pseudo-labeling losses to align source and target feature distributions across multiple source domains. It enhances target domain classification accuracy while reducing dependence on source data during adaptation. MLDP-SFOD~\cite{Hong2024GIS} consists of domain perturbation, multi-level alignment, and prototype-based distillation within a mean teacher framework. With this design, it effectively extracts domain-invariant features to adapt source-pretrained models to target domains without accessing source data. SFUDA~\cite{Mohammadi2024RSE} proposes a source-free unsupervised domain adaptation (UDA) method for cross-regional and cross-time crop type mapping, effectively addressing the limitations of traditional UDA methods that require source data during adaptation. This method demonstrates strong generalization capabilities across different countries, sensors, and time periods. SFOD~\cite{Liu2024Arxiv} also addresses the limitations of traditional UDA methods that rely on access to source data during adaptation. It effectively extracts domain-invariant features without requiring source images. APD-SFDA~\cite{Gao2024GRSL} pioneers source-free domain adaptation segmentation for remote sensing images, eliminating the need for source data during adaptation. By integrating vision foundation models with attention-guided prompt tuning and a similarity-based feature alignment strategy, the method effectively adapts pretrained models to diverse target images. LPLDA~\cite{Kim2025GRSL} distills low-confidence proposals and introduces an instance consistency loss. This approach enhances the robustness of small object representations under domain shifts, making it particularly effective for fine-grained targets in aerial scenes. SD-SFDA~\cite{Zhang2025GRSL} proposes a source-free cross-sensor adaptation method for ship detection in SAR images, addressing the limitations of fixed-threshold pseudo-labeling in heterogeneous domains like optical-to-SAR transfer. S$^{3}$AHI~\cite{Ding2024IJCNN} adapts source-trained models to target domains using only unlabeled data during test-time training. It integrates slicing aided hyper inference (SAHI) and instance-level contrastive learning under a mean-teacher framework to overcome both data scarcity and domain shift challenges. SMNA~\cite{Zhu2025TGRS} incorporates a model statistics-guided alignment strategy alongside a noise adaptation module, enabling the target domain features to align with the source distribution while mitigating the impact of label noise.
 \subsection{Others}
 \subsubsection{Domain Adaptation across Varying Label Spaces} While our focus remains on domain adaptation under shared label spaces, real-world remote sensing scenarios often involve mismatched category configurations between source and target domains. Cross-label settings like partial, open-set, and universal domain adaptation pose additional challenges beyond distribution alignment. Since these approaches primarily tackle label space discrepancies, we briefly mention them here without delving into detailed discussion.
 \par
\begin{figure}[!t]
  \centering
 \includegraphics[width=0.48\textwidth]{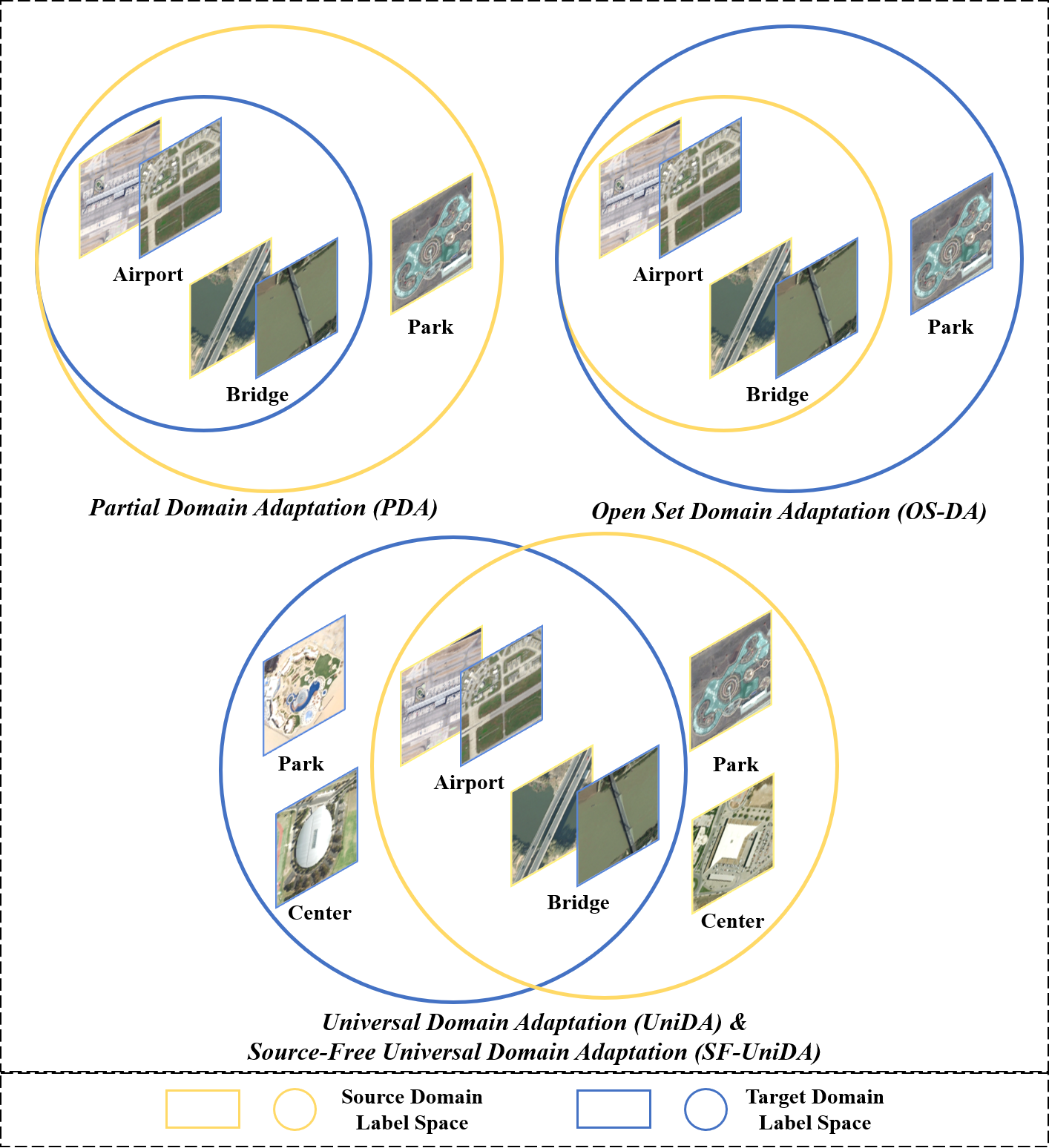}
  \caption{Illustration of the domain adaptation paradigm across varying label space.}
  \label{Fig7}
\end{figure}

 As shown in Fig.~\ref{Fig7}, partial domain adaptation (PDA) assumes that target label sets are considered a subset of source label sets. In contrast, open set domain adaptation (OS-DA) assumes that the target label set contains unknown classes not present in the source label set, which means that the source label sets are considered a subset of target label sets. Universal domain adaptation (UniDA) poses no prior knowledge on the label sets of source and target domains. Instead, it assumes that each domain contains a mix of shared and private (domain-specific) label sets. Building on this, source-free universal domain adaptation (SF-UniDA) further assumes that the source data is inaccessible during adaptation, and only a trained source model and unlabeled target data are available. In the field of remote sensing, numerous recent advances have been made to address the challenges of PDA~\cite{Zheng2023TGRS, Ma2023RS}, OS-DA~\cite{Zhang2020IGARSS, Zhao2022GRSL, ZhangJun2022TGRS2, Wang2023TGRS, Niu2023GRSL, Zhao2024GRSL, Zheng2024ISPRS, Wang2024BigData, ZhangXiaokang2024TGRS, Liu2025TETCI}, UniDA~\cite{Li2024IGARSS, Chen2025CIS, Guo2024GRSL, LiQingmei2024TGRS, LiQingmei2025TGRS}, and SF-UniDA~\cite{XuQingsong2022IGARSS, Xu2023TGRS, Zhao2023GRSL}. 
 \par \textbf{PDA Remote Sensing Methods.} PDA~\cite{Zheng2023TGRS} makes the first attempts to address the partial domain adaptation problem in remote sensing scene classification, where the source label space subsumes the target's. A progressive auxiliary domain module, multi-weight adversarial learning, and attentive entropy regularization are introduced to mitigate negative transfer and improve adaptation performance. PDANN~\cite{Ma2023RS} designs a partial domain adversarial neural network to address domain shift and label space mismatch in crop yield prediction, improving transferability by downweighing outlier source samples during domain alignment. 
 \par \textbf{OS-DA Remote Sensing Methods.} OSDANet~\cite{Zhang2020IGARSS} tackles open set domain adaptation for remote sensing image scene classification by leveraging adversarial learning between a feature generator and a classifier to distinguish shared classes and reject unknown target samples. ODA-SAR~\cite{Zhao2022GRSL} proposes a spherical space domain adaptation network for SAR image classification under open set conditions, where features are mapped to a hypersphere to enhance class separability. OSDA-ETD~\cite{ZhangJun2022TGRS2} jointly explores transferability to reduce interdomain and intraclass discrepancies, while enhancing interclass separability through discriminability, thereby effectively addressing the challenges of unseen categories and strong interclass similarity in remote sensing images. SSOUDA~\cite{Wang2023TGRS} addresses open-set unsupervised domain adaptation in optical remote sensing by integrating contrastive self-supervised learning with consistency self-training, enabling the extraction of discriminative target features and reliable unknown class samples. DST~\cite{Zhao2024GRSL} is a novel approach designed to tackle open-set UDA challenges in heterogeneous SAR image classification. The proposed framework adaptively distinguishes unknown target classes from known ones, enabling more robust cross-domain recognition. IAFAN~\cite{Niu2023GRSL} employs an instance affinity-based mechanism to separate unknown samples and a sample discriminability loss to enhance class separation, while a novel Mask-MMD metric ensures precise alignment of known classes without negatively transferring unknown ones. MAOSDAN~\cite{Zheng2024ISPRS} enhances model robustness in complex real-world scenarios with partially overlapping label spaces. It integrates attention-aware backpropagation, auxiliary adversarial learning, and adaptive entropy suppression to effectively separate known and unknown target classes while mitigating negative transfer. OSDA-ST~\cite{Wang2024BigData} introduces max-logit score thresholding and dynamic source selection to estimate unknown classes. PUMCL~\cite{ZhangXiaokang2024TGRS} effectively handles unknown classes and improves cross-domain generalization by aligning features and refining pseudo-labels. DFEN~\cite{Liu2025TETCI} jointly captures local discriminative patterns and global semantic correlations to enhance the detection of unseen categories by leveraging sample similarity to known classes.
 \par \textbf{UniDA Remote Sensing Methods.} C$^3$DA~\cite{Guo2024GRSL} proposes a universal domain adaptation method for remote sensing scene recognition, which introduces a novel ``C$^3$'' criterion that integrates confidence, consistency, and certainty to effectively identify unknown classes. By optimizing the loss function design, it enhances training efficiency and improves adaptability across diverse DA scenarios. HyUniDA~\cite{Li2024IGARSS, LiQingmei2024TGRS} eliminates reliance on prior label space knowledge. It effectively identifies shared classes and estimates target domain structure, enabling robust knowledge transfer across diverse HSI domains. SPOT~\cite{Chen2025CIS} leverages unbalanced optimal transport and a sample complement mechanism to effectively distinguish shared and private classes. By enhancing inter-class separability and intra-class compactness, it achieves improved accuracy and robustness under complex domain shifts. DCMix~\cite{LiQingmei2025TGRS} integrates closed-set and open-set classifiers with a feature mixup strategy, which enhances target domain generalization and enables more robust cross-domain transfer.
 \par \textbf{SF-UniDA Remote Sensing Methods.} SDG-UniDA~\cite{XuQingsong2022IGARSS} first generates synthetic source data from a pre-trained model and then adapts using a transferable weight mechanism to distinguish shared and unknown classes. SDG-MA~\cite{Xu2023TGRS} proposes a UniDA framework for remote sensing image scene classification without prior knowledge of label sets. By synthesizing source data from a pretrained model and introducing transferable weights, the method effectively identifies shared categories while detecting unknown classes in the target domain. OKRA~\cite{Zhao2023GRSL} employs neighborhood similarity for knowledge distillation and utilizes energy-based uncertainty modeling to distinguish known and unknown target samples effectively, without requiring source data or manual thresholding.
 \subsubsection{DA-RSCD} As previously mentioned, the DA-RSCD task differs from DA-RSCls, DS-RSSeg, and DA-RSDet in that the domain shift occurs within paired images, rather than across separate domains (source and target domains). The DA-RSCD task, also known as Cross-Domain Change Detection (CDCD), focuses on detecting changes in remote sensing image pairs acquired from different sensors or under varying imaging conditions. \textit{Chen et al.}~\cite{Chen2024JSTARS} provide the first systematic review of DA-RSCD. This survey examines the key components of this research direction, including image preprocessing, feature representation and change detection strategies, while identifying current challenges and future research avenues. \textit{Cheng et al.}~\cite{Cheng2024RS} also investigate the domain adaptation issue in the realm of change detection tasks.
\begin{table*}[!ht]
  \scriptsize
  \centering
  \caption{The representative benchmark datasets along with their detailed information for different DA-RSCls task.}
  \scalebox{1.0}{
  \begin{tabular}
  {p{0.14\textwidth}|p{0.1\textwidth}|p{0.09\textwidth}|p{0.17\textwidth} |p{0.12\textwidth}|p{0.09\textwidth}|p{0.1\textwidth}}
    \hline\hline
    \makecell[c]{\textbf{Dataset}} & \makecell[c]{\textbf{Image Size}} & 
    \makecell[c]{\textbf{Resolution}} & \makecell[c]{\textbf{Location/Source}} & \makecell[c]{\textbf{Imaging Mode}} & \makecell[c]{\textbf{Available Date}} & \makecell[c]{\textbf{Categories}}\\
  \hline \hline
  \rowcolor{gray!30!} \multicolumn{7}{c} {\textbf{Hyperspectral Image (HSI) Dataset}}  \\
    \hline
    \makecell[c]{ Houston2013~\cite{Debes2014JSTARS} \\ Houston2018~\cite{Bertrand2018GRSM}} & \makecell[c]{$349\times1905$ \\ $601\times2384$} & \makecell[c]{$2.5m$ \\ $0.5m\sim1m$} & \makecell[c]{University of Houston campus \\ and its neighboring area, \\ USA} & \makecell[c]{HSI, LiDAR} & \makecell[c]{$2013$ \\ $2018$} & \makecell[c]{$15$ \\ $20$} \\
    \hline
    \makecell[c]{Pavia University\footnotemark[1] \\ Pavia Center\footnotemark[1]} & \makecell[c]{$610\times340$ \\ $1096\times715$} & \makecell[c]{$1.3m$} & \makecell[c]{Pavia \\ Northern Italy} & \makecell[c]{HSI} & \makecell[c]{$2003$} & \makecell[c]{$9$ \\ $9$} \\
    \hline
    \makecell[c]{Botswana May\footnotemark[2] \\ Botswana June\footnotemark[2] \\ Botswana July\footnotemark[2]} & \makecell[c]{$1496\times256$} & \makecell[c]{$30m$} & \makecell[c]{Okavango Delta, \\ Botswana} & \makecell[c]{HSI} & \makecell[c]{$2001$} & \makecell[c]{$9$ \\ $9$ \\ $9$} \\
    \hline
    \makecell[c]{HyRank Dioni~\cite{HyRANK} \\ HyRank Loukia~\cite{HyRANK}} & \makecell[c]{$1376\times176$ \\ $945\times176$ \\ } & \makecell[c]{$30m$} & \makecell[c]{Athens} & \makecell[c]{HSI} & \makecell[c]{$2018$} & \makecell[c]{$14$ \\ $14$} \\
    \hline
    \makecell[c]{KSC\footnotemark[3]} & \makecell[c]{$512\times614$} & \makecell[c]{$18m$} & \makecell[c]{Kennedy Space Center, \\ Florida, USA} & \makecell[c]{HSI} & \makecell[c]{$1996$} & \makecell[c]{$13$} \\
    \hline
    \makecell[c]{Hangzhou~\cite{HZSH} \\ Shanghai~\cite{HZSH}} & \makecell[c]{$590\times230$ \\ $1600\times230$} & \makecell[c]{$30m$} & \makecell[c]{Hangzhou and Shanghai, \\ China} & \makecell[c]{HSI} & \makecell[c]{$2002$} & \makecell[c]{$3$} \\
    \hline
    \makecell[c]{Indian Pines~\cite{IndianPines}} & \makecell[c]{$145\times145$} & \makecell[c]{$20m$} & \makecell[c]{North-Western Indiana, \\ USA} & \makecell[c]{HSI} & \makecell[c]{$1992$} & \makecell[c]{$16$} \\
    \hline
    \makecell[c]{Salinas\footnotemark[4]} & \makecell[c]{$512\times207$} & \makecell[c]{$3.7m$} & \makecell[c]{Salinas Valley, California, \\ USA} & \makecell[c]{HSI} & \makecell[c]{$1998$} & \makecell[c]{$16$} \\
    \hline
    \rowcolor{gray!30!} \multicolumn{7}{c} {\textbf{Aerial and Satellite Optical Image Dataset}}  \\
    \hline
    \makecell[c]{AID~\cite{AID}} & \makecell[c]{$600\times600$} & \makecell[c]{$0.5m\sim8m$} & \makecell[c]{From Google Earth} & \makecell[c]{R-G-B} & \makecell[c]{$2017$} & \makecell[c]{$30$} \\
    \hline
    \makecell[c]{NWPU~\cite{NWPU}} & \makecell[c]{$256\times256$} & \makecell[c]{$0.2m\sim30m$} & \makecell[c]{From Google Earth} & \makecell[c]{R-G-B} & \makecell[c]{$2016$} & \makecell[c]{$45$} \\
    \hline
    \makecell[c]{UC-Merced~\cite{UC-Merced}} & \makecell[c]{$256\times256$} & \makecell[c]{$0.3m$} & \makecell[c]{National Urban Area, \\ USA} & \makecell[c]{R-G-B} & \makecell[c]{$2010$} & \makecell[c]{$21$} \\
    \hline
    \makecell[c]{WHU-RS19~\cite{WHU-RS19-1, WHU-RS19-2}} & \makecell[c]{$600\times600$} & \makecell[c]{$0.5m$} & \makecell[c]{From Google Earth} & \makecell[c]{R-G-B} & \makecell[c]{$2012$} & \makecell[c]{$19$} \\
    \hline
    \makecell[c]{RSSCN7~\cite{RSSCN7}} & \makecell[c]{$400\times400$} & \makecell[c]{-} & \makecell[c]{From Google Earth} & \makecell[c]{R-G-B} & \makecell[c]{$2015$} & \makecell[c]{$7$} \\
    \hline
    \makecell[c]{PatternNet~\cite{PatternNet}} & \makecell[c]{$256\times256$} & \makecell[c]{$0.6m\sim4.7m$} & \makecell[c]{US Cites From Google Earth} & \makecell[c]{R-G-B} & \makecell[c]{$2018$} & \makecell[c]{$38$} \\
    \hline
  \rowcolor{gray!30!} \multicolumn{7}{c} {\textbf{SAR Dataset}}  \\
    \hline
    \makecell[c]{SAMPLE~\cite{SAMPLE} } & \makecell[c]{$128\times128$} & \makecell[c]{$0.3m$} & \makecell[c]{-} & \makecell[c]{SAR} & \makecell[c]{$2019$} & \makecell[c]{$10$} \\
    \hline
    \makecell[c]{FUSAR-Ship~\cite{FUSAR-Ship}} & \makecell[c]{$512\times512$} & \makecell[c]{up to $1m$} & \makecell[c]{From Gaofen-3 (GF-3)} & \makecell[c]{SAR} & \makecell[c]{$2020$} & \makecell[c]{$15$} \\
    \hline
    \makecell[c]{OpenSARShip~\cite{OpenSARShip} \\ OpenSARShip 2.0~\cite{OpenSARShip-2.0}} & \makecell[c]{$30\times30 \sim$ \\ $120\times120$} & \makecell[c]{$\sim10m$} & \makecell[c]{From Sentinel-1} & \makecell[c]{SAR} & \makecell[c]{$2017$ \\ $2017$} & \makecell[c]{$17$ \\ $16$} \\
    \hline\hline  
\end{tabular}}
  \label{Tab5}
\end{table*}
\footnotetext[1]{\url{http://www.ehu.eus/ccwintco/index.php?title=Hyperspectral_Remote_Sensing_Scenes\#Pavia_Centre_and_University}}
\footnotetext[2]{\url{http://www.ehu.eus/ccwintco/index.php?title=Hyperspectral_Remote_Sensing_Scenes\#Botswana}}
\footnotetext[3]{\url{ 
https://www.ehu.eus/ccwintco/index.php?title=Hyperspectral_Remote_Sensing_Scenes\#Kennedy_Space_Center_.28KSC.29}}
\footnotetext[4]{\url{http://www.ehu.eus/ccwintco/index.php?title=Hyperspectral_Remote
 _Sensing_Scenes\#Salinas}}
\section{Benchmark Performance} % 2-3 pages
\label{performance}
In this section, we first introduce the primary datasets commonly used in the field of DA-RS. We then present state-of-the-art methods evaluated on these primary datasets. It is worth noting that our focus is on representative algorithms that are widely adopted for comparative analysis.
\subsection{Benchmark Datasets for DA-RS}
As shown in Tab.~\ref{Tab5}, Tab.~\ref{Tab6}, Tab.~\ref{Tab7} we will summarize the dominant benchmark datasets widely used in DA-RS research. These datasets are designed to support various domain adaptation tasks, including DA-RSCls, DA-RSSeg, and DA-RSDet. For each dataset, we highlight its imaging mode, characteristics, and suitability for evaluation.
\subsubsection{Benchmark Datasets for DA-RSCls} For DA-RSCls, the benchmark datasets (Tab.~\ref{Tab5}) cover hyperspectral (HSI), optical, and SAR imagery, reflecting the task's global classification characteristics. This diversity captures varying spectral, spatial, and modality features, providing a comprehensive basis for evaluating domain adaptation methods.
\par \textbf{The Houston Datasets} (Houston2013~\cite{Debes2014JSTARS}, Houston2018~\cite{Bertrand2018GRSM}) consist of hyperspectral images from 2013 and 2018, collected over the University of Houston area using different sensors. After spectral and spatial alignment, a shared region of size $209\times955$ with 48 spectral bands and 7 consistent land cover categories is extracted from the Houston2013 scene to correspond with the Houston2018 scene.
\par \textbf{Pavia University and Center Datasets (PU and PC)\footnotemark[1]} are acquired by the ROSIS sensor during a flight campaign over Pavia, northern Italy. The number of spectral bands is 102 for PC and 103 for PU. Both datasets contain 9 categories, though their category definitions are not entirely identical. For domain adaptation experiments, only the 7 classes shared between the two datasets are selected to ensure label consistency.
\par \textbf{Botswana Dataset (May, June, and July)\footnotemark[2]} consists of three multitemporal hyperspectral images captured by the NASA EO-1 Hyperion sensor over the Okavango Delta, Botswana, in May, June, and July 2001. Each image contains 9 annotated categories with 242 spectral bands covering the 357–2576 nm range at a 10 nm spectral resolution and a 30 m spatial resolution. After removing uncalibrated and noisy bands, 145 bands are retained for experiments.
\par \textbf{The HyRANK Dataset}~\cite{HyRANK} comprises satellite hyperspectral imagery acquired by the Hyperion sensor (EO-1, USGS), with 176 spectral bands. The dataset includes two labeled scenes, Dioni and Loukia. Both scenes share 12 consistent land cover categories.
\par \textbf{The KSC (Kennedy Space Center) Dataset}\footnotemark[3] is captured by the AVIRIS sensor with 224 spectral bands and 18 m spatial resolution. After removing noisy and water absorption bands, 176 bands remain for use. It contains two spatially disjoint subsets, which share 10 common categories. These two subsets are commonly used as source and target in domain adaptation experiments.
\par \textbf{The Indian Pines Dataset}~\cite{IndianPines} is collected by NASA's AVIRIS sensor. The original data set consists of 224 bands. After preprocessing, the data contain 200 spectral bands, covering a 3 km area with 20-meter spatial resolution. To simulate domain adaptation task, each hyperspectral image is split into two parts by grouping its spectral bands. One subset is designated as the source domain and the other as the target domain, mimicking data acquired from different sensors with varying spectral coverages.
\par \textbf{The Salines Dataset}\footnotemark[4] is acquired by the AVIRIS sensor. After removing 20 water absorption bands, the dataset includes 204 spectral bands at 3.7 m resolution, with 16 labeled land-cover categories used for classification tasks.
\par \textbf{The Hangzhou-Shanghai Datasets}~\cite{HZSH} are acquired by the EO-1 Hyperion sensor in 2002, each retaining 198 spectral bands after preprocessing. Both datasets cover urban and rural regions with 3 labeled land-cover categories. Hangzhou-Shanghai datasets are commonly used for domain adaptation due to their similar geographical scene composition.
\par \textbf{Aerial Image Dataset (AID)}~\cite{AID} contains 30 scene categories and a total of 10,000 large-scale RGB images, each with a size of 600 $\times$ 600 pixels. The number of images per category ranges from 220 to 420. The spatial resolution varies between 0.5 and 8 meters.
\par \textbf{NWPU-RESISC45 Dataset}~\cite{NWPU} comprises 31,500 RGB images across 45 scene categories with sized at 256 $\times$ 256 pixels. Every category includes 700 samples. The spatial resolution spans a wide range from 0.2 to 30 meters.
\par\textbf{UC-Merced Dataset}~\cite{UC-Merced} contains 2,100 RGB images evenly distributed across 21 scene categories, with each category comprising 100 images. All images have a spatial resolution of 0.3 meters and a fixed size of 256 $\times$ 256 pixels.
\par \textbf{WHU-RS19 Dataset}~\cite{WHU-RS19-1, WHU-RS19-2} includes 1,005 high-resolution scene images, grouped into 19 semantic categories. Each category is represented by approximately 50 to 70 samples, with all images uniformly sized at 600 $\times$ 600 pixels.
\par \textbf{RSSCN7 Dataset}~\cite{RSSCN7} consists of 2,800 images across 7 categories. The images are 400 $\times$ 400 pixels and sourced from Google Earth under varying seasons and lighting conditions.
\par \textbf{PatternNet Dataset}~\cite{PatternNet} is a large-scale remote sensing dataset designed for image retrieval, containing 30,400 images across 38 categories. Each category has 800 samples (256 $\times$ 256 pixels) of US cities collected from Google Earth.
\par In domain adaptation experiments, aerial and satellite optical image datasets such as AID~\cite{AID}, NWPU-RESISC45~\cite{NWPU}, UC-Merced~\cite{UC-Merced}, WHU-RS19~\cite{WHU-RS19-1, WHU-RS19-2}, RSSCN7~\cite{RSSCN7}, PatternNet~\cite{PatternNet} are used by selecting shared categories and treating them as source and target domains for each other.
\par \textbf{SAMPLE Dataset}~\cite{SAMPLE} contains 1,345 paired synthetic and measured SAR images across 10 vehicle types, generated under matched imaging parameters from the MSTAR dataset\footnote{https://www.sdms.afrl.af.mil/index.php?collection=mstar}. Synthetic and Measured images are respectively produced using CAD models and real SAR captures. 
\par \textbf{FUSAR-Ship Dataset}~\cite{FUSAR-Ship} comprises over 5,000 high-resolution ship chips cropped from 126 GF-3 scenes. It introduces an extensible SAR ship taxonomy with 15 categories and 98 subcategories, enabling comprehensive development, evaluation, and benchmarking of ship recognition algorithms.
\par \textbf{OpenSARShip Dataset}~\cite{OpenSARShip} is a large-scale SAR ship detection benchmark built from Sentinel-1 imagery. It contains 11,346 ship chips from 41 SAR scenes covering 17 AIS-defined ship types with detailed subtype annotations. This dataset is valuable for detection, fine-grained classification as well as domain adaptation research. 
\par \textbf{OpenSARShip 2.0 Dataset}~\cite{OpenSARShip-2.0} expands on its predecessor by offering 34,528 SAR ship chips extracted from 87 Sentinel-1 images. It offers multiple data formats, including original, grayscale, pseudo-color, and calibrated versions, supporting deeper analysis of ship targets in SAR imagery.
\begin{table*}[!ht]
  \scriptsize
  \centering
  \caption{The representative benchmark datasets along with their detailed information for different DA-RSSeg task.}
  \scalebox{1.0}{
  \begin{tabular}{p{0.16\textwidth}|p{0.18\textwidth}|p{0.07\textwidth}|p{0.18\textwidth} |p{0.11\textwidth}|p{0.09\textwidth}|p{0.07\textwidth}}
    \hline\hline
    \makecell[c]{\textbf{Dataset}} & \makecell[c]{\textbf{Image Size}} & 
    \makecell[c]{\textbf{Resolution}} & \makecell[c]{\textbf{Location/Source}} & \makecell[c]{\textbf{Imaging Mode}} & \makecell[c]{\textbf{Available Date}} & \makecell[c]{\textbf{Categories}}\\
  \hline \hline
    \rowcolor{gray!30!} \multicolumn{7}{c} {\textbf{Aerial and Satellite Optical Image Dataset}}  \\
    \hline
    \makecell[c]{ISPRS Potsdam~\cite{ISPRS} \\ ISPRS Vaihingen~\cite{ISPRS}} & \makecell[c]{$6000\times6000$ \\ $\sim 2000\times2000$} & \makecell[c]{$0.05m$ \\ $0.09m$} & \makecell[c]{Potsdam \& Vaihingen, \\ Germany} & \makecell[c]{R-G-B, IR-R-G, \\ R-G-B-IR, IR-R-G} & \makecell[c]{$2014$} & \makecell[c]{$6$} \\
    \hline
    \makecell[c]{LoveDA Urban~\cite{LoveDA} \\ LoveDA Rural~\cite{LoveDA}} & \makecell[c]{$1024\times1024$} & \makecell[c]{$0.3m$} & \makecell[c]{Nanjing, Changzhou, Wuhan, \\ China} & \makecell[c]{R-G-B} & \makecell[c]{$2021$} & \makecell[c]{$7$} \\
    \hline
    \makecell[c]{CITY-OSM~\cite{CITY-OSM}} & \makecell[c]{$500\times500$} & \makecell[c]{$\sim0.1m$} & \makecell[c]{ Berlin, Chicago, Zurich, \\ Pairs, Tokyo from Google Earth} & \makecell[c]{R-G-B} & \makecell[c]{$2017$} & \makecell[c]{$3$} \\
    \hline
    \makecell[c]{WHU Aerial~\cite{WHU} \\ WHU Satellite~\cite{WHU}} & \makecell[c]{$512\times512$} & \makecell[c]{$0.075m$ \\ $2.7m$} & \makecell[c]{Christchurch, New Zealand \\ Europe, America, New Zealand} & \makecell[c]{R-G-B} & \makecell[c]{$2018$} & \makecell[c]{$2$} \\
    \hline
    \makecell[c]{BLU~\cite{BLU}} & \makecell[c]{$2048\times2048$} & \makecell[c]{$0.8m$} & \makecell[c]{Beijing, China} & \makecell[c]{R-G-B} & \makecell[c]{$2018$} & \makecell[c]{$6$} \\
    \hline
    \makecell[c]{OpenEarthMap~\cite{OpenEarthMap}} & \makecell[c]{$1024\times1024$} & \makecell[c]{$0.25m\sim$ \\ $0.5m$} & \makecell[c]{44 countries on 6 continents} & \makecell[c]{R-G-B} & \makecell[c]{$2023$} & \makecell[c]{$8$} \\
    \hline
    \makecell[c]{DeepGlobe~\cite{DeepGlobe}} & \makecell[c]{$1024\times1024$} & \makecell[c]{$0.5m$} & \makecell[c]{Thailand, Indonesia, India} & \makecell[c]{R-G-B} & \makecell[c]{$2018$} & \makecell[c]{$2$} \\
    \hline
    \makecell[c]{Road Detection~\cite{RD}} & \makecell[c]{$600\times600$} & \makecell[c]{$1.2m$} & \makecell[c]{From Google Earth } & \makecell[c]{R-G-B} & \makecell[c]{$2017$} & \makecell[c]{$2$} \\
    \hline
  \rowcolor{gray!30!} \multicolumn{7}{c} {\textbf{Other Dataset}}  \\
    \hline
    \makecell[c]{C2Seg-AB~\cite{Hong2023RSE} \\ C2Seg-BW~\cite{Hong2023RSE}} & \makecell[c]{$2465\times811$, $886\times1360$ \\ $13474\times8706$, $6225\times8670$} & \makecell[c]{$10m$} & \makecell[c]{Berlin, Augsburg in Germany \\ Beijing, Wuhan in China} & \makecell[c]{HSI, MSI, SAR} & \makecell[c]{$2023$} & \makecell[c]{$13$} \\
    \hline
    \makecell[c]{OpenEarthMap-SAR~\cite{OpenEarthMap-SAR} \\ OpenEarthMap-Optic~\cite{OpenEarthMap-SAR}} & \makecell[c]{$1024\times1024$} & \makecell[c]{$0.15m\sim$ \\ $0.5m$} & \makecell[c]{USA, Japan, France} & \makecell[c]{Optic, SAR} & \makecell[c]{$2025$} & \makecell[c]{$8$} \\
    \hline\hline  
\end{tabular}}
  \label{Tab6}
\end{table*}
\subsubsection{Benchmark Datasets for DA-RSSeg}
For DA-RSSeg, the benchmark datasets (Tab.~\ref{Tab6}) mainly comprise HR aerial and satellite optical imagery, offering fine-grained spatial details suitable for semantic segmentation. These datasets feature diverse geographic regions and acquisition conditions, enabling comprehensive evaluation of domain adaptation methods.
\par \textbf{ISPRS Dataset}~\cite{ISPRS} offers pixel-level annotations across 6 land-cover categories and includes Potsdam and Vaihingen datasets. Potsdam contains 38 VHR orthophotos (6000$\times$6000 pixels) in IR-R-G, R-G-B, and R-G-B-IR modes, while Vaihingen has 33 orthophotos ($\sim$2000$\times$2000 pixels) in IR-R-G mode. The domain adaptation segmentation experiments are conducted between the Potsdam and Vaihingen datasets.
\par \textbf{LoveDA Dataset}~\cite{LoveDA} provides 5,987 VHR remote sensing images (1024$\times$1024) from Nanjing, Changzhou, and Wuhan, covering both urban and rural environments. It spans urban and rural domains to evaluate model generalization across diverse geographic features.
\par \textbf{CITY-OSM Dataset}~\cite{CITY-OSM} is an urban-focused dataset that provides pixel-level annotations for cities such as Berlin, Chicago, Zurich, Paris, and Tokyo. It highlights key urban elements with 3 categories: ``background'', ``road'', and ``building''.
\par \textbf{WHU Dataset}~\cite{WHU} delivers extensive pixel-level annotations for building extraction, including aerial and satellite imagery. The aerial subset contains 8,189 RGB images from New Zealand with 0.3 m resolution. The Satellite II subset comprises 17,388 RGB patches from East Asia with 2.7 m resolution. Both subsets provide diverse building types and are available for domain adaptation experiments.
\par \textbf{BLU Dataset}~\cite{BLU} is collected by the Beijing-2 satellite with a 0.8 m GSD. It has fine-grained pixel-level annotations for 6 land-use categories across diverse urban and rural scenes in Beijing. The dataset comprises 4 large high-resolution tiles (15680$\times$15680 pixels each), cropped into 256 sub-images with nonoverlapping splits for training, validation, and testing.
\par \textbf{OpenEarthMap Dataset}~\cite{OpenEarthMap} is a comprehensive global dataset designed for high-resolution land cover mapping, featuring diverse aerial and satellite imagery from 97 regions across 44 countries. It provides 8-class annotations at 0.25$\sim$0.5 m resolution, supporting the development of models with strong cross-region generalization.
\par \textbf{DeepGlobe Dataset}~\cite{DeepGlobe} is designed to support accurate road extraction. It contains 8,570 high-resolution RGB images (0.5 m/pixel) from Thailand, Indonesia, and India, covering both urban and rural areas.
\par \textbf{Road Detection Dataset}~\cite{RD} comprises 224 VHR Google Earth images (1.2 m/pixel) with manually annotated road segmentation. Each image has a size of at least 600$\times$600 pixels, with road widths of about 12–15 pixels. The presence of complex backgrounds and frequent occlusions makes detection highly challenging.
\par \textbf{C2Seg Dataset}~\cite{Hong2023RSE} is a pioneering large-scale benchmark designed for cross-city multimodal remote sensing semantic segmentation, incorporating hyperspectral, multispectral, and SAR data. It consists of two scenarios: C2Seg-AB sourced from EnMAP, Sentinel-2, and Sentinel-1, and C2Seg-BW sourced from Gaofen-5, Gaofen-6, and Gaofen-3. C2Seg integrates three RS modalities while maintaining a 10 m GSD, and covers 13 categories.
\par \textbf{OpenEarthMap-SAR Dataset}~\cite{OpenEarthMap-SAR} is a large-scale benchmark dataset for global high-resolution land cover mapping using SAR imagery, addressing the lack of SAR-specific benchmarks. It contains 1.5 million segments from 5033 images across 35 regions in Japan, France, and the USA, labeled into 8 categories. Serving as the official ``IEEE GRSS Data Fusion Contest Track I'' dataset, it supports semantic segmentation research under challenging real-world settings.
\begin{table*}[!ht]
  \scriptsize
  \centering
  \caption{The representative benchmark datasets along with their detailed information for different DA-RSDet task.}
  \scalebox{1.0}{
  \begin{tabular}{p{0.16\textwidth}|p{0.18\textwidth}|p{0.07\textwidth}|p{0.19\textwidth} |p{0.09\textwidth}|p{0.09\textwidth}|p{0.07\textwidth}}
    \hline\hline
    \makecell[c]{\textbf{Dataset}} & \makecell[c]{\textbf{Image Size}} & 
    \makecell[c]{\textbf{Resolution}} & \makecell[c]{\textbf{Location/Source}} & \makecell[c]{\textbf{Imaging Mode}} & \makecell[c]{\textbf{Available Date}} & \makecell[c]{\textbf{Categories}}\\
  \hline \hline
    \rowcolor{gray!30!} \multicolumn{7}{c} {\textbf{Aerial and Satellite Optical Image Dataset}}  \\
    \hline
    \makecell[c]{DOTA~\cite{DOTA}} & \makecell[c]{$800\times800$ $\sim 4000\times4000$} & \makecell[c]{$0.1m\sim$ \\ $4.5m$} & \makecell[c]{From Google Earth} & \makecell[c]{R-G-B} & \makecell[c]{$2018$} & \makecell[c]{$15$} \\
    \hline
    \makecell[c]{xView~\cite{xView}} & \makecell[c]{$2500\times2500$ \\ $\sim 5000\times5000$} & \makecell[c]{$0.3m$} & \makecell[c]{From WorldView-3 satellites} & \makecell[c]{R-G-B} & \makecell[c]{$2018$} & \makecell[c]{$60$} \\
    \hline
    \makecell[c]{DIOR~\cite{DIOR}} & \makecell[c]{$800\times800$} & \makecell[c]{$0.5m\sim$ \\ $30m$} & \makecell[c]{From Google Earth} & \makecell[c]{R-G-B} & \makecell[c]{$2018$} & \makecell[c]{$20$} \\
    \hline
    \makecell[c]{UCAS-AOD~\cite{UCAS-AOD}} & \makecell[c]{$1280\times659$} & \makecell[c]{-} & \makecell[c]{From Google Earth} & \makecell[c]{R-G-B} & \makecell[c]{$2014$} & \makecell[c]{$2$} \\
    \hline
    \makecell[c]{CARPK~\cite{CARPK}} & \makecell[c]{$1280\times720$} & \makecell[c]{$0.01m\sim$ \\ $0.04m$} & \makecell[c]{From UAV} & \makecell[c]{R-G-B} & \makecell[c]{$2017$} & \makecell[c]{$1$} \\
    \hline
    \makecell[c]{NWPU VHR-10~\cite{NWPU-VHR-10}} & \makecell[c]{$500\times500 \sim 1100\times1100$} & \makecell[c]{$0.5m\sim$ \\ $2m$} & \makecell[c]{From Google Earth} & \makecell[c]{R-G-B} & \makecell[c]{$2014$} & \makecell[c]{$10$} \\
    \hline
    \makecell[c]{UAVDT~\cite{UAVDT}} & \makecell[c]{$1080\times540$} & \makecell[c]{-} & \makecell[c]{From UAV (DJI Inspire 2)} & \makecell[c]{R-G-B} & \makecell[c]{$2018$} & \makecell[c]{$3$} \\
    \hline
    \makecell[c]{Visdrone~\cite{Visdrone}} & \makecell[c]{$960\times540\sim 1920\times1080$} & \makecell[c]{-} & \makecell[c]{14 different cities, China} & \makecell[c]{R-G-B} & \makecell[c]{$2019\sim2024$} & \makecell[c]{$10$} \\
    \hline
    \makecell[c]{HRRSD~\cite{HRRSD}} & \makecell[c]{$1000\times1000\sim$ \\ $ 2000\times2000$} & \makecell[c]{$0.15m\sim$ \\ $1.2m$} & \makecell[c]{From Google Earth \\ and Baidu Map} & \makecell[c]{R-G-B} & \makecell[c]{$2017$} & \makecell[c]{$13$} \\
    \hline
    \makecell[c]{LEVIR~\cite{LEVIR}} & \makecell[c]{$800\times600$} & \makecell[c]{$0.2m\sim$ \\ $1m$} & \makecell[c]{From Google Earth} & \makecell[c]{R-G-B} & \makecell[c]{$2017$} & \makecell[c]{$3$} \\
    \hline
  \rowcolor{gray!30!} \multicolumn{7}{c} {\textbf{SAR Dataset}}  \\
    \hline
    \makecell[c]{SSDD~\cite{SSDD}} & \makecell[c]{$300\times300 \sim 500\times500$} & \makecell[c]{$1m\sim$ \\ $15m$} & \makecell[c]{From RadarSat-2, \\ TerraSAR-X, Sentinel-1} & \makecell[c]{SAR} & \makecell[c]{$2017$} & \makecell[c]{$1$} \\
    \hline
    \makecell[c]{SAR-Ship-Dataset~\cite{SSD}} & \makecell[c]{$256\times256$} & \makecell[c]{$1.7m\sim$ \\ $15m$} & \makecell[c]{From Gaofen-3 (GF-3), \\ Sentinel-1} & \makecell[c]{SAR} & \makecell[c]{$2019$} & \makecell[c]{$1$} \\
    \hline
    \makecell[c]{FARAD\footnotemark[1] \\ miniSAR\footnotemark[1]} & \makecell[c]{$1300\times580\sim1700\times1850$ \\ $1638\times2510$} & \makecell[c]{$4\quad inch$ \\ $4\quad inch$} & \makecell[c]{University of New Mexico, USA \\ Kirt land Air Force Base, USA} & \makecell[c]{SAR} & \makecell[c]{$2015$ \\ $2005$} & \makecell[c]{$1$ \\ $1$} \\
    \hline
    \makecell[c]{AIR-SARship-1.0~\cite{AIR-SARship-1.0}} & \makecell[c]{$3000\times3000$} & \makecell[c]{$1m, 3m$} & \makecell[c]{From Gaofen-3 (GF-3)} & \makecell[c]{SAR} & \makecell[c]{$2019$} & \makecell[c]{$1$} \\
    \hline \hline  
\end{tabular}}
  \label{Tab7}
\end{table*}
\footnotetext[1]{\url{https://www.sandia.gov/radar/complex-data/}}
\subsubsection{Benchmark Datasets for DA-RSDet} For DA-RSDet, the benchmark datasets (Tab.~\ref{Tab7}) mainly consist of high-resolution aerial and satellite imagery. SAR-based detection datasets are also included. These datasets emphasizes accurate object localization and recognition in complex geospatial scenes, which provides a basis for evaluating DA-RSDet task.
\par  \textbf{DOTA Dataset}~\cite{DOTA} is a large-scale benchmark for object detection in aerial imagery, featuring images from multiple sensors and platforms. It contains diverse objects with varying scales, orientations, and shapes, annotated by experts using arbitrary quadrilaterals. The dataset includes 15 categories, 2,806 images, and over 188k instances. This dataset has been expanded across several versions~\cite{DOTA-1.5, DOTA-2.0}.
\par \textbf{xView Dataset}~\cite{xView} is a large-scale overhead object detection dataset built from WorldView-3 imagery with 0.3 m resolution, covering over 1,400 km$^2$. It contains more than 1 million objects across 60 categories, annotated through a rigorous multi-stage quality control process.
\par \textbf{DIOR Dataset}~\cite{DIOR} is a large-scale benchmark for object detection in remote sensing, containing 23,463 images and 192,472 manually annotated instances across 20 categories. It is notable for its wide range of object size variations, spanning different resolutions and diverse intra- and inter-class scales.
\par \textbf{UCAS-AOD Dataset}~\cite{UCAS-AOD} is constructed from Google Earth imagery and focuses on vehicles and planes with diverse orientations. The vehicle dataset consists of 310 images with 2,819 annotated vehicle instances, while the plane dataset includes 600 images containing 3,210 plane instances. Its balanced orientation distribution and fine-grained annotations for object detection in remote sensing.
\par \textbf{CARPK Dataset}~\cite{CARPK} is the first large-scale UAV-based parking lot dataset for vehicle counting. It contains 1,448 images and 89,774 annotated vehicles. Each image has a resolution of 1280$\times$720 pixels.
\par \textbf{NWPU VHR-10 Dataset}~\cite{NWPU-VHR-10} provides a diverse and challenging benchmark for multi-class geospatial object detection. It covers 10 categories with 3,651 manually annotated instances and offers both positive and negative samples for comprehensive evaluation.
\par \textbf{UAVDT Dataset}~\cite{UAVDT} provides a large-scale benchmark for UAV-based object detection. It consists of 179 videos with over 80,000 frames. An experimental subset of 10,000 images is also available, containing three annotated object categories.
\par \textbf{Visdrone Dataset}~\cite{Visdrone} is a large-scale UAV benchmark comprising over 10,000 frames collected from more than 6 hours of video. It focuses on three common object categories with image resolutions between 540p and 1080p. The dataset is divided into 6,883 training images and 546 testing images.
\par \textbf{HRRSD Dataset}~\cite{HRRSD} ensures balanced distributions across training, validation, and testing subsets. It includes 13 object categories with 26,722 images collected from Google Earth and Baidu Map, at spatial resolutions between 0.15 m and 1.2 m. 
\par \textbf{LEVIR Dataset}~\cite{LEVIR} contains over 22,000 high-resolution Google Earth images at 800$\times$600 pixels with a spatial resolution of 0.2$\sim$1.0 m per pixel. The dataset includes three object categories—airplanes, ships, and oil tanks—with a total of 11,000 annotated bounding boxes. It is widely used in domain adaptation studies by focusing solely on ships and excluding irrelevant categories.
\par \textbf{SAR Ship Detection Dataset (SSDD) Dataset}~\cite{SSDD} contains 1160 SAR images with 2456 ship instances, covering diverse sea states, resolutions, and sensors. It is commonly paired with HRRSD dataset~\cite{HRRSD} to evaluate cross-modality domain adaptation from optical to SAR images.
\par \textbf{SAR-Ship-Dataset}~\cite{SSD}, released by the Chinese Academy of Sciences, is built mainly from GF-3 SAR imagery and provides 43,819 ship chips with diverse backgrounds. It serves as a large-scale benchmark for SAR-based ship detection.
\par \textbf{The miniSAR and FARAD datasets\footnotemark[1]} are SAR image collections acquired by Sandia National Laboratories in 2005 and 2015, respectively. The miniSAR dataset includes 9 images, with 7 for training and 2 for testing. FARAD contains 106 images, split into 78 training and 28 testing images. Both datasets exhibit significant domain differences, which makes them suitable benchmarks for evaluating cross-domain SAR image analysis methods.
\par \textbf{AIR-SARship-1.0}~\cite{AIR-SARship-1.0} contains 31 large-scene SAR images with a total of 879 ships, captured at resolutions of 1 m and 3 m. It is commonly used for domain adaptation for SAR-based ship detection experiments.
\subsection{Comparison and Analysis on Experimental Results}
In this section, we review state-of-the-art methods for DA-RS tasks, with a particular focus on the mainstream benchmark datasets summarized in Tab.\ref{Tab5}, Tab.\ref{Tab6}, and Tab.~\ref{Tab7}. We highlight representative algorithms that are widely used for comparative evaluation. \textbf{Since different papers re-implemented other methods for comparison, the results in the table are reported according to the original papers.}
\par \textbf{In this survey, we focus on the widely studied UDA setting, which mainly refers to the single-source single-target scenario. We also include other settings where relatively systematic studies have been conducted.}
\subsubsection{Benchmark Performance on DA-RSCls}
In DA-RSCls task, the commonly applied evaluation metrics are Overall Accuracy (OA), Average Accuracy (AA), and kappa coefficient ($\kappa$). OA measures the proportion of correctly classified samples over the entire dataset, which reflects the global classification performance. AA computes the mean of per-class accuracies, providing a fairer evaluation when the class distribution is imbalanced. The $\kappa$ considers the agreement between prediction and ground truth beyond chance, offering a more rigorous assessment of classification consistency. For the sake of experimental consistency across different methods, this survey mainly adopts OA as the primary evaluation metric.
\begin{table}[t]
  \scriptsize
  \centering
  \caption{The performance of representative methods on HSI datasets for UDA-RSCls task. The model with the best performance is denoted in bold. Methods marked with “*” are trained with only 5\% of the source samples, while others utilize the full set of labeled samples.}
  \scalebox{1.0}{
  \begin{tabular}{p{0.11\textwidth}|p{0.09\textwidth}|p{0.09\textwidth}
  |p{0.11\textwidth}}
    %\toprule %{c|l|m{7cm}}
    \hline\hline
    \makecell[c]{\textbf{Method}} & \makecell[c]{\textbf{Source}} & \makecell[c]{\textbf{Tasks}} & \makecell[c]{\textbf{OA (\%)}}\\ 
    \hline\hline
  \rowcolor{gray!15!}   \multicolumn{4}{c}{\textbf{The Houston Datasets}} \\
    \hline\hline
    \makecell[c]{TSTnet*~\cite{Zhang2023TNNLS}} & \makecell[c]{TNNLS 2021} & \makecell[c]{\multirow{7}{*}{\shortstack{HS13$\rightarrow$HS18 \\ / \\ HS18$\rightarrow$HS13}}} & \makecell[c]{75.34/\textbf{88.38}} \\
    \makecell[c]{S$^4$DL*~\cite{Feng2025TNNLS}} & \makecell[c]{TNNLS 2025} & & \makecell[c]{72.10/-} \\
    \makecell[c]{DAN\_MFAC*~\cite{Wu2025KBS}} & \makecell[c]{KBS 2025} & & \makecell[c]{\textbf{79.88}/81.50} \\
    \makecell[c]{TAADA~\cite{Huang2022TGRS}} & \makecell[c]{TGRS 2022} & & \makecell[c]{81.63/-} \\
    \makecell[c]{SDENet~\cite{Zhang2023TIP}} & \makecell[c]{TGRS 2023} & & \makecell[c]{79.96/-} \\
    \makecell[c]{MLUDA~\cite{Cai2024TGRS}} & \makecell[c]{TGRS 2024} & & \makecell[c]{76.64/-} \\
    \makecell[c]{S$^{2}$AMSnet~\cite{ChenXi2024TGRS}} & \makecell[c]{TGRS 2024} & & \makecell[c]{\textbf{82.70}/-} \\
    \hline
    \rowcolor{gray!15!} \multicolumn{4}{c}{\textbf{The HyRANK Dataset}} \\
    \hline\hline
    \makecell[c]{TSTnet*~\cite{Zhang2023TNNLS}} & \makecell[c]{TNNLS 2021} & \makecell[c]{\multirow{4}{*}{\shortstack{Dioni$\rightarrow$Loukia \\ / \\ Loukia$\rightarrow$Dioni}}} & \makecell[c]{63.31/-} \\
    \makecell[c]{S$^4$DL*~\cite{Feng2025TNNLS}} & \makecell[c]{TNNLS 2025} & & \makecell[c]{\textbf{65.00}/-} \\
    \makecell[c]{DAN\_MFAC*~\cite{Wu2025KBS}} & \makecell[c]{KBS 2025} & & \makecell[c]{64.24/\textbf{68.55}} \\
    \makecell[c]{TAADA~\cite{Huang2022TGRS}} & \makecell[c]{TGRS 2022} & & \makecell[c]{\textbf{68.03}/-} \\
    \hline
    \rowcolor{gray!15!} \multicolumn{4}{c}{\textbf{The Hangzhou-Shanghai Dataset}} \\
    \hline\hline
    \makecell[c]{TSTnet*~\cite{Zhang2023TNNLS}} & \makecell[c]{TNNLS 2021} & \makecell[c]{\multirow{5}{*}{\shortstack{HZ$\rightarrow$SH \\ / \\ SH$\rightarrow$HZ}}} & \makecell[c]{90.36/-} \\
    \makecell[c]{S$^4$DL*~\cite{Feng2025TNNLS}} & \makecell[c]{TNNLS 2025} & & \makecell[c]{\textbf{92.40}/-} \\
    \makecell[c]{TAADA~\cite{Huang2022TGRS}} & \makecell[c]{TGRS 2022} & & \makecell[c]{\textbf{94.17}/-} \\
    \makecell[c]{MLUDA~\cite{Cai2024TGRS}} & \makecell[c]{TGRS 2024} & & \makecell[c]{-/\textbf{92.15}} \\
    \makecell[c]{S$^{2}$AMSnet~\cite{ChenXi2024TGRS}} & \makecell[c]{TGRS 2024} & & \makecell[c]{90.62/-} \\
    \hline\hline
   \rowcolor{gray!15!}  \multicolumn{4}{c}{\textbf{The Pavia University and Center Datasets}} \\
    \hline\hline
    \makecell[c]{DAN\_MFAC*~\cite{Wu2025KBS}} & \makecell[c]{KBS 2025} & \makecell[c]{\multirow{7}{*}{\shortstack{PU$\rightarrow$PC \\ / \\ PC$\rightarrow$PU}}} & \makecell[c]{75.56/69.78} \\
    \makecell[c]{ADA-Net~\cite{Yu2022GRSL}} & \makecell[c]{TGRS 2020} & & \makecell[c]{88.25/-} \\
    \makecell[c]{JDA~\cite{Miao2021JSTARs}} & \makecell[c]{JSTARs 2021} & & \makecell[c]{83.55/-} \\
    \makecell[c]{UDACA~\cite{Yu2022GRSL}} & \makecell[c]{GRSL 2022} & & \makecell[c]{\textbf{92.87}/\textbf{78.04}} \\
    \makecell[c]{SDENet~\cite{Zhang2023TIP}} & \makecell[c]{TGRS 2023} & & \makecell[c]{81.94/-} \\
    \makecell[c]{MLUDA~\cite{Cai2024TGRS}} & \makecell[c]{TGRS 2024} & & \makecell[c]{91.26/-} \\
    \makecell[c]{S$^{2}$AMSnet~\cite{ChenXi2024TGRS}} & \makecell[c]{TGRS 2024} & & \makecell[c]{86.18/-} \\
    \hline \hline
    \rowcolor{gray!15!} \multicolumn{4}{c}{\textbf{The Botswana Datasets}} \\
    \hline\hline
    \makecell[c]{CDA~\cite{Liu2021TGRS}} & \makecell[c]{TGRS 2020} & \makecell[c]{\multirow{6}{*}{\shortstack{May$\rightarrow$June/ \\ June$\rightarrow$May/ \\ May$\rightarrow$July/ \\ July$\rightarrow$May/ \\ June$\rightarrow$July/ \\
    July$\rightarrow$June }}} & \makecell[c]{92.62/87.85/91.27/ \\ 79.78/95.05/93.14} \\ 
    \makecell[c]{JCGNN~\cite{Wang2021JSTARs}} & \makecell[c]{JSTARs 2021} & & \makecell[c]{\textbf{94.11}/\textbf{92.91}/91.11/ \\ \textbf{83.54}/\textbf{95.85}/\textbf{94.67}} \\
    \makecell[c]{CMC~\cite{Wei2021JSTARs}} & \makecell[c]{JSTARs 2021} & & \makecell[c]{91.41/85.25/\textbf{91.70}/ \\ 83.29/95.78/93.96} \\
    \makecell[c]{JDA~\cite{Miao2021JSTARs}} & \makecell[c]{JSTARs 2021} & & \makecell[c]{89.69/81.39/90.30/ \\ 80.63/90.81/91.82}  \\
    \hline \hline
  \end{tabular}}
  \label{Tab8}
\end{table}
\par \textbf{Benchmark Performance on HSI datasets.} Tab.~\ref{Tab8} summarizes the performance of representative UDA-RSCls methods on the HSI benchmark datasets listed in Tab.~\ref{Tab5}.
\par In the Houston dataset, Houston2013 and Houston2018 are respectively served as source and target sets. 7 consistent land cover categories are extracted from the Houston2013 scene to correspond with the Houston2018 scene. The 2 domain adaptation tasks are listed as follows:
\begin{itemize}
\item Adapt Houston2013 to Houston2018 (HS13 $\rightarrow$ HS18).
\item Adapt Houston2018 to Houston2013 (HS18 $\rightarrow$ HS13).
\end{itemize} 
\par In the HyRANK dataset, Dioni and Loukia are respectively served as source and target sets. 12 consistent land cover categories between Dioni and Loukia are extracted. The 2 domain adaptation tasks are listed as follows:
\begin{itemize}
\item Adapt Dioni to Loukia (Dioni $\rightarrow$ Loukia).
\item Adapt Loukia to Dioni (Loukia $\rightarrow$ Dioni).
\end{itemize}
\par In the Hangzhou-Shanghai dataset, Hangzhou and Shanghai are respectively served as source and target sets. 3 consistent land cover categories between Hangzhou and Shanghai are extracted. The 2 domain adaptation tasks are listed as follows:
\begin{itemize}
\item Adapt Hangzhou to Shanghai (HZ $\rightarrow$ SH).
\item Adapt Shanghai to Hangzhou (SH $\rightarrow$ HZ).
\end{itemize} 
\par In the Pavia University and Center Datasets, Pavia University and Pavia Center are respectively served as source and target sets. 7 consistent land cover categories between Pavia University and Pavia Center are extracted. The 2 domain adaptation tasks are listed as follows:
\begin{itemize}
\item Adapt Pavia University to Pavia Center (PU $\rightarrow$ PC).
\item Adapt Pavia Center to Pavia University (PC $\rightarrow$ PU).
\end{itemize}
\begin{table*}[ht]
  \scriptsize
  \centering
  \caption{The performance of representative methods on aerial and satellite optical datasets for UDA-RSCls task. ``5/6/8-c'' denotes that 5, 6, or 8 common categories are extracted from the selected datasets. OA is adopted as metric.}
  \scalebox{1.1}{
  \begin{tabular}{c|c|c|c|c|c|c|c}
  %\begin{tabular}{p{0.05\textwidth}|p{0.06\textwidth}|p{0.06\textwidth}|p{0.06\textwidth}|p{0.06\textwidth}|p{0.06\textwidth}|p{0.06\textwidth}|p{0.06\textwidth}|p{0.06\textwidth}|p{0.06\textwidth}|p{0.06\textwidth}}
  \hline \hline
  \multirow{4}{*}{\makecell[c]{\textbf{Tasks}}} & \multicolumn{7}{c}{\textbf{UDA-RSCls Methods on Aerial and Satellite Optical Image Datasets}}
  \\ \cline{2-8}
  {} & \makecell[c]{5-c} & \multicolumn{5}{c|}{6-c} & \makecell[c]{8-c} \\ \cline{2-8}
  {} & \makecell[c]{DFENet~\cite{ZhangXufei2022TGRS}} & \makecell[c]{DATSNET~\cite{ZhengZhendong2021TGRS}} & \makecell[c]{MRDAN~\cite{Niu2022TGRS}} & \makecell[c]{AMRAN~\cite{Zhu2022TGRS}} & \makecell[c]{SRKT~\cite{Zhao2023TGRS}} & \makecell[c]{DDCI~\cite{Zhao2025TGRS}} & \makecell[c]{ADA-DDA~\cite{Yang2022TGRS}} \\ 
  {} & \makecell[c]{TGRS 2022} & \makecell[c]{TGRS 2021} & \makecell[c]{TGRS 2022} & \makecell[c]{TGRS 2022} & \makecell[c]{TGRS 2023} & \makecell[c]{TGRS 2025} & \makecell[c]{TGRS 2022} \\ \hline
  \makecell[c]{U$\rightarrow$A} & - & 76.3 & 90.8 & 84.8 & 95.3 & 98.6 & 77.8
  \\ \cline{1-8}
  \makecell[c]{A$\rightarrow$U} & - & 82.6 & 89.1 & 88.2 & 84.8 & 89.3 & 87.5
  \\ \cline{1-8}
  \makecell[c]{U$\rightarrow$W} & - & 92.1 & 96.1 & 89.6 & 94.0 & 96.5 & - 
  \\ \cline{1-8}
  \makecell[c]{W$\rightarrow$U} & - & 83.1 & 89.1 & 85.2 & 88.3 & 96.5 & - 
  \\ \cline{1-8}
  \makecell[c]{A$\rightarrow$W} & - & 99.4 & 99.8 & 99.7 & 98.7 & 99.0 & -
  \\ \cline{1-8}
  \makecell[c]{W$\rightarrow$A} & - & 95.8 & 96.4 & 94.8 & 97.0 & 98.3 & -
  \\ \cline{1-8}
  \makecell[c]{U$\rightarrow$N} & - & - & - & - & - & - & 74.8 
  \\ \cline{1-8}
  \makecell[c]{N$\rightarrow$U} & - & - & - & - & - & - & 89.6 
  \\ \cline{1-8}
  \makecell[c]{A$\rightarrow$N} & - & - & - & - & - & - & 90.7 
  \\ \cline{1-8}
  \makecell[c]{N$\rightarrow$A} & - & - & - & - & - & - & 92.0
  \\ \cline{1-8}
  %\makecell[c]{W$\rightarrow$N} & - & - & - & - & - & - & -
  %\\ \cline{1-8}
  %\makecell[c]{N$\rightarrow$W} & - & - & - & - & - & - & -
  %\\ \cline{1-8}
  \makecell[c]{U$\rightarrow$R} & 85.1 & 60.6 & 78.6 & 69.0 & 74.5 & 75.3 & -
  \\ \cline{1-8}
  \makecell[c]{R$\rightarrow$U} & 88.5 & 88.7 & 90.9 & 87.6 & 89.2 & 91.3 & -
  \\ \cline{1-8}
  \makecell[c]{A$\rightarrow$R} & 90.7 & 77.0 & 85.1 & 81.0 & 85.3 & 82.0 & -
  \\ \cline{1-8}
  \makecell[c]{R$\rightarrow$A} & 94.1 & 90.5 & 96.9 & 94.5 & 98.4 & 97.7 & -
  \\ \cline{1-8}
  \makecell[c]{N$\rightarrow$R} & 92.6 & - & - & - & - & - & -
  \\ \cline{1-8}
  \makecell[c]{R$\rightarrow$N} & 95.0 & - & - & - & - & - & -
  \\ \cline{1-8}
  \makecell[c]{W$\rightarrow$R} & - & 76.0 & 87.0 & 79.5 & 85.6 & 79.2 & -
  \\ \cline{1-8}
  \makecell[c]{R$\rightarrow$W} & - & 95.9 & 97.4 & 94.6 & 96.8 & 98.4 & -
  \\ \cline{1-8}
  \hline \hline
   \end{tabular}
  }
  \label{Tab9}
\end{table*}
\par In the Botswana dataset, three subsets (May, June, July) are used as source and target domains with 9 consistent land cover categories. Each subset can serve as source while the other two act as target, resulting in 6 domain adaptation tasks.
\begin{itemize}
\item Adapt May to June (May $\rightarrow$ June).
\item Adapt June to May (June $\rightarrow$ May).
\item Adapt May to July (May $\rightarrow$ July).
\item Adapt July to May (July $\rightarrow$ May).
\item Adapt June to July (June $\rightarrow$ July).
\item Adapt July to June (July $\rightarrow$ June).
\end{itemize}
\par From the performance comparison in Tab.~\ref{Tab8}, we conduct an in-depth analysis from the following perspectives. (1) TSTnet~\cite{Zhang2023TNNLS}, S$^4$DL~\cite{Feng2025TNNLS} and DAN\_MFAC~\cite{Wu2025KBS} investigate the challenging scenario of training with only 5\% of source samples, and demonstrate competitive performance on the ``Houston'', ``HyRANK'', ``Hangzhou-Shanghai'', and ``Pavia University and Center'' datasets. In contrast, the other methods do not follow this restricted setting, enabling a more direct comparison under full supervision. TAADA~\cite{Huang2022TGRS} achieves the highest performance on the adaptation task of the HyRANK dataset. On the ``HZ$\rightarrow$SH'' and ``SH$\rightarrow$HZ'' tasks, TAADA~\cite{Huang2022TGRS} and MLUDA~\cite{Cai2024TGRS} achieve the best results, respectively. UDACA~\cite{Yu2022GRSL} represents the state-of-the-art on the Pavia University and Center datasets. For the Botswana dataset, JCGNN~\cite{Wang2021JSTARs} demonstrates clear superiority, securing the best results on 5 out of 6 tasks. (2) In the DA-RSCls task on HSI datasets, experimental configurations differ considerably across methods. Therefore, we recommend checking each paper's implementation details carefully when following these methods. (3) The preprocessing stage of hyperspectral data plays a crucial role. Therefore, the comparison of methods is not merely about deep learning techniques, but rather constitute a comprehensive evaluation encompassing preprocessing, hyperparameter settings, and other relevant factors. (4) Some datasets have only been explored in isolated methods, and systematic or continuous explorations are relatively limited. Therefore, the results on all HSI datasets are not presented in the Tab.~\ref{Tab8}.
\par \textbf{Benchmark Performance on Aerial and Satellite Optical Image datasets.} Tab.~\ref{Tab9} summarizes the performance of representative UDA-RSCls methods on the aerial and satellite optical benchmark datasets listed in Tab.~\ref{Tab5}.
\par As shown in Tab.\ref{Tab9}, we select 5 widely applied datasets, namely AID (A)\cite{AID}, UC-Merced (U)\cite{UC-Merced}, WHU-RS19 (W)\cite{WHU-RS19-1}, NWPU-RESISC45 (N)~\cite{NWPU} and RSSCN7 (R)~\cite{RSSCN7} to construct the following UDA-RSCls tasks. Since different methods employ different combinations of datasets in their experiments, the number of shared categories varies. Therefore, we present a comprehensive summary in Tab.~\ref{Tab9}.
\begin{itemize}
\item Adaptation between UC-Merced and AID \\ (A $\rightarrow$ U \& U $\rightarrow$ A).
\item Adaptation between UC-Merced and WHU-RS19 \\ (U $\rightarrow$ W \& W $\rightarrow$ U).
\item Adaptation between UC-Merced and NWPU-RESISC45 \\ (U $\rightarrow$ N \& N $\rightarrow$ U).
\item Adaptation between AID and WHU-RS19 \\ (A $\rightarrow$ W \& W $\rightarrow$ A).
\item Adaptation between AID and NWPU-RESISC45 \\ (A $\rightarrow$ N \& N $\rightarrow$ A).
%\item Adaptation between WHU-RS19 and NWPU-RESISC45 \\ (W $\rightarrow$ N \& N $\rightarrow$ W).
\item Adaptation between UC-Merced and RSSCN7 \\ (U $\rightarrow$ R \& R $\rightarrow$ U).
\item Adaptation between AID and RSSCN7 \\ (A $\rightarrow$ R \& R $\rightarrow$ A).
\item Adaptation between NWPU-RESISC45 and RSSCN7 \\ (N $\rightarrow$ R \& R $\rightarrow$ N).
\item Adaptation between WHU-RS19 and RSSCN7 \\ (W $\rightarrow$ R \& R $\rightarrow$ W).
\end{itemize}
\par From the performance comparison in Tab.~\ref{Tab9}, we will analyze in the following points. (1) DATSNET~\cite{ZhengZhendong2021TGRS}, MRDAN~\cite{Niu2022TGRS}, AMRAN~\cite{Zhu2022TGRS}, SRKT~\cite{Zhao2023TGRS}, DDCI~\cite{Zhao2025TGRS} adopt similar task configurations and implementation details. Among them, DDCI~\cite{Zhao2025TGRS} achieves the best overall performance. (2) In UDA-RSCls tasks, adversarial learning mechanism are often employed for feature alignment, but they may be less economical for this setting. Since classification relies on global semantic features rather than fine-grained local details, adversarial training can introduce unnecessary complexity and instability. In contrast, more direct distribution measurement algorithm can provide a simpler and more stable way to reduce the domain gap while maintaining efficiency.
\begin{table}[t]
  \scriptsize
  \centering
  \caption{The performance of representative methods on SAR datasets for UDA-RSCls task. The model with the best performance is denoted in bold. }
  \scalebox{1.0}{
  \begin{tabular}{p{0.11\textwidth}|p{0.09\textwidth}|p{0.09\textwidth}
  |p{0.11\textwidth}}
    %\toprule %{c|l|m{7cm}}
    \hline\hline
    \makecell[c]{\textbf{Method}} & \makecell[c]{\textbf{Source}} & \makecell[c]{\textbf{Tasks}} & \makecell[c]{\textbf{OA (\%)}} \\ 
    \hline\hline
    \rowcolor{gray!15!} \multicolumn{4}{c}{\textbf{The SAMPLE Dataset}} \\
    \hline\hline
    \makecell[c]{PFDA~\cite{Chen2022GRSL}} & \makecell[c]{GRSL 2022} & \makecell[c]{\multirow{2}{*}{S$\rightarrow$M}} & \makecell[c]{97.81} \\
    \makecell[c]{VSFA~\cite{Zhang2023TGRS}} & \makecell[c]{TGRS 2023} & & \makecell[c]{\textbf{98.18}} \\
    \hline
    \rowcolor{gray!15!} \multicolumn{4}{c}{\textbf{The Hangzhou-Shanghai Dataset}} \\
    \hline\hline
    \makecell[c]{TSIC~\cite{Zhao2022GRSL}} & \makecell[c]{GRSL 2022} & \makecell[c]{\multirow{2}{*}{F$\rightarrow$O / O$\rightarrow$F}} & \makecell[c]{71.4/74.2} \\
    \makecell[c]{DST~\cite{Zhao2024GRSL}} & \makecell[c]{GRSL 2024} & & \makecell[c]{\textbf{75.6}/\textbf{77.5}} \\
    \hline \hline
  \end{tabular}}
  \label{Tab10}
\end{table}
\par \textbf{Benchmark Performance on SAR datasets.} Although research on UDA-RSCls with SAR datasets is relatively limited, we still summarize representative UDA-RSCls methods on the SAR benchmark datasets in Tab.~\ref{Tab10}. 
\par In the SAMPLE~\cite{SAMPLE} dataset, the UDA-RSCls task is carried out between the synthetic and measured subsets, with the synthetic subset serving as the source and the measured subset as the target.
\begin{itemize}
\item Adapt synthetic to measured (S $\rightarrow$ M).
\end{itemize}
\par The reverse setting (M $\rightarrow$ S) is not considered since it lacks practical relevance.
\par In the UDA SAR ship classification task, FUSAR-Ship~\cite{FUSAR-Ship} and OpenSARShip~\cite{OpenSARShip} are utilized. To ensure consistency, five shared classes are selected. The tasks are formulated by adapting across the two datasets.
\begin{itemize}
\item Adapt FUSAR-Ship to OpenSARShip (F $\rightarrow$ O).
\item Adapt OpenSARShip to FUSAR-Ship (O $\rightarrow$ F).
\end{itemize}
\par As shown in Tab.~\ref{Tab10}, PFDA~\cite{Chen2022GRSL}, VSFA~\cite{Zhang2023TGRS}, TSIC~\cite{Zhao2022GRSL} and DST~\cite{Zhao2024GRSL} provide insights into the domain adaptation for SAR image classification. Among them, VSFA~\cite{Zhang2023TGRS} and DST~\cite{Zhao2024GRSL} achieve the best performance on their respective experimental settings. 
\subsubsection{Benchmark Performance on DA-RSSeg}
In DA-RSSeg task, model performance is evaluated using $IoU/mIoU$ and $F1/mF$-score. For class $i$, $IoU_{i} = tp_{i} / (tp_{i} + fp_{i} + fn_{i})$, where $tp_{i}$, $fp_{i}$, and $fn_{i}$ denote true positive, false positive, and false negative, respectively. $mIoU$ is the average across all classes. The $F1$-score is defined as $2 \times Precision \times Recall / (Precision + Recall)$, and $mF$-score is its mean over all classes. To ensure fair and consistent comparison across methods, this survey primarily reports results using $mIoU$ as the main evaluation metric.
\par As previously noted, most recent DA-RSSeg methods focus on HR aerial and satellite optical images. In this survey, we systematically compare their performance on two notable benchmark datasets, ISPRS~\cite{ISPRS} and LoveDA~\cite{LoveDA} datasets. The results are summarized in Tab.~\ref{Tab11}. 
\begin{table}[t]
  \scriptsize
  \centering
  \caption{The performance of representative methods for UDA-RSSeg task. $mIoU$ (\%) is adopted as metric. The model with the best performance is denoted in bold.}
  \scalebox{0.9}{
  \begin{tabular}
  %{p{0.107\textwidth}|p{0.068\textwidth}|p{0.058\textwidth}|p{0.058\textwidth}|p{0.058\textwidth}|p{0.058\textwidth}}
    {c|c|cccc}
    \hline\hline
   \makecell[c]{\multirow{2}{*}{\textbf{Method}}} & \makecell[c]{\multirow{2}{*}{\textbf{Source}}} & \multicolumn{4}{c}{\textbf{Tasks on ISPRS dataset (w/ ``Clutter'')}} \\ \cline{3-6}
  {} & {} & \makecell[c]{P $\rightarrow$ V\textcircled{1}} & \makecell[c]{V $\rightarrow$ P\textcircled{1}} & \makecell[c]{P $\rightarrow$ V\textcircled{2}} & \makecell[c]{V $\rightarrow$ P\textcircled{2}} \\
    \hline\hline
    \makecell[c]{CSLG~\cite{ZhangBo2022TGRS}} & \makecell[c]{TGRS 2021} & \makecell[c]{52.03} & \makecell[c]{47.87} & \makecell[c]{48.06} & \makecell[c]{43.17} \\ \hline
    \makecell[c]{IterDANet~\cite{Cai2022TGRS}} & \makecell[c]{TGRS 2022} & \makecell[c]{-} & \makecell[c]{-} & \makecell[c]{42.20} & \makecell[c]{40.80} \\ \hline
    \makecell[c]{DNT~\cite{Chen2022JAG}} & \makecell[c]{JAG 2022} & \makecell[c]{54.19} & \makecell[c]{43.74} & \makecell[c]{52.60} & \makecell[c]{41.45} \\ \hline
    \makecell[c]{STADA~\cite{Liang2023GRSL}} & \makecell[c]{GRSL 2023} & \makecell[c]{48.05} & \makecell[c]{46.69} & \makecell[c]{-} & \makecell[c]{-} \\ \hline
    \makecell[c]{MMDANet~\cite{Zhou2023TGRS}} & \makecell[c]{TGRS 2023} & \makecell[c]{51.68} & \makecell[c]{47.59} & \makecell[c]{45.93} & \makecell[c]{41.20} \\ \hline
    \makecell[c]{FGUDA~\cite{Wang2023JSTARs}} & \makecell[c]{JSTARs 2023} & \makecell[c]{53.63} & \makecell[c]{50.94} & \makecell[c]{50.18} & \makecell[c]{45.86} \\ \hline
    \makecell[c]{ResiDualGAN~\cite{Zhao2023RS}} & \makecell[c]{RS 2023} & \makecell[c]{55.83} & \makecell[c]{-} & \makecell[c]{46.62} & \makecell[c]{-} \\ \hline
    % \makecell[c]{MemoryAdaptNet~\cite{Zhu2023TGRS}} & \makecell[c]{TGRS 2023} & \makecell[c]{56.05} & \makecell[c]{49.82} & \makecell[c]{46.88} & \makecell[c]{-} \\ \hline
    \makecell[c]{JDAF~\cite{Huang2024TGRS}} & \makecell[c]{TGRS 2024} & \makecell[c]{55.52} & \makecell[c]{51.30} & \makecell[c]{52.10} & \makecell[c]{45.79} \\ \hline
    \makecell[c]{CPCA~\cite{Zhu2024TGRS}} & \makecell[c]{TGRS 2024} & \makecell[c]{60.75} & \makecell[c]{50.72} & \makecell[c]{47.67} & \makecell[c]{-} \\ \hline
    \makecell[c]{ST-DASegNet~\cite{Zhao2024JAG}} & \makecell[c]{JAG 2024} & \makecell[c]{64.33} & \makecell[c]{59.65} & \makecell[c]{55.16} & \makecell[c]{56.86} \\ \hline
    \makecell[c]{PFM-JONet~\cite{Lyu2025TGRS}} & \makecell[c]{TGRS 2025} & \makecell[c]{66.86} & \makecell[c]{61.21} & \makecell[c]{56.61} & \makecell[c]{57.60} \\ \hline
    {DDF~\cite{Ran2024TGRS}} & \makecell[c]{TGRS 2024} & \makecell[c]{68.69} & \makecell[c]{63.93} & \makecell[c]{66.81} & \makecell[c]{\textbf{60.18}} \\ \hline
    \makecell[c]{MEBS~\cite{Li2025TGRS}} & \makecell[c]{TGRS 2025} & \makecell[c]{\textbf{72.28}} & \makecell[c]{\textbf{67.06}} & \makecell[c]{\textbf{69.77}} & \makecell[c]{-} \\
    \hline\hline
   \makecell[c]{\multirow{2}{*}{\textbf{Method}}} & \makecell[c]{\multirow{2}{*}{\textbf{Source}}} & \multicolumn{4}{c}{\textbf{Tasks on ISPRS dataset (w/o ``Clutter'')}} \\ \cline{3-6}
  {} & {} & \makecell[c]{P $\rightarrow$ V\textcircled{1}} & \makecell[c]{V $\rightarrow$ P\textcircled{1}} & \makecell[c]{P $\rightarrow$ V\textcircled{2}} & \makecell[c]{V $\rightarrow$ P\textcircled{2}} \\ \hline
  \makecell[c]{MIDANet~\cite{ChenHongyu2022TGRS}} & \makecell[c]{TGRS 2022} & \makecell[c]{62.25} & \makecell[c]{-} & \makecell[c]{59.82} & \makecell[c]{59.01} \\ \hline
  \makecell[c]{MBATA-GAN~\cite{Ma2023TGRS}} & \makecell[c]{TGRS 2023} & \makecell[c]{63.50} & \makecell[c]{-} & \makecell[c]{52.31} & \makecell[c]{48.42} \\ \hline
  \makecell[c]{DeGLGAN~\cite{Ma2024TGRS}} & \makecell[c]{TGRS 2024} & \makecell[c]{\textbf{68.09}} & \makecell[c]{-} & \makecell[c]{\textbf{63.09}} & \makecell[c]{\textbf{62.17}} \\
    \hline\hline
    \makecell[c]{\multirow{2}{*}{\textbf{Method}}} & \makecell[c]{\multirow{2}{*}{\textbf{Source}}} & \multicolumn{4}{c}{\textbf{Tasks on LoveDA dataset}} \\ \cline{3-6}
  {} & {} & \multicolumn{2}{c}{U $\rightarrow$ R} & \multicolumn{2}{c}{R $\rightarrow$ U} \\ \hline
  \makecell[c]{DCA~\cite{Wu2022TGRS}} & \makecell[c]{TGRS 2022} & \multicolumn{2}{c}{45.17} & \multicolumn{2}{c}{46.36}  \\ \hline
   \makecell[c]{JDAF~\cite{Huang2024TGRS}} & \makecell[c]{TGRS 2024} & \multicolumn{2}{c}{47.31} & \multicolumn{2}{c}{47.12}  \\ \hline
  \makecell[c]{ST-DASegNet~\cite{Zhao2024JAG}} & \makecell[c]{JAG 2024} & \multicolumn{2}{c}{\textbf{50.08}} & \multicolumn{2}{c}{50.28}  \\ \hline
  {DeGLGAN~\cite{Ma2024TGRS}} & \makecell[c]{TGRS 2024} & \multicolumn{2}{c}{38.58} & \multicolumn{2}{c}{\textbf{55.50}} \\
  \hline \hline
  \end{tabular}}
  \label{Tab11}
\end{table}
\par ISPRS~\cite{ISPRS} is the most widely used benchmark dataset for the UDA-RSSeg task. It provides comprehensive coverage for DA tasks, with Potsdam offering large-scale multi-modal VHR images (IR-R-G and R-G-B) and Vaihingen contributing IR-R-G images. Existing methods generally follow two evaluation settings on this dataset: some methods consider all 6 categories including ``Clutter'' (w/ ``Clutter''), while others adopt a 5-class setting that excludes it (w/o ``Clutter''). The DA tasks on this dataset are listed as follows. 
\begin{itemize}
\item Adapt Potsdam IR-R-G to Vaihingen IR-R-G \\ (P $\rightarrow$ V\textcircled{1}).
\item Adapt Vaihingen IR-R-G to Potsdam IR-R-G \\ (V $\rightarrow$ P\textcircled{1}).
\item Adapt Potsdam R-G-B to Vaihingen IR-R-G \\ (P $\rightarrow$ V\textcircled{2}).
\item Adapt Vaihingen IR-R-G to Potsdam R-G-B \\ (V $\rightarrow$ P\textcircled{2}).
\end{itemize}
\par LoveDA~\cite{LoveDA} is another notable benchmark dataset for DA-RSSeg task. It supports two cross-domain tasks (listed as follows) and its test set is evaluated through an online server~\footnote{https://github.com/Junjue-Wang/LoveDA}.
\begin{itemize}
\item Adapt Urban to Rural (U $\rightarrow$ R).
\item Adapt Rural to Urban (R $\rightarrow$ U).
\end{itemize}
\par As shown in Tab.~\ref{Tab11}, the following observations can be made: (1) On the ISPRS dataset, MEBS~\cite{Li2025TGRS} achieves the best performance under the ``w/ Clutter'' setting, and DeGLGAN~\cite{Ma2024TGRS} outperforms all other methods in the ``w/o Clutter'' configuration. In addition, DeGLGAN shows strong results on the ``R $\rightarrow$ U'' task of the LoveDA dataset, while ST-DASegNet~\cite{Zhao2024JAG} attains the best performance on the ``U $\rightarrow$ R'' task. (2) PFM-JONet~\cite{Lyu2025TGRS} does not reach state-of-the-art performance, highlighting that methods based on foundation models do not always surpass specialized models on specific downstream tasks. (3) Class-wise $IoU$ results in UDA-RSSeg task provide a more detailed view of model performance across different categories, which can be referred to original papers for further analysis. (4) It should be noted that the LoveDA leaderboard contains many methods, but not all of them are associated with published papers.
\begin{table}[t]
  \scriptsize
  \centering
  \caption{The performance of representative methods for UDA-RSDet task. $mAP_{50}$ (\%) is adopted as the metric.}
  \scalebox{0.9}{
  \begin{tabular}
    {c|c|cc}
    \hline\hline
    \makecell[c]{\multirow{2}{*}{\textbf{Method}}} & \makecell[c]{\multirow{2}{*}{\textbf{Source}}} & \multicolumn{2}{c}{\textbf{Tasks on xView and DOTA datasets}} \\ \cline{3-4}
  {} & {} & \multicolumn{2}{c}{xView $\rightarrow$ DOTA-3} \\ \hline
  \makecell[c]{FADA~\cite{Xu2022TGRS}} & \makecell[c]{TGRS 2022} & \multicolumn{2}{c}{52.5}  \\ \hline
  \makecell[c]{RST~\cite{Han2024TGRS}} & \makecell[c]{TGRS 2024} & \multicolumn{2}{c}{63.3}  \\ \hline
  \makecell[c]{ML-UDA~\cite{Luo2024JSTARs}} & \makecell[c]{JSTARs 2024} & \multicolumn{2}{c}{66.2}  \\ \hline
  \makecell[c]{CDST~\cite{Luo2024TGRS}} & \makecell[c]{TGRS 2024} & \multicolumn{2}{c}{72.8}  \\ \hline
  \makecell[c]{FIE-Net~\cite{ZhangJun2025TGRS}} & \makecell[c]{TGRS 2025} & \multicolumn{2}{c}{55.1}  \\ 
  \hline \hline
  \makecell[c]{\multirow{2}{*}{\textbf{Method}}} & \makecell[c]{\multirow{2}{*}{\textbf{Source}}} & \multicolumn{2}{c}{\textbf{Tasks on DIOR and DOTA datasets}} \\ \cline{3-4}
  {} & {} & \multicolumn{2}{c}{DIOR $\rightarrow$ DOTA-10} \\ \hline
  \makecell[c]{DCLDA~\cite{Biswas2024JSTARs}} & \makecell[c]{JSTARs 2024} & \multicolumn{2}{c}{50.6}  \\ \hline
  \makecell[c]{FIE-Net~\cite{ZhangJun2025TGRS}} & \makecell[c]{TGRS 2025} & \multicolumn{2}{c}{57.3}  \\ 
  \hline \hline
  \makecell[c]{\multirow{2}{*}{\textbf{Method}}} & \makecell[c]{\multirow{2}{*}{\textbf{Source}}} & \multicolumn{2}{c}{\textbf{Tasks on NWPU VHR-10 and DIOR datasets}} \\ \cline{3-4}
  {} & {} & \multicolumn{2}{c}{NWPU VHR-10 $\rightarrow$ DIOR} \\ \hline
  \makecell[c]{RFA-Net~\cite{Zhu2022JSTARs}} & \makecell[c]{JSTARs 2022} & \multicolumn{2}{c}{51.6}  \\ \hline
  \makecell[c]{MGDAT~\cite{Fang2025RS}} & \makecell[c]{RS 2025} & \multicolumn{2}{c}{55.1}  \\ 
  \hline \hline
  \makecell[c]{\multirow{2}{*}{\textbf{Method}}} & \makecell[c]{\multirow{2}{*}{\textbf{Source}}} & \multicolumn{2}{c}{\textbf{Tasks on UCAS-AOD and CARPK datasets}} \\ \cline{3-4}
  {} & {} & \makecell[c]{UCAS-AOD $\rightarrow$ CARPK} & \makecell[c]{UCARPK $\rightarrow$ UCAS-AOD} \\ \hline
  \makecell[c]{RST~\cite{Han2024TGRS}} & \makecell[c]{TGRS 2024} & \makecell[c]{76.2} & \makecell[c]{75.6}  \\ \hline
  \makecell[c]{FIE-Net~\cite{ZhangJun2025TGRS}} & \makecell[c]{TGRS 2025} & \makecell[c]{79.4} & \makecell[c]{78.9}  \\ 
  \hline \hline
  \makecell[c]{\multirow{2}{*}{\textbf{Method}}} & \makecell[c]{\multirow{2}{*}{\textbf{Source}}} & \multicolumn{2}{c}{\textbf{Tasks on UCAS-AOD and ISPRS Potsdam datasets}} \\ \cline{3-4}
  {} & {} & \multicolumn{2}{c}{UCAS-AOD $\rightarrow$ ISPRS Potsdam} \\ \hline
  \makecell[c]{ML-UDA~\cite{Luo2024JSTARs}} & \makecell[c]{JSTARs 2024} & \multicolumn{2}{c}{91.8}  \\ \hline
  \makecell[c]{CDST~\cite{Luo2024TGRS}} & \makecell[c]{TGRS 2024} & \multicolumn{2}{c}{92.6}  \\ 
  \hline \hline
  \makecell[c]{\multirow{2}{*}{\textbf{Method}}} & \makecell[c]{\multirow{2}{*}{\textbf{Source}}} & \multicolumn{2}{c}{\textbf{Tasks on HRRSD and SSDD datasets}} \\ \cline{3-4}
  {} & {} & \multicolumn{2}{c}{HRRSD $\rightarrow$ SSDD} \\ \hline
  \makecell[c]{HSANet~\cite{ZhangJun2022TGRS}} & \makecell[c]{TGRS 2022} & \multicolumn{2}{c}{58.1}  \\ \hline
  \makecell[c]{RST~\cite{Han2024TGRS}} & \makecell[c]{TGRS 2024} & \multicolumn{2}{c}{58.5}  \\ \hline
  \makecell[c]{FIE-Net~\cite{ZhangJun2025TGRS}} & \makecell[c]{TGRS 2025} & \multicolumn{2}{c}{58.5}  \\ 
  \hline \hline
  \makecell[c]{\multirow{2}{*}{\textbf{Method}}} & \makecell[c]{\multirow{2}{*}{\textbf{Source}}} & \multicolumn{2}{c}{\textbf{Tasks on LEVIR and SSDD datasets}} \\ \cline{3-4}
  {} & {} & \multicolumn{2}{c}{LEVIR $\rightarrow$ SSDD} \\ \hline
  \makecell[c]{HSANet~\cite{ZhangJun2022TGRS}} & \makecell[c]{TGRS 2022} & \multicolumn{2}{c}{58.2}  \\ \hline
  \makecell[c]{IDA~\cite{Pan2023TGRS}} & \makecell[c]{TGRS 2023} & \multicolumn{2}{c}{55.1}  \\ 
  \hline \hline
  \end{tabular}}
  \label{Tab12}
\end{table}
\subsubsection{Benchmark Performance on DA-RSDet}
In the DA-RSDet task, model performance is commonly evaluated using $AP/mAP$, $AR/mAR$, and $F1/mF$-score. For each class $i$, the Average Precision ($AP_i$) is defined as the area under the precision–recall curve, while the mean Average Precision ($mAP$) is the average of $AP_i$ across all classes. The Average Recall ($AR_i$) reflects the proportion of correctly detected objects, and $mAR$ is its mean across classes. The $F1$-score is defined as $2 \times Precision \times Recall / (Precision + Recall)$, with $mF$-score denoting the mean across classes. In this survey, $mAP@50$ ($mAP_{50}$) is adopted as the primary evaluation metric, as it is used by most existing methods.
\par We select the datasets in Tab.~\ref{Tab7} to systematically report the performance of UDA-RSDet methods, with the detailed tasks listed as follows.
\begin{itemize}
\item Adaptation between xView and DOTA \\ (xView $\rightarrow$ DOTA-3).
\item Adaptation between DIOR and DOTA \\ (DIOR $\rightarrow$ DOTA-10).
\item Adaptation between NWPU VHR-10 and DIOR \\ (NWPU VHR-10 $\rightarrow$ DIOR).
\item Adaptation between UCAS-AOD and CARPK \\ (UCAS-AOD $\rightarrow$ CARPK \& CARPK $\rightarrow$ UCAS-AOD).
\item Adaptation between UCAS-AOD and ISPRS Potsdam \\ (UCAS-AOD $\rightarrow$ ISPRS Potsdam).
\item Adaptation between HRRSD and SSDD \\ (HRRSD $\rightarrow$ SSDD).
\item Adaptation between LEVIR and SSDD \\ (LEVIR $\rightarrow$ SSDD).
\end{itemize}
\par Among UDA-RSDet tasks, the ``HRRSD $\rightarrow$ SSDD'' task represents an optical-to-SAR scenario, while the ``LEVIR $\rightarrow$ SSDD'' task corresponds to a typical SAR-to-SAR setting. All remaining tasks fall under the optical-to-optical category.
\par The results are summarized in Tab.~\ref{Tab12}. \textbf{Here, we first clarify that some methods use different experimental settings on the same datasets, such as different dataset splits and model choices. Therefore, the comparisons may not be strictly fair. This survey only reports the results from original paper listed in Tab.~\ref{Tab7}. For more details, please refer to the original papers.}
From the results and the detailed information provided in the original papers, it can be observed that recent approaches such as FIE-Net~\cite{ZhangJun2025TGRS} and CDST~\cite{Luo2024TGRS} consistently achieve leading performance across multiple cross-domain tasks. 
\par Based on experimental results and previous reviews, our analysis is as follows. Current UDA-RSDet methods mostly use hybrid strategies, with self-training at the core to gradually adapt to the target domain using high-quality pseudo-labels. Adversarial alignment acts as an auxiliary tool, often combined with class-wise constraints or contrastive learning to improve feature consistency.
\section{Future Works} % one page 
\label{future}
%% 针对图中4点分别讲一下
In recent years, deep learning based DA-RS methods has achieved notable progress across classification, segmentation, detection, and change detection. Despite the huge progress, several challenges remain for practical deployment and further breakthroughs. 
\par \textbf{Research on Limited Computation Resources.}
Most DA-RS methods rely on computationally demanding deep models, which restricts their use in real-world platforms such as satellites, UAVs, and edge devices. Future work should focus on lightweight architectures, efficient optimization, and model compression techniques to balance performance with practical deployment.
\par \textbf{Extension to Multiple-Source/Target Settings.}
Most DA-RS methods focus on a single-source to single-target setting, and only a few consider the issue concerning multi-source or multi-target. In practice, applications always face diverse domains with different conditions. Extending DA-RS to multi-source and multi-target settings can improve scalability and robustness, which requires new strategies for domain fusion and adaptive generalization.
\par \textbf{Further Exploration on Transformer Structure.} Transformers have achieved success in vision and multimodal learning, but their potential in DA-RS remains underexplored. Future methods should investigate how attention mechanisms and hierarchical representations can capture cross-domain invariances, improve semantic consistency, and enhance adaptability across tasks.
\par \textbf{Exploration on Foundation Model Paradigm.} Most foundation models are built on transformer architectures, their emergence brings new opportunities for DA-RS. Leveraging vision or vision-language foundation models and adapting them to remote sensing tasks can bring new advantages. They may provide stronger cross-domain generalization, support open-vocabulary recognition, and enable task transferability. Together, these advances point to a new paradigm for domain adaptation research.
\section{Conclusions}
\label{conclusion}
This survey provides a comprehensive survey of the recent advancements of deep learning based domain adaptation methods in remote sensing, which have rapidly evolved in recent years. We review existing related methods through a unified perspective that considers task categorization, input mode, supervision paradigm, and algorithmic granularity, thereby offering a systematic understanding of the field. Moreover, we summarize benchmarking efforts including widely used datasets and performance evaluations of state-of-the-art methods. Finally, the discussion of future research directions has outlined promising avenues for further exploration. We hope this survey helps advance the understanding of domain adaptation in remote sensing and serves as a reference to support and guide future research.

\bibliographystyle{IEEEtran}
\bibliography{main} % at least 300 references 5 pages 
%\onecolumn

\end{document}